\documentclass[12pt]{article}
\newcommand{\filename}{2LM-M\_LLR\_monomer\_v7}
\usepackage{authblk}
\usepackage{amsfonts}
\usepackage{amssymb}
\usepackage{amsmath}
\usepackage{amsthm}
\usepackage{framed}
\usepackage{graphicx}

\usepackage{enumitem}
\usepackage{geometry}
\usepackage{url}
\usepackage{booktabs}

\usepackage{multirow}

\usepackage{mdframed}
\usepackage{mathtools}
\usepackage{algorithm}  
\usepackage{algorithmic}

\newcommand{\lf}{\mathrm{lf}}  

\newcommand{\newone}{\marginpar{\bf NEW!}}

\newcommand{\eledeg}{\mathrm{eledeg}}  
\newcommand{\eledegC}{\mathrm{eledeg}_\mathrm{C}}   
\newcommand{\eledegT}{\mathrm{eledeg}_\mathrm{T}}   
\newcommand{\eledegF}{\mathrm{eledeg}_\mathrm{F}}   
\newcommand{\eledegX}{\mathrm{eledeg}_\mathrm{X}}   

\newcommand{\vion}{\mathrm{v}_\mathrm{ion}}  

\newcommand{\ttH}{{\tt H}}  
\newcommand{\ttC}{{\tt C}}  
\newcommand{\ttO}{{\tt O}}  
\newcommand{\ttN}{{\tt N}}  
\newcommand{\ttP}{{\tt P}}  
\newcommand{\ttF}{{\tt F}}  
\newcommand{\ttCl}{{\tt Cl}}  
\newcommand{\ttS}{{\tt S}}  
\newcommand{\ttSi}{{\tt Si}}

\newcommand{\oH}{\overline{{\tt H}}}  

\newcommand{\Z}{\mathbb{Z}}  
  
\newcommand{\C}{\mathbb{C}}  
\newcommand{\Co}{\mathbb{C}}

\newcommand{\anC}{\langle \mathbb{C} \rangle}  
\newcommand{\anpsi}{\langle \psi \rangle}  

\newcommand{\VH}{V_{\tt H}}

\newcommand{\R}{\mathbb{R}} 
\newcommand{\RK}{\mathbb{R}^K} 
\newcommand{\RKw}{\mathbb{R}^{K+1}}

\newcommand{\deghyd}{\deg^\mathrm{hyd}}

\newcommand{\FrC}{\mathcal{F}^\mathrm{C}} 
\newcommand{\FrT}{\mathcal{F}^\mathrm{T}} 
\newcommand{\FrF}{\mathcal{F}^\mathrm{F}} 
\newcommand{\FrX}{\mathcal{F}^\mathrm{X}}  

\newcommand{\dcp}{\mathrm{dcp}}

\newcommand{\Vleaf}{V_\mathrm{leaf}} 
\newcommand{\Eleaf}{E_\mathrm{leaf}} 

\newcommand{\sint}{\sigma_\mathrm{int}} 
\newcommand{\sce}{\sigma_\mathrm{ce}}

\newcommand{\Ez}{E_{(0/1)}}
\newcommand{\Ew}{E_{(\geq 1)}}
\newcommand{\Et}{E_{(\geq 2)}}
\newcommand{\Eew}{E_{(=1)}}

\newcommand{\Iz}{I_{(0/1)}}
\newcommand{\Iw}{I_{(\geq 1)}}
\newcommand{\It}{I_{(\geq 2)}}
\newcommand{\Iew}{I_{(=1)}}

\newcommand{\Gac}{\Gamma_\mathrm{ac}} 
\newcommand{\Gacs}{\Gamma_\mathrm{ac,<}} 
\newcommand{\Gace}{\Gamma_\mathrm{ac,=}} 
\newcommand{\Gacl}{\Gamma_\mathrm{ac,>}}

\newcommand{\tGacC}{\widetilde{\Gamma}_\mathrm{ac}^\mathrm{C}}  
\newcommand{\tGacT}{\widetilde{\Gamma}_\mathrm{ac}^\mathrm{T}} 
\newcommand{\tGacF}{\widetilde{\Gamma}_\mathrm{ac}^\mathrm{F}}  
\newcommand{\tGacCT}{\widetilde{\Gamma}_\mathrm{ac}^\mathrm{CT}}  
\newcommand{\tGacTC}{\widetilde{\Gamma}_\mathrm{ac}^\mathrm{TC}}  
\newcommand{\tGacCF}{\widetilde{\Gamma}_\mathrm{ac}^\mathrm{CF}}  
\newcommand{\tGacTF}{\widetilde{\Gamma}_\mathrm{ac}^\mathrm{TF}}

\newcommand{\tLdgX}{\widetilde{\Lambda}_\mathrm{dg}^\mathrm{X}}   
\newcommand{\tLdgC}{\widetilde{\Lambda}_\mathrm{dg}^\mathrm{C}}   
\newcommand{\tLdgT}{\widetilde{\Lambda}_\mathrm{dg}^\mathrm{T}}   
\newcommand{\tLdgF}{\widetilde{\Lambda}_\mathrm{dg}^\mathrm{F}}

\newcommand{\tGecC}{\widetilde{\Gamma}_\mathrm{ec}^\mathrm{C}}  
\newcommand{\tGecT}{\widetilde{\Gamma}_\mathrm{ec}^\mathrm{T}} 
\newcommand{\tGecF}{\widetilde{\Gamma}_\mathrm{ec}^\mathrm{F}}  
\newcommand{\tGecCT}{\widetilde{\Gamma}_\mathrm{ec}^\mathrm{CT}}  
\newcommand{\tGecTC}{\widetilde{\Gamma}_\mathrm{ec}^\mathrm{TC}}  
\newcommand{\tGecCF}{\widetilde{\Gamma}_\mathrm{ec}^\mathrm{CF}}  
\newcommand{\tGecTF}{\widetilde{\Gamma}_\mathrm{ec}^\mathrm{TF}}

\newcommand{\typ}{\mathrm{t}}

\newcommand{\w}{w}
\newcommand{\x}{x}

\newcommand{\ta}{{\tt a}}
\newcommand{\tb}{{\tt b}}
\newcommand{\Ldg}{\Lambda_{\mathrm{dg}}}

\newcommand{\fc}{\mathrm{fc}} 
\newcommand{\betar}{\beta_\mathrm{r}}

\newcommand{\val}{\mathrm{val}}

\newcommand{\inte}{\mathrm{int}}

\newcommand{\F}{\mathcal{F}}

\newcommand{\T}{\mathcal{T}}

\newcommand{\nint}{\mathrm{n}^\mathrm{int}}

\newcommand{\h}{\mathrm{ht}}

\newcommand{\cs}{\mathrm{cs}}
\newcommand{\ch}{\mathrm{ch}}

\newcommand{\dg}{\mathrm{dg}}

\newcommand{\na}{\mathrm{na}}
 
\newcommand{\naX}{\mathrm{na}_\mathrm{X}}

\newcommand{\naC}{\mathrm{na}_\mathrm{C}}
\newcommand{\naT}{\mathrm{na}_\mathrm{T}}
\newcommand{\naF}{\mathrm{na}_\mathrm{F}}

\newcommand{\ecX}{\mathrm{ec}_\mathrm{X}}

\newcommand{\ecC}{\mathrm{ec}_\mathrm{C}}
\newcommand{\ecT}{\mathrm{ec}_\mathrm{T}}
\newcommand{\ecF}{\mathrm{ec}_\mathrm{F}}
\newcommand{\ecCT}{\mathrm{ec}_\mathrm{CT}}
\newcommand{\ecTC}{\mathrm{ec}_\mathrm{TC}}
\newcommand{\ecTF}{\mathrm{ec}_\mathrm{TF}}
\newcommand{\ecCF}{\mathrm{ec}_\mathrm{CF}}
 
\newcommand{\acX}{\mathrm{ac}_\mathrm{X}}

\newcommand{\acC}{\mathrm{ac}_\mathrm{C}}
\newcommand{\acT}{\mathrm{ac}_\mathrm{T}}
\newcommand{\acF}{\mathrm{ac}_\mathrm{F}}
\newcommand{\acCT}{\mathrm{ac}_\mathrm{CT}}
\newcommand{\acTC}{\mathrm{ac}_\mathrm{TC}}
\newcommand{\acTF}{\mathrm{ac}_\mathrm{TF}}
\newcommand{\acCF}{\mathrm{ac}_\mathrm{CF}}

\newcommand{\bdX}{\mathrm{bd}_\mathrm{X}}

\newcommand{\bdC}{\mathrm{bd}_\mathrm{C}}
\newcommand{\bdT}{\mathrm{bd}_\mathrm{T}}
\newcommand{\bdF}{\mathrm{bd}_\mathrm{F}}
\newcommand{\bdCT}{\mathrm{bd}_\mathrm{CT}}
\newcommand{\bdTC}{\mathrm{bd}_\mathrm{TC}}
\newcommand{\bdTF}{\mathrm{bd}_\mathrm{TF}}
\newcommand{\bdCF}{\mathrm{bd}_\mathrm{CF}}

\newcommand{\ns}{\mathrm{ns}}

\newcommand{\ec}{\mathrm{ec}}
\newcommand{\ac}{\mathrm{ac}}

\newcommand{\bl}{\mathrm{bl}}

\newcommand{\bd}{\mathrm{bd}}

\newcommand{\UB}{\mathrm{UB}}
\newcommand{\LB}{\mathrm{LB}}

\newcommand{\ex}{\mathrm{ex}}

\newcommand{\GC}{G_\mathrm{C}}

\newcommand{\mC}{m_\mathrm{C}}

\newcommand{\hC}{h^\mathrm{C}}
\newcommand{\hT}{h^\mathrm{T}} 
 
\newcommand{\hX}{h^\mathrm{X}}

\newcommand{\VF}{V_\mathrm{F}}
\newcommand{\VT}{V_\mathrm{T}}
\newcommand{\VC}{V_\mathrm{C}} 
 
\newcommand{\VX}{V_\mathrm{X}}

\newcommand{\ET}{E_\mathrm{T}}
\newcommand{\EC}{E_\mathrm{C}}

\newcommand{\EF}{E_\mathrm{F}}
\newcommand{\ECT}{E_\mathrm{CT}}
\newcommand{\ETC}{E_\mathrm{TC}}
\newcommand{\ETF}{E_\mathrm{TF}}

\newcommand{\ECF}{E_\mathrm{CF}}

\newcommand{\EX}{E_\mathrm{X}}

\newcommand{\vT}{{v^\mathrm{T}}}
\newcommand{\vC}{{v^\mathrm{C}}} 
  
\newcommand{\vX}{{v^\mathrm{X}}}

\newcommand{\vF}{{v^\mathrm{F}}}

\newcommand{\eF}{{e^\mathrm{F}}}
\newcommand{\eT}{{e^\mathrm{T}}}
\newcommand{\eC}{{e^\mathrm{C}}} 
 
\newcommand{\eX}{{e^\mathrm{X}}}

\newcommand{\eCF}{{e^\mathrm{CF}}}

\newcommand{\eCT}{{e^\mathrm{CT}}}
\newcommand{\eTC}{{e^\mathrm{TC}}}
\newcommand{\eTF}{{e^\mathrm{TF}}}

\newcommand{\tT}{{t_\mathrm{T}}}
\newcommand{\tC}{{t_\mathrm{C}}} 
\newcommand{\tF}{{t_\mathrm{F}}} 
 
\newcommand{\tX}{{t_\mathrm{X}}}

\newcommand{\IC}{{I_\mathrm{C}}}

\newcommand{\degCint}{{\deg_\mathrm{C}^\mathrm{int}}}
\newcommand{\degTint}{{\deg_\mathrm{T}^\mathrm{int}}}
\newcommand{\degFint}{{\deg_\mathrm{F}^\mathrm{int}}}
\newcommand{\degXint}{{\deg_\mathrm{X}^\mathrm{int}}}

\newcommand{\hyddeg}{\mathrm{hyddeg}}

\newcommand{\hyddegX}{\mathrm{hyddeg}^\mathrm{X}}

\newcommand{\degCex}{{\deg_\mathrm{C}^\mathrm{ex}}}
\newcommand{\degTex}{{\deg_\mathrm{T}^\mathrm{ex}}}
\newcommand{\degFex}{{\deg_\mathrm{F}^\mathrm{ex}}}
\newcommand{\degXex}{{\deg_\mathrm{X}^\mathrm{ex}}}
\newcommand{\degF}{{\deg^\mathrm{F}}}
\newcommand{\degT}{{\deg^\mathrm{T}}}
\newcommand{\degC}{{\deg^\mathrm{C}}} 
 
\newcommand{\degX}{{\deg^\mathrm{X}}}

\newcommand{\degCT}{\deg_\mathrm{CT}}
\newcommand{\degTC}{\deg_\mathrm{TC}}

\newcommand{\degCTT}{\deg^\mathrm{CT}_\mathrm{T}}
\newcommand{\degTCT}{\deg^\mathrm{TC}_\mathrm{T}}
\newcommand{\degCFF}{\deg^\mathrm{CF}_\mathrm{F}}
\newcommand{\degTFF}{\deg^\mathrm{TF}_\mathrm{F}}

\newcommand{\tldgC}{{\widetilde{\deg}_\mathrm{C}} }

\newcommand{\cF}{{c_\mathrm{F}}}

\newcommand{\kC}{{k_\mathrm{C}}}

\newcommand{\chiF}{{\chi^\mathrm{F}}} 
\newcommand{\dclrF}{\delta_{\chi}^\mathrm{F}}
\newcommand{\clrF}{\mathrm{clr}^{\mathrm{F}}}

\newcommand{\chiT}{{\chi^\mathrm{T}}}
\newcommand{\dclrT}{\delta_{\chi}^\mathrm{T}}
\newcommand{\clrT}{\mathrm{clr}^{\mathrm{T}}}

\newcommand{\tail}{\mathrm{tail}} 
\newcommand{\hd}{\mathrm{head}}

\newcommand{\dlfrF}{\delta_\mathrm{fr}^\mathrm{F}}

\newcommand{\dlfrC}{\delta_\mathrm{fr}^\mathrm{C}} 
 
\newcommand{\dlfrX}{\delta_\mathrm{fr}^\mathrm{X}}
 
\newcommand{\ddgF}{\delta_\mathrm{dg}^\mathrm{F}}
\newcommand{\ddgT}{\delta_\mathrm{dg}^\mathrm{T}}
\newcommand{\ddgC}{\delta_\mathrm{dg}^\mathrm{C}} 
 
\newcommand{\ddgX}{\delta_\mathrm{dg}^\mathrm{X}}

\newcommand{\ddgFint}{\delta_\mathrm{dg,F}^\mathrm{int}}
\newcommand{\ddgTint}{\delta_\mathrm{dg,T}^\mathrm{int}}
\newcommand{\ddgCint}{\delta_\mathrm{dg,C}^\mathrm{int}}  
\newcommand{\ddgXint}{\delta_\mathrm{dg,X}^\mathrm{int}}

\newcommand{\bF}{\beta^\mathrm{F}}
\newcommand{\bT}{\beta^\mathrm{T}}
\newcommand{\bC}{\beta^\mathrm{C}} 
\newcommand{\bX}{\beta^\mathrm{X}}
  
\newcommand{\bCT}{\beta^\mathrm{CT}}
\newcommand{\bTC}{\beta^\mathrm{TC}} 
\newcommand{\bTF}{\beta^\mathrm{TF}} 
\newcommand{\bCF}{\beta^\mathrm{CF}} 
\newcommand{\bXF}{\beta^\mathrm{XF}} 
\newcommand{\bsF}{\beta^{*\mathrm{F}}}

\newcommand{\bFex}{\beta^\mathrm{F}_\mathrm{ex}} 
\newcommand{\bTex}{\beta^\mathrm{T}_\mathrm{ex}} 
\newcommand{\bCex}{\beta^\mathrm{C}_\mathrm{ex}} 
\newcommand{\bXex}{\beta^\mathrm{X}_\mathrm{ex}}

\newcommand{\delbF}{\delta_{\beta}^\mathrm{F}}
\newcommand{\delbT}{\delta_{\beta}^\mathrm{T}}
\newcommand{\delbC}{\delta_{\beta}^\mathrm{C}}

\newcommand{\delbCT}{\delta_{\beta}^\mathrm{CT}}
\newcommand{\delbTC}{\delta_{\beta}^\mathrm{TC}}

\newcommand{\delbsF}{\delta_{\beta}^{*\mathrm{F}}} 
\newcommand{\delbX}{\delta_{\beta}^\mathrm{X}}

\newcommand{\aF}{{\alpha}^\mathrm{F}}
\newcommand{\aT}{{\alpha}^\mathrm{T}}
\newcommand{\aC}{{\alpha}^\mathrm{C}}  
 
\newcommand{\aX}{{\alpha}^\mathrm{X}}
\newcommand{\aCT}{{\alpha}^\mathrm{CT}}
\newcommand{\aTC}{{\alpha}^\mathrm{TC}}

\newcommand{\aCF}{{\alpha}^\mathrm{CF}}  
\newcommand{\aTF}{{\alpha}^\mathrm{TF}}

\newcommand{\delaC}{\delta_\mathrm{\alpha}^{\mathrm{C}}}
\newcommand{\delaT}{\delta_\mathrm{\alpha}^{\mathrm{T}}}
\newcommand{\delaF}{\delta_\mathrm{\alpha}^{\mathrm{F}}}
\newcommand{\delaX}{\delta_\mathrm{\alpha}^{\mathrm{X}}}

\newcommand{\dlnsF}{\delta_{\mathrm{ns}}^\mathrm{F}}
\newcommand{\dlnsT}{\delta_{\mathrm{ns}}^\mathrm{T}}
\newcommand{\dlnsC}{\delta_{\mathrm{ns}}^\mathrm{C}} 
\newcommand{\dlnsX}{\delta_{\mathrm{ns}}^\mathrm{X}}

\newcommand{\dlacF}{\delta_{\mathrm{ac}}^\mathrm{F}}
\newcommand{\dlacT}{\delta_{\mathrm{ac}}^\mathrm{T}}
\newcommand{\dlacC}{\delta_{\mathrm{ac}}^\mathrm{C}}

\newcommand{\dlacCT}{\delta_{\mathrm{ac}}^\mathrm{CT}}
\newcommand{\dlacTC}{\delta_{\mathrm{ac}}^\mathrm{TC}}

\newcommand{\dlacCF}{\delta_{\mathrm{ac}}^\mathrm{CF}} 
\newcommand{\dlacTF}{\delta_{\mathrm{ac}}^\mathrm{TF}} 
\newcommand{\dlacX}{\delta_{\mathrm{ac}}^\mathrm{X}}

\newcommand{\DlacFp}{\Delta_{\mathrm{ac}}^\mathrm{F+}}
\newcommand{\DlacTp}{\Delta_{\mathrm{ac}}^\mathrm{T+}}
\newcommand{\DlacCp}{\Delta_{\mathrm{ac}}^\mathrm{C+}}

\newcommand{\DlacCTp}{\Delta_{\mathrm{ac}}^\mathrm{CT+}}
\newcommand{\DlacTCp}{\Delta_{\mathrm{ac}}^\mathrm{TC+}}

\newcommand{\DlacCFp}{\Delta_{\mathrm{ac}}^\mathrm{CF+}} 
\newcommand{\DlacTFp}{\Delta_{\mathrm{ac}}^\mathrm{TF+}} 
\newcommand{\DlacXp}{\Delta_{\mathrm{ac}}^\mathrm{X+}}

\newcommand{\DlacFm}{\Delta_{\mathrm{ac}}^\mathrm{F-}}
\newcommand{\DlacTm}{\Delta_{\mathrm{ac}}^\mathrm{T-}}
\newcommand{\DlacCm}{\Delta_{\mathrm{ac}}^\mathrm{C-}}

\newcommand{\DlacCTm}{\Delta_{\mathrm{ac}}^\mathrm{CT-}}
\newcommand{\DlacTCm}{\Delta_{\mathrm{ac}}^\mathrm{TC-}}

\newcommand{\DlacCFm}{\Delta_{\mathrm{ac}}^\mathrm{CF-}} 
\newcommand{\DlacTFm}{\Delta_{\mathrm{ac}}^\mathrm{TF-}} 
\newcommand{\DlacXm}{\Delta_{\mathrm{ac}}^\mathrm{X-}}

\newcommand{\dlecF}{\delta_{\mathrm{ec}}^\mathrm{F}}
\newcommand{\dlecT}{\delta_{\mathrm{ec}}^\mathrm{T}}
\newcommand{\dlecC}{\delta_{\mathrm{ec}}^\mathrm{C}}

\newcommand{\dlecCTC}{\delta_{\mathrm{ec,C}}^\mathrm{CT}}

\newcommand{\dlecTCC}{\delta_{\mathrm{ec,C}}^\mathrm{TC}}

\newcommand{\dlecCFC}{\delta_{\mathrm{ec,C}}^\mathrm{CF}}

\newcommand{\dlecTFT}{\delta_{\mathrm{ec,T}}^\mathrm{TF}} 
 
\newcommand{\dlecX}{\delta_{\mathrm{ec}}^\mathrm{X}}

\newcommand{\DlecFp}{\Delta_{\mathrm{ec}}^\mathrm{F+}}
\newcommand{\DlecTp}{\Delta_{\mathrm{ec}}^\mathrm{T+}}
\newcommand{\DlecCp}{\Delta_{\mathrm{ec}}^\mathrm{C+}}

\newcommand{\DlecCTp}{\Delta_{\mathrm{ec}}^\mathrm{CT+}}
\newcommand{\DlecTCp}{\Delta_{\mathrm{ec}}^\mathrm{TC+}}

\newcommand{\DlecCFp}{\Delta_{\mathrm{ec}}^\mathrm{CF+}} 
\newcommand{\DlecTFp}{\Delta_{\mathrm{ec}}^\mathrm{TF+}} 
\newcommand{\DlecXp}{\Delta_{\mathrm{ec}}^\mathrm{X+}}

\newcommand{\DlecFm}{\Delta_{\mathrm{ec}}^\mathrm{F-}}
\newcommand{\DlecTm}{\Delta_{\mathrm{ec}}^\mathrm{T-}}
\newcommand{\DlecCm}{\Delta_{\mathrm{ec}}^\mathrm{C-}}

\newcommand{\DlecCTm}{\Delta_{\mathrm{ec}}^\mathrm{CT-}}
\newcommand{\DlecTCm}{\Delta_{\mathrm{ec}}^\mathrm{TC-}}

\newcommand{\DlecCFm}{\Delta_{\mathrm{ec}}^\mathrm{CF-}} 
\newcommand{\DlecTFm}{\Delta_{\mathrm{ec}}^\mathrm{TF-}} 
\newcommand{\DlecXm}{\Delta_{\mathrm{ec}}^\mathrm{X-}}

\begin{document} 

\begin{center}
   {\Large\bf 
   An Inverse QSAR Method Based on 
   Linear Regression and Integer Programming}
\end{center} 

\begin{center}
Jianshen Zhu$^1$, 
Naveed Ahmed Azam$^1$, 
Kazuya Haraguchi$^{1}$, 
Liang Zhao$^2$, 
Hiroshi Nagamochi$^1$ 
 and  
 Tatsuya Akutsu$^3$ 
\end{center} 
%
%
{\small 
 1.    Department of Applied Mathematics and Physics, Kyoto University, Kyoto 606-8501, Japan\\
  2.  Graduate School of Advanced Integrated Studies in Human Survavibility   (Shishu-Kan),  
  Kyoto University, Kyoto 606-8306, Japan \\
  3.  Bioinformatics Center,  Institute for Chemical Research, 
  Kyoto University, Uji 611-0011, Japan 
}

\begin{quote}  
{\bf Abstract}\\  
Recently a novel framework has been proposed for designing 
the molecular structure of chemical compounds
 using both artificial neural networks (ANNs)
 and mixed integer linear programming (MILP).
In the framework, we  first  define 
 a feature vector $f(\C)$ of a chemical graph  $\C$
and construct  an ANN that maps $x=f(\C)$ to 
 a predicted value $\eta(x)$ of a chemical property $\pi$ to $\C$.
 After this,  we formulate  an MILP that  simulates
  the  computation process of $f(\C)$ from $\C$  and 
  that of $\eta(x)$ from $x$.
Given a  target value $y^*$ of the chemical property $\pi$, 
we   infer   a chemical graph $\C^\dagger$
such that $\eta(f(\C^\dagger))=y^*$ by solving the MILP.  
In this paper, we use linear regression to construct a prediction function
$\eta$ instead of ANNs.
For this, we derive an MILP formulation that simulates the computation process
of a prediction function by linear regression.
The results of computational experiments suggest 
our method can infer  chemical graphs 
with around up to 50 non-hydrogen atoms.



\end{quote}

\section{Introduction}\label{sec:introduction}

\noindent {\bf Background~}
Analysis of chemical compounds is one of the important applications of
intelligent computing.
Indeed, various machine learning methods have been applied to
the prediction of chemical activities from their structural data,
where such a problem is often referred to as
\emph{quantitative structure activity relationship} (QSAR)
\cite{Lo18,Tetko20}.
Recently, neural networks and deep-learning technologies have extensively
been applied to QSAR \cite{Ghasemi18}.

In addition to QSAR, extensive studies have been done on
inverse quantitative structure activity relationship
(inverse QSAR), which seeks for chemical structures having
desired chemical activities under some constraints.
Since it is difficult to directly handle chemical structures
in both QSAR and inverse QSAR,
chemical compounds are usually represented 
as vectors of real or integer numbers,
which are often called \emph{descriptors} in chemoinformatics and
correspond to \emph{feature vectors} in machine learning.
One major approach in inverse QSAR is 
to infer feature vectors from given chemical activities and constraints
and then reconstruct chemical structures from these feature
vectors~\cite{Miyao16,Ikebata17,Rupakheti15},
where chemical structures are usually treated as undirected graphs.
However, the  reconstruction itself is a challenging task
because the number of possible chemical graphs is huge.
 For example, 
chemical graphs with up to 30 atoms (vertices)
{\tt C}, {\tt N}, {\tt O}, and  {\tt S}
may  exceed~$10^{60}$~\cite{BMG96}. 
Indeed, it is NP-hard to infer a chemical graph from a given feature vector
except for some simple cases~\cite{AFJS12}.  
Due to this inherent difficulty, most existing methods for inverse QSAR
do not guarantee optimal or exact solutions.

As a new approach,
extensive studies have recently been done for inverse QSAR using 
\emph{artificial neural networks} (ANNs),
especially using graph convolutional networks~\cite{Kipf16}.
For example, recurrent neural networks~\cite{Segler18,Yang17}, 
variational autoencoders~\cite{Gomez18}, 
grammar variational autoencoders~\cite{Kusner17},
generative adversarial networks~\cite{DeCao18},
and invertible flow models~\cite{Madhawa19,Shi20}
have been applied.
However, these methods do not yet guarantee optimal or exact solutions.  

\begin{figure}[!ht]  \begin{center}
\includegraphics[width=.77\columnwidth]{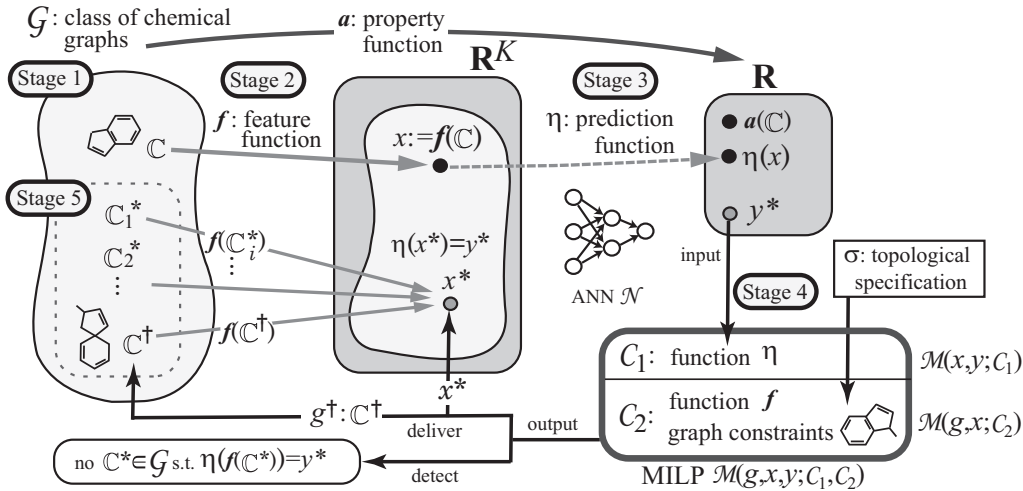}
\end{center} \caption{An illustration of a framework for inferring
a set of chemical graphs $\C^*$.   } 
\label{fig:framework}  \end{figure}    

\smallskip
\noindent {\bf Framework~}
Akutsu and Nagamochi~\cite{AN19} proved that  
the computation process of a given ANN can be simulated
with a mixed integer linear programming (MILP).
Based on this,
a novel  framework for inferring chemical graphs has been developed
\cite{ACZSNA20,ZZCSNA20}, 
as illustrated in Figure~\ref{fig:framework}. 
It constructs a prediction function in the first phase and
infers a chemical graph in the second phase. 
The first phase of the framework consists of three stages.
In Stage~1, we choose a chemical property $\pi$ and a class $\mathcal{G}$ 
of graphs, where a property function
$a$ is defined so that $a(\C)$ is the value of $\pi$  for a compound $\C\in \mathcal{G}$,
and collect a data set $D_{\pi}$ of chemical graphs in  $\mathcal{G}$ 
such that $a(\C)$ is available for every $\C \in D_{\pi}$.
In Stage~2, we introduce a feature function $f: \mathcal{G}\to \mathbb{R}^K$ 
for a positive integer $K$.    
In Stage~3, we construct a prediction function $\eta$ 
with an ANN $\mathcal{N}$ that,  
given a   vector  $x\in \mathbb{R}^K$, 
returns a value $y=\eta(x)\in \mathbb{R}$    
so that $\eta(f(\C))$ serves as a predicted value
to the real value $a(\C)$ of $\pi$ for each $\C\in D_\pi$.  
Given a  target chemical value $y^*$,
the  second phase infers  chemical graphs $\C^*$
with $\eta(f(\C^*))=y^*$ in the next two stages. 
We have obtained a  feature function $f$ and a  prediction function $\eta$
and call an additional constraint on the substructures of target chemical graphs 
a {\em topological specification}. 
In Stage~4, we prepare the following two  MILP formulations: 
\begin{enumerate}[nosep,  leftmargin=*]
\item[-]
 MILP $\mathcal{M}(x,y;\mathcal{C}_1)$
with a set $\mathcal{C}_1$ of linear constraints on variables $x$ and $y$
(and some other auxiliary variables) 
  simulates the  process of computing $y:=\eta(x)$ from a vector $x$; and
\item[-]
 MILP $\mathcal{M}(g,x;\mathcal{C}_2)$
with a set $\mathcal{C}_2$ of linear constraints on  variable  $x$ and
 a variable vector  $g$ that represents a chemical graph $\C$
(and some other auxiliary variables)  
  simulates the  process of computing $x:=f(\C)$ from a chemical graph $\C$
and chooses a chemical graph $\C$ that satisfies the given topological specification
$\sigma$. 
\end{enumerate} 
Given a target value $y^*\in \mathbb{R}$, we solve the combined
MILP $\mathcal{M}(g,x,y;\mathcal{C}_1,\mathcal{C}_2)$
to find  a feature vector $x^*\in \mathbb{R}^K$
 and a chemical graph $\C^{\dagger}$  with the specification
$\sigma$ such that $f(\C^\dagger)=x^*$ and  $\eta(x^*)=y^*$
(where if the MILP instance is infeasible then this suggests that there 
does  not exist
such a desired chemical graph).
In Stage~5,  we generate other  chemical graphs $\C^*$
such that $\eta(f(\C^*))=y^*$ based on the output chemical graph $\C^\dagger$.

MILP formulations required in Stage~4 have been designed   
for chemical compounds with 
cycle index 0 (i.e., acyclic) \cite{ZZCSNA20,AZSSSZNA20},
cycle index 1~\cite{IAWSNA20} and 
cycle index 2~\cite{ZCSNA20}, where no sophisticated topological  specification
was available yet. 
Azam~et~al.~\cite{AZSSSZNA20} 
 introduced a restricted class of acyclic graphs 
 that is  characterized  by an integer ${\rho}$, called
 a ``branch-parameter''
such that the restricted class still covers  most of the acyclic
chemical compounds in the database. 
 Akutsu and Nagamochi~\cite{AN20} 
extended the idea to define a restricted class of  cyclic graphs,
called ``${\rho}$-lean cyclic graphs'' 
and introduced a set of flexible rules for  describing a topological specification.
Recently, Tanaka et~al.~\cite{TZAHZNA21} used a decision tree
to construct a prediction function $\eta$ in Stage~3 in the framework
and derived an MILP $\mathcal{M}(x,y;\mathcal{C}_1)$ 
that simulates the computation process of  a decision~tree.

\smallskip
\noindent {\bf Two-layered Model~}
Recently Shi et~al.~\cite{SZAHZNA21} 
  proposed a new model, called a {\em two-layered model}  for representing
the feature of a chemical graph in order to deal with an arbitrary graph 
in the framework and refined  
the set of rules for  describing a topological specification
 so that a prescribed structure 
can be included in both of the acyclic and cyclic parts of  $\C$. 
In the two-layered model, a chemical graph  $\C$
with a parameter ${\rho}\geq 1$ is regarded as two parts:
the exterior and the interior of the hydrogen-suppressed chemical graph  
$\anC$ obtained from  $\C$ by removing hydrogen.
The exterior  consists of  maximal acyclic induced subgraphs with height
at most ${\rho}$ in $\anC$ and
the interior is the connected subgraph of $\anC$ 
obtained by ignoring the exterior.  
 Shi et~al.~\cite{SZAHZNA21}  defined 
 a feature vector $f(\C)$ of a chemical graph  $\C$
to be a combination of the frequency of adjacent atom pairs in the interior  and
the frequency of chemical acyclic graphs among the set of  chemical rooted trees $T_u$
rooted at interior-vertices $u$. 
Recently, Tanaka et~al.~\cite{TZAHZNA21} 
extend the model to treat a chemical graph with hydrogens directly
so that   more variety of chemical rooted trees   represent
 the feature of the exterior. 

\smallskip
\noindent {\bf Contribution~} 
In this paper, we first make a slight modification to a model of chemical graphs
proposed by Tanaka et~al.~\cite{TZAHZNA21}
so that we can treat a chemical element with multi-valence such as sulfur {\tt S}
and a chemical graph with cations and anions. 

The quality of a prediction function $\eta$ constructed in Stage~3
is one of the most important factors in the framework.
It is also pointed out that overfitting is a major issue in ANN-based approaches
for QSAR because ANNs have many parameters to be optimized \cite{Ghasemi18}. 
Tanaka et~al.~\cite{TZAHZNA21}  observed 
that decision trees perform better than ANNs for some chemical properties
and 
 used a decision tree for constructing a prediction function $\eta$ in Stage~3.  
In this paper, we use linear regression to construct a prediction function in Stage~3.  
Linear regression is much simpler than ANNs and decision trees and
thereby we regard the performance of a prediction function by  linear regression 
as the basis for other more sophisticated machine learning methods. 
In this paper, we derive  an MILP formulation  $\mathcal{M}(x,y;\mathcal{C}_1)$
 that simulates the computation process of a prediction function by linear regression.
For  an MILP  formulation $\mathcal{M}(g,x;\mathcal{C}_2)$ 
that   represents  a feature function $f$ and a specification $\sigma$ in Stage~4,  
we can use the same  formulation proposed by Tanaka et~al.~\cite{TZAHZNA21} 
with a slight modification 
 (the detail of the MILP $\mathcal{M}(g,x;\mathcal{C}_2)$ can be found in  
Appendix~\ref{sec:full_milp}).  
To generate target chemical graphs $\C^*$ in Stage~5,
we can also use the dynamic programming algorithm 
due to Tanaka et~al.~\cite{TZAHZNA21} 
with a slight modification and omit the details in this paper.  

We implemented the framework based on the refined two-layered model 
and a prediction function by linear regression. 
 The results  of our computational experiments 
 reveal a set of chemical properties to which 
 a prediction function constructed with linear regression  on our feature function
 performs well. 
We also observe that 
 the proposed method can infer chemical graphs 
with  up to 50 non-hydrogen atoms. 

The paper is organized as follows.  
Section~\ref{sec:preliminary} introduces some notions on graphs,
 a modeling of chemical compounds and a choice of descriptors.  
Section~\ref{sec:2LM} describes our modification to the two-layered model.
Section~\ref{sec:linear_regression}  reviews the idea of linear regression
and formulates an MILP $\mathcal{M}(x,y;\mathcal{C}_1)$ 
that simulates a process of computing 
a prediction function constructed by linear regression.
%
%
%
%
Section~\ref{sec:experiment} reports the results on some computational 
experiments conducted for  18 chemical properties such as 
vapor density and 
optical rotation.
Section~\ref{sec:conclude} makes some concluding remarks.   
Some technical details are given in Appendices:   
 Appendix~\ref{sec:descriptor} for  all descriptors in our feature function; 
 Appendix~\ref{sec:specification} for a full description of 
a topological specification;
Appendix~\ref{sec:test_instances} for the detail of test instances
used in our computational experiment for Stages~4 and 5; and 
 Appendix~\ref{sec:full_milp}
  for the details of   our MILP formulation  $\mathcal{M}(g,x;\mathcal{C}_2)$.

%

\section{Preliminary}\label{sec:preliminary}

This section  introduces some notions and terminologies on graphs,
  modeling of chemical compounds and our choice of descriptors. 
 
Let $\mathbb{R}$, $\mathbb{R}_+$, $\mathbb{Z}$  and $\mathbb{Z}_+$ 
denote the sets of reals,  non-negative reals, 
integers and non-negative integers, respectively.
For two integers $a$ and $b$, let $[a,b]$ denote the set of 
integers $i$ with $a\leq i\leq b$.

\bigskip\noindent
{\bf  Graph} 
Given a  graph $G$, let $V(G)$ and $E(G)$ denote the sets
of vertices and edges, respectively.     
For a subset $V'\subseteq V(G)$ (resp., $E'\subseteq E(G))$ of
a graph $G$, 
let $G-V'$ (resp., $G-E'$) denote the graph obtained from $G$
by removing the vertices in $V'$ (resp.,  the edges in $E'$),
where we remove all edges incident to a vertex in $V'$
in $G-V'$. 
An edge subset $E'\subseteq E(G)$ in a connected graph $G$ is called
{\em separating} (resp., {\em non-separating})
if $G-E'$ remains connected 
(resp., $G-E'$ becomes disconnected). 
The {\em rank}  $\mathrm{r}(G)$ of a graph $G$  is defined to be 
the minimum $|F|$ of an edge subset $F\subseteq E(G)$
such that $G-F$ contains no cycle, where  $\mathrm{r}(G)=|E(G)|-|V(G)|+1$
for a connected graph $G$.  
Observe that   $\mathrm{r}(G-E')=\mathrm{r}(G)-|E'|$ holds
for any non-separating edge subset $E'\subseteq E(G)$. 
An edge $e\in E(G)$ in a connected graph $G$
  is called a {\em bridge} if $\{e\}$ is separating.
For a connected cyclic graph $G$, an edge $e$ is called a {\em core-edge} if
it is in a cycle of $G$ or is a bridge $e=u_1u_2$ such that
each of the connected graphs $G_i$, $i=1,2$ of $G-e$ contains a cycle. 
A vertex incident to a core-edge is called a {\em core-vertex} of $G$. 
A path with two end-vertices $u$ and $v$ is called a {\em $u,v$-path}. 
 
A vertex designated in a graph $G$ is called a {\em root}.
In this paper, we designate at most two vertices as roots, 
and denote by $\mathrm{Rt}(G)$ the set of roots of $G$.
We call a graph $G$   {\em rooted} (resp., {\em bi-rooted})
if $|\mathrm{Rt}(G)|=1$ (resp., $|\mathrm{Rt}(G)|=2$),
where we call $G$ {\em unrooted} if $\mathrm{Rt}(G)=\emptyset$.

 For a graph $G$ possibly with roots
 a {\em leaf-vertex} is defined to be a non-root vertex 
 $v\in  V(G)\setminus \mathrm{Rt}(G)$
 with degree 1, 
 call  the edge $uv$ incident to a leaf vertex $v$ a {\em leaf-edge},
 and denote $\Vleaf(G)$ and $\Eleaf(G)$
  the sets of leaf-vertices and leaf-edges  in $G$, respectively.
 For a graph  or a rooted graph $G$,
 we define graphs $G_i, i\in \mathbb{Z}_+$ obtained from $G$
 by removing the set of leaf-vertices $i$ times so that
\[ G_0:=G; ~~ G_{i+1}:=G_i - \Vleaf(G_i), \]
where we call a vertex $v\in  \Vleaf(G_k)$ 
a {\em leaf $k$-branch} and
we say that a vertex $v\in  \Vleaf(G_k)$ has
height {\em height} $\h(v)=k$ in $G$. 
The {\em height} $\h(T)$ of a rooted tree $T$ is defined
to be the maximum of $\h(v)$ of a vertex $v\in V(T)$. 
For an integer $k\geq 0$, we call a  rooted tree $T$
 {\em $k$-lean} if $T$ has at most one leaf $k$-branch.
For an unrooted cyclic graph $G$, we regard the set of non-core-edges in $G$ induces
a collection $\mathcal{T}$ of trees each of which is rooted at a core-vertex,
where we call $G$  {\em $k$-lean} if each of the rooted trees in $\mathcal{T}$ 
is $k$-lean. 
 
\subsection{Modeling of Chemical Compounds}\label{sec:chemical_model}

To represent a chemical compound, 
we introduce a set  of   chemical elements such as 
  {\tt H} (hydrogen),   
 {\tt C} (carbon), {\tt O} (oxygen), {\tt N} (nitrogen)  and so on.
 To distinguish a chemical element $\ta$ with multiple valences such as {\tt S} (sulfur),
 we denote a chemical element $\ta$ with a valence $i$ by $\ta_{(i)}$,
 where we do not use such a suffix $(i)$ 
 for a chemical element $\ta$ with a unique valence. 
Let $\Lambda$ be a set of chemical elements $\ta_{(i)}$.
For example,  $\Lambda=\{\ttH,  \ttC, \ttO, \ttN, \ttP, \ttS_{(2)}, \ttS_{(4)}, \ttS_{(6)}\}$. 
Let $\val: \Lambda\to [1,6]$ be a valence function.
For example, $\val(\ttH)=1$, $\val(\ttC)=4$, $\val(\ttO)=2$, $\val(\ttP)=5$,
$\val(\ttS_{(2)})=2$, $\val(\ttS_{(4)})=4$ and $\val(\ttS_{(6)})=6$.
 For each  chemical element $\ta\in \Lambda$, 
let $\mathrm{mass}(\ta)$  denote the mass   of  $\ta$.  

A chemical compound  is represented by a {\em chemical graph} defined to be
a tuple $\C=(H,\alpha,\beta)$  of
  a simple, connected undirected graph $H$ and  
    functions   $\alpha:V(H)\to \Lambda$  and  $\beta: E(H)\to [1,3]$.
The set of atoms and the set of bonds in the compound 
are represented by the vertex set $V(H)$ and the edge set $E(H)$, respectively.
The chemical element assigned to a vertex $v\in V(H)$
is represented by $\alpha(v)$ and 
 the bond-multiplicity  between two adjacent vertices  $u,v\in V(H)$
is represented by $\beta(e)$ of the edge $e=uv\in E(H)$.
We say that two tuples $(H_i,\alpha_i,\beta_i), i=1,2$ are
{\em isomorphic} if they admit an isomorphism $\phi$,
i.e.,  a bijection $\phi: V(H_1)\to V(H_2)$
such that
 $uv\in E(H_1), \alpha_1(u)=\ta, \alpha_1(v)=\tb, \beta_1(uv)=m$
 $\leftrightarrow$  
 $\phi(u)\phi(v) \in E(H_2), \alpha_2(\phi(u))=\ta, 
 \alpha_2(\phi(v))=\tb, \beta_2(\phi(u)\phi(v))=m$. 
 When $H_i$ is rooted at a vertex $r_i,  i=1,2$,
 $(H_i,\alpha_i,\beta_i), i=1,2$ are
{\em rooted-isomorphic} (r-isomorphic) if 
they admit an isomorphism $\phi$ such that $\phi(r_1)=r_2$. 
 
 For a notational convenience, we  use
 a function $\beta_\C: V(H)\to [0,12]$ 
 for a chemical graph $\C=(H,\alpha,\beta)$ 
  such that $\beta_\C(u)$ means the sum of bond-multiplicities
 of edges incident to a vertex $u$; i.e., 
\[ \beta_\C(u) \triangleq \sum_{uv\in E(H) }\beta(uv) 
\mbox{ for each vertex $u\in V(H)$.}\]
For each vertex $u\in V(H)$, 
 define the {\em electron-degree} $\eledeg_\C(u)$  to be 
\[  \eledeg_\C(u) \triangleq  \beta_\C(u) - \val(\alpha(u)). \]
For each  vertex $u\in V(H)$ and each chemical element $\ta\in \Lambda$,
let $\deg_\C(v;\ta)$ denote the number of atoms with $\ta$ 
 adjacent to a vertex $v$ in $\C$. 
  
  For a chemical   graph  $\C=(H,\alpha,\beta)$, 
  let  $V_{\ta}(\C)$, $\ta\in \Lambda$
   denote the set  vertices $v\in V(H)$ such that $\alpha(v)=\ta$ in $\C$
  and define the {\em hydrogen-suppressed chemical graph} $\anC$ 
to be  the graph obtained from $H$ by
  removing all the vertices $v\in \VH(\C)$.
  

\begin{figure}[h!] \begin{center}
\includegraphics[width=.80\columnwidth]{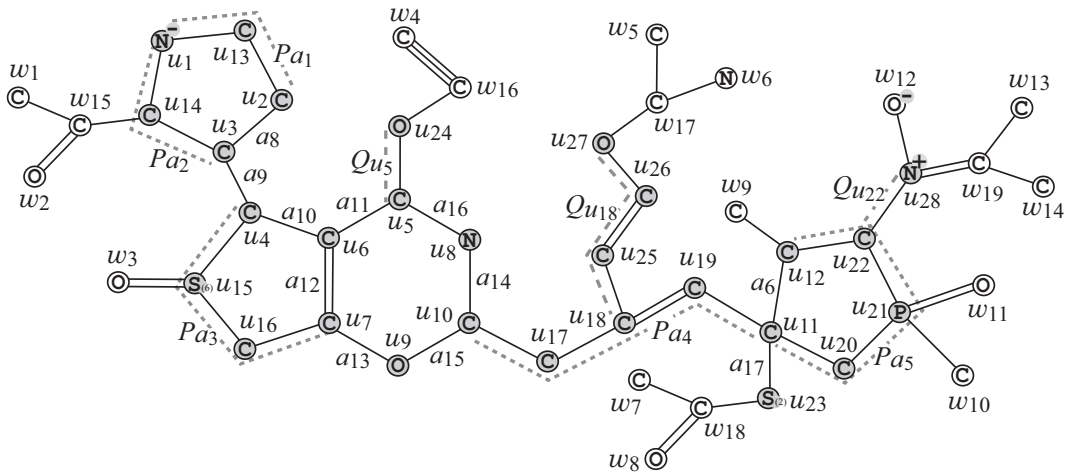}
\end{center} \caption{An illustration of  a hydrogen-suppressed chemical graph  
$\anC$ obtained from a chemical graph $\C$ with $\mathrm{r}(\C)=4$ 
by removing all the 
 hydrogens, 
where for  ${\rho}=2$,  
$V^\ex(\C)=\{w_i \mid i\in [1,19]\}$ and
$V^\inte(\C)=\{u_i \mid i\in [1,28]\}$.  
 }
\label{fig:example_chemical_graph} \end{figure}

\section{Two-layered Model}\label{sec:2LM}
This section reviews the two-layered model and 
describes our modification to the model.

 Let  $\C=(H,\alpha,\beta)$ be a chemical graph
 and  ${\rho}\geq 1$ be an integer, which we call a {\em branch-parameter}.
 
  A {\em two-layered model} of $\C$ is a partition of
 the hydrogen-suppressed chemical graph $\anC$ into
 an ``interior'' and an ``exterior'' in the following way. 
 We call a vertex $v\in V(\anC)$
   (resp., an edge $e\in E(\anC))$ of   $G$
   an {\em exterior-vertex} (resp.,    {\em exterior-edge}) if
    $\h(v)< {\rho}$ (resp., $e$ is incident to an  exterior-vertex)
and denote the sets of exterior-vertices and exterior-edges 
by $V^\ex(\C)$ and $E^\ex(\C)$, respectively
and denote  $V^\inte(\C)=V(\anC)\setminus  V^\ex(\C)$ and 
$E^\inte(\C)=E(\anC)\setminus E^\ex(\C)$, respectively.
We call a vertex in $V^\inte(\C)$ (resp.,   an edge in $E^\inte(\C)$) 
   an {\em interior-vertex} (resp.,    {\em interior-edge}). 
 The set  $E^\ex(\C)$ of  exterior-edges forms 
a collection of connected graphs each of which is
regarded as a rooted tree $T$ rooted at 
the vertex $v\in V(T)$ with the maximum $\h(v)$. 
Let $\mathcal{T}^\ex(\anC)$ denote 
the set of these chemical rooted trees in $\anC$. 
The {\em interior} $\C^\inte$ of $\C$ is defined to be the subgraph
 $(V^\inte(\C),E^\inte(\C))$ of $\anC$. 

Figure~\ref{fig:example_chemical_graph}
 illustrates an example of a hydrogen-suppressed chemical graph $\anC$.
For a branch-parameter ${\rho}=2$, 
the interior of  the chemical graph $\anC$ in Figure~\ref{fig:example_chemical_graph} 
is obtained by removing the set of vertices with degree 1 ${\rho}=2$ times; i.e., 
first remove  
the set  $V_1=\{w_1,w_2,\ldots,w_{14}\}$ of vertices of degree 1 in $\anC$ 
and then remove  the set
 $V_2=\{w_{15},w_{16},\ldots,w_{19}\}$ of vertices of degree 1 in $\anC-V_1$,
 where the removed vertices become the exterior-vertices of $\anC$.


  
\begin{figure}[h!] \begin{center}
\includegraphics[width=.84\columnwidth]{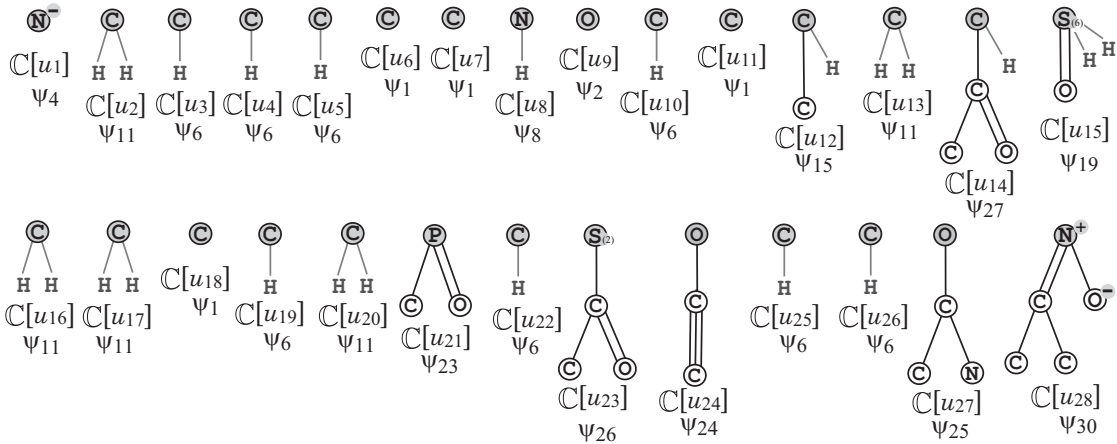}
\end{center} \caption{
The set $\mathcal{T}(\C)$ of  2-fringe-trees  $\C[u_i], i\in [1,28]$ of the example $\C$
in Figure~\ref{fig:example_chemical_graph}, 
where the root of each tree is depicted with a gray circle and
 the hydrogens attached to non-root vertices are omitted in the figure.  
 }
\label{fig:example_fringe-tree} \end{figure}

For each interior-vertex $u\in V^\inte(\C)$,
let $T_u\in \mathcal{T}^\ex(\anC)$ denote the chemical tree rooted at $u$
(where possibly $T_u$ consists of vertex $u$)
and 
define the {\em $\rho$-fringe-tree} $\C[u]$ 
to be  
the chemical rooted tree obtained from $T_u$ by putting back
 the hydrogens originally attached $T_u$ in $\C$. 
Let $\mathcal{T}(\C)$ denote the set of $\rho$-fringe-trees 
$\C[u], u \in V^\inte(\C)$. 
Figure~\ref{fig:example_fringe-tree}  illustrates
the set  $\mathcal{T}(\C)=\{\C[u_i]\mid i\in [1,28]\}$ of the 2-fringe-trees 
  of the example $\C$
in Figure~\ref{fig:example_chemical_graph}. 

\smallskip
\noindent {\bf Feature Function~} 
 The feature of an  interior-edge $e=uv\in E^\inte(\C)$ 
 such that $\alpha(u)=\ta$, $\deg_{\anC}(u)=d$, 
 $\alpha(v)=\tb$, $\deg_{\anC}(v)=d'$  and $\beta(e)=m$  is represented by 
 a tuple $(\ta d, \tb d', m)$, which is called the {\em edge-configuration} 
  of the edge $e$, where 
  we call the tuple $(\ta, \tb, m)$ 
 the {\em adjacency-configuration} of the edge $e$. 
 
For an integer $K$, a feature vector $f(\C)$ of a chemical graph $\C$
is defined by a {\em feature function} $f$ that consists of $K$ descriptors. 
We call  $\RK$ {\em  the feature space}.

 Tanaka et~al.~\cite{TZAHZNA21} 
  defined  a feature vector $f(\C)\in \RK$  
to be a combination of the frequency 
of edge-configurations of   the interior-edges  and
the frequency of chemical rooted trees among the set 
of  chemical rooted trees $\C[u]$ over all interior-vertices $u$. 
In this paper, we introduce the rank and the adjacency-configuration of leaf-edges
as new descriptors  in a feature vector of a chemical graph. 
 
\smallskip
\noindent {\bf Topological Specification~}   
A topological specification is described
as a set of the following rules proposed by \cite{SZAHZNA21}
and modified by Tanaka et~al.~\cite{TZAHZNA21}:
\begin{enumerate}[nosep]
\item[(i)]
a {\em seed graph} $\GC$ as an  abstract form of  a target chemical graph $\C$;
\item[(ii)]
 a set $\mathcal{F}$ of chemical rooted trees  as candidates
 for a tree  $\C[u]$ rooted at each interior-vertex $u$ in $\C$; 
and 
\item[(iii)]
lower and upper bounds on the number of components 
 in a target chemical graph such as  chemical elements, 
double/triple bonds and the interior-vertices in $\C$. 
\end{enumerate} 

\begin{figure}[h!] \begin{center}
\includegraphics[width=.98\columnwidth]{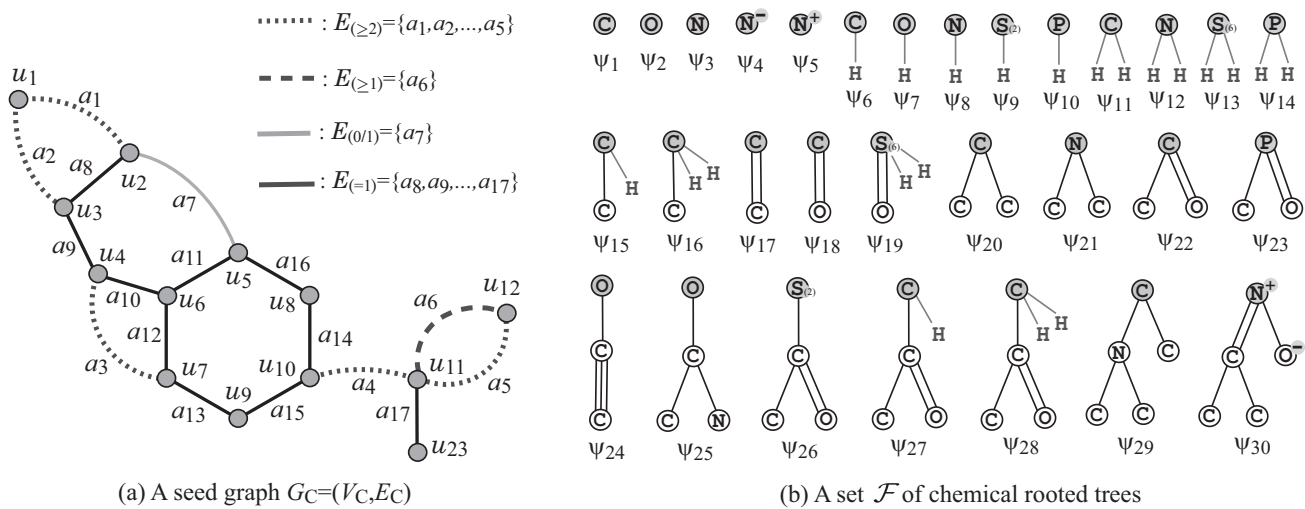}
\end{center} \caption{
(a) An illustration of a seed graph $\GC$ with $\mathrm{r}(\GC)=5$
where the vertices in $\VC$ are depicted with gray circles,
the edges in $\Et$ are depicted with dotted lines,
the edges in $\Ew$ are depicted with dashed lines,
the edges in $\Ez$ are depicted with gray bold lines and  
the edges in $\Eew$ are depicted with black solid lines;
(b) A set $\mathcal{F}=\{\psi_1,\psi_2,\ldots,\psi_{30}\}\subseteq
\mathcal{F}(D_\pi)$ of 30 chemical rooted trees
$\psi_i, i\in [1,30]$, where the root of each tree is depicted with a gray circle, 
where  the hydrogens attached to non-root vertices are omitted in the figure.    }
\label{fig:specification_example_1} \end{figure}  

Figure~\ref{fig:specification_example_1}(a) and (b)
 illustrate  examples of  a  seed graph  $\GC$ and 
 a set $\mathcal{F}$ of chemical rooted trees, respectively. 
 Given a seed graph $\GC$, 
 the interior of   a target chemical graph $\C$ is constructed
 from $\GC$ by replacing some edges $a=uv$ 
 with paths $P_a$ between the end-vertices
 $u$ and $v$ and by attaching new paths $Q_v$ to some vertices $v$.  
%
For example, a chemical graph $C$ 
in Figure~\ref{fig:example_chemical_graph} is constructed
from the seed  graph  $\GC$ in Figure~\ref{fig:specification_example_1}(a)
as follows.
\begin{enumerate}[nosep,  leftmargin=*]
\item[-]
First replace  five edges
 $a_1=u_1 u_{2},  a_2=u_1 u_{3},  a_3=u_4 u_{7}, a_4=u_{10}u_{11}$
and $a_5=u_{11}u_{12}$ in  $\GC$ 
 with new paths  
$P_{a_1}=(u_1,u_{13},u_{2})$, 
$P_{a_2}=(u_{1},u_{14},u_{3})$,
$P_{a_3}=(u_{4},u_{15},u_{16},u_{7})$, 
 $P_{a_4}=(u_{10},u_{17},u_{18},u_{19},u_{11})$ and
 $P_{a_5}=(u_{11},u_{20},u_{21},u_{22},u_{12})$, respectively
 to obtain a subgraph $G_1$ of $\anC$. 
\item[-]
Next attach to this graph  $G_1$ three new paths 
$Q_{u_5}=(u_5,u_{24})$, 
$Q_{u_{18}}=(u_{18},u_{25},u_{26},u_{27})$ and 
$Q_{u_{22}}=(u_{22},u_{28})$
to obtain  
the interior of  $\anC$ in Figure~\ref{fig:example_chemical_graph}.
\item[-]
Finally  attach to the interior   28 trees selected from the set $\mathcal{F}$ 
and assign chemical elements and bond-multiplicities in the interior
to  obtain a chemical graph $\C$  in Figure~\ref{fig:example_chemical_graph}. 
In Figure~\ref{fig:example_fringe-tree},  XXX Check the next again XXXX
  $\psi_1\in \mathcal{F}$ is selected for $\C[u_i]$, $i\in\{6,7,11\}$.
  Similarly 
  $\psi_2$  for  $\C[u_9]$,
  $\psi_4$   for $\C[u_i]$, $i\in\{3,4,5,10,18,19,22,25,26\}$,
  $\psi_5$    for $\C[u_i]$, $i\in\{2,13,16,17,20\}$,
  $\psi_7$  for $\C[u_i]$, $i\in\{1,8\}$,
  $\psi_9$   for $\C[u_{12}]$,
   $\psi_{11}$    for $\C[u_{15}]$,
   $\psi_{13}$    for $\C[u_{21}]$,
   $\psi_{16}$    for $\C[u_{24}]$,
   $\psi_{17}$    for $\C[u_{27}]$, 
   $\psi_{18}$  for $\C[u_{14}]$,
   $\psi_{19}$   for $\C[u_{23}]$   
   and 
    $\psi_{21}$  for $\C[u_{28}]$. 
\end{enumerate} 

%
%

Our definition of a topological specification is analogous with the one  by 
 Tanaka et~al.~\cite{TZAHZNA21} 
   except for a necessary modification due to the introduction 
   of multiple valences of chemical elements, cations and anions 
(see Appendix~\ref{sec:specification} for a full description of topological specification).

\section{Linear Regressions}\label{sec:linear_regression} 

For an integer $p\geq 1$ and a vector $\x\in \R^p$, the $j$-th entry of $\x$ 
is denoted by $\x(j), j\in [1,p]$.

Let $D$ be a data set   of chemical graphs $\C$ with
an observed value $a(\C)\in \R$,
where we denote by $a_i=a(\C_i)$ 
for an indexed graph $\C_i$. 

Let $ f$ be a feature function that maps a chemical graph $\C$
to a vector $ f(\C)\in \RK$
where we denote by $\x_i= f(\C_i)$ 
for an indexed graph $\C_i$. 
For  a prediction function $\eta: \RK\to \R$, 
define an error function 
\[ \mathrm{Err}(\eta;D)  \triangleq 
\sum_{\C_i\in D}(a_i - \eta(f(\C_i)))^2=\sum_{\C_i\in D}(a_i - \eta(\x_i))^2, \]
and define the {\em coefficient of determination}
 $\mathrm{R}^2(\eta,D)$ 
  to be 
\[ \displaystyle{ \mathrm{R}^2(\eta,D)\triangleq 
  1- \frac{\mathrm{Err}(\eta;D) } 
  {\sum_{ \C_i\in D  } (a_i-\widetilde{a})^2}  
  \mbox{   for  }
   \widetilde{a}= \frac{1}{|D |}\sum_{ \C\in D }a(\C).  } \]

For a feature space $\RK$, a hyperplane is defined to be 
a pair  $(\w,b)$ of a vector $\w\in \RK$ and a real $b\in \R$.
Given a hyperplane $(\w,b)\in \RKw$,
a prediction function $\eta_{\w,b}:\RK\to \R$ is defined by setting
\[ \eta_{\w,b}(\x) \triangleq \w\cdot \x +b=\sum_{j\in [1,K]}\w(j)\x(j) +b. \]
We can observe that such a prediction function can be represented as an ANN
with an input layer with $K$ nodes $u_{j}, j\in [1,K]$
and an output layer with a single node $v$
such that the weight of edge arc $(u_{j},v)$ is set to be $w(j)$, 
 the bias of node $u$ is set to be $b$ 
 and the activation function at node $u$ is set to be a linear function.
 However, a learning algorithm for an ANN may not find a set of weights
 $w(j),j\in[1,K]$ and $b$ that minimizes the error function, since
 the algorithm simply iterates modification of the current weights and biases
 until it terminates at a local optima in the minimization. 

We wish to find a hyperplane $(\w,b)$  that minimizes the error function
$\mathrm{Err}(\eta_{\w,b};D)$.
In many cases, a feature vector $f$ contains descriptors that do not play
an essential role in constructing a good prediction function.
When we solve the minimization problem, the entries $\w(j)$ for some descriptors $j\in [1,K]$
in the resulting hyperplane $(\w,b)$ become zero, which means that these descriptors were
not necessarily important for finding a prediction function $\eta_{\w,b}$. 
It is proposed that solving the minimization with an additional penalty term $\tau$ to the error function 
often results in a more number of  entries $\w(j)=0$, reducing a set of descriptors
necessary for defining   a prediction function $\eta_{\w,b}$. 
For an error function with such a penalty term,   
a Ridge function  $\frac{1}{2|D|}\mathrm{Err}(\eta_{\w,b};D)
    + \lambda  [\sum_{j\in[1,K]}w(j)^2 +  b^2]$~\cite{Ridge88}
 and  a Lasso function 
$\frac{1}{2|D|}\mathrm{Err}(\eta_{\w,b};D)
+   \lambda [\sum_{j\in[1,K]}  |w(j)| +  |b| ]$~\cite{Lasso96}
are known, where $\lambda \in \R$ is a given real number.

Given a prediction function $\eta_{\w,b}$,
we can simulate a  process of computing the output $\eta_{\w,b}(\x)$ for an input $\x\in \RK$
as an MILP   $\mathcal{M}(x,y;\mathcal{C}_1)$ in the framework.  
By solving such an MILP for a specified target value $y^*$ ,   
we can find a vector $x^*\in \RK$ such that $\eta_{\w,b}(\x^*)=y^*$.
Instead of specifying a single target value $y^*$, we use
  lower and upper bounds $\underline{y}^*, \overline{y}^*\in \R$ 
  on the value $a(\C)$ of a chemical graph $\C$ to be inferred.
We can control the range between $\underline{y}^*$ and $\overline{y}^*$
for searching a chemical graph $\C$  
by setting  $\underline{y}^*$ and $\overline{y}^*$ to be close or different values.  
A desired MILP is formulated as follows. 

\paragraph{$\mathcal{M}(x,y;\mathcal{C}_1)$: 
 An MILP formulation  for the inverse problem to prediction function }
~\\

\smallskip\noindent
{\bf constants: } 
\begin{enumerate}  [nosep,  leftmargin=*]
\item[-]  
A hyperplane $(\w,b)$ with $\w\in \RK$ and $b\in \R$;

  
\item[-]
Real values $\underline{y}^*, \overline{y}^*\in \R$ such that $\underline{y}^*< \overline{y}^*$; 

\item[-] 
A set $I_{\Z}$ of indices $j\in [1,K]$ such that 
the $j$-th descriptor $\dcp_j(\C)$ is always an integer;

\item[-] 
A set $I_{+}$ of indices $j\in [1,K]$ such that 
the $j$-th descriptor $\dcp_j(\C)$ is always non-negative;

\item[-] 
$\ell(j), u(j)\in \R, j\in [1,K]$: lower and upper bounds on the $j$-th descriptor;
 \end{enumerate}

\smallskip\noindent
{\bf variables: } 
\begin{enumerate} [nosep,  leftmargin=*]

\item[-] 
Non-negative integer variable $x(j)\in \Z_+, j\in I_{\Z}\cap I_+$;

\item[-] 
Integer variable $x(j)\in \Z, j\in I_{\Z}\setminus I_+$;

\item[-] 
Non-negative real variable $x(j)\in \Z_+, j\in  I_+\setminus I_{\Z}$;

\item[-] 
Real variable $x(j)\in \Z, j\in [1,K] \setminus (I_{\Z}\cup I_+)$;
 \end{enumerate}
 
\smallskip\noindent
{\bf constraints: }   
\begin{align} 
 \ell(j) \leq x(j)  \leq  u(j), j\in [1,K], \label{eq:1} 
\end{align}   
\begin{align} 
\underline{y}^* \leq 
\sum_{j\in [1,K]} w(j)x(j)+b \leq  \overline{y}^*,  \label{eq:2} 
\end{align}   
 
\smallskip\noindent
{\bf objective function: }   \\
~~ none.\\

The number of variables and constraints in the above MILP formulation is $O(K)$. 
It is not difficult to see that the above MILP is an NP-hard problem. 

The entire MILP for Stage~4 consists of the two MILPs
$\mathcal{M}(x,y;\mathcal{C}_1)$ and $\mathcal{M}(g,x;\mathcal{C}_2)$ 
with no objective function. 
The latter  represents the computation process of our feature function $f$ and 
a given topological specification.   
See Appendix~\ref{sec:full_milp}  for the details of  MILP $\mathcal{M}(g,x;\mathcal{C}_2)$.

\clearpage

\section{Results}\label{sec:experiment}

We implemented our method of Stages~1 to 5 
for inferring chemical graphs under a given topological specification and
conducted experiments  to evaluate the computational efficiency. 
We executed the experiments on a PC with 
 Processor:  Core i7-9700 (3.0 GHz; 4.7 GHz at the maximum) and 
Memory: 16 GB RAM DDR4. 

\medskip \noindent
{\bf Results on Phase~1.  }
We have conducted experiments of linear regression for 
37 chemical properties   among which
we report the following 18 properties  to which the test  
 coefficient of determination ${\rm R}^2$ 
attains at least 0.8:
octanol/water partition coefficient ({\sc Kow}), 
heat of combustion ({\sc Hc}), 
vapor density ({\sc Vd}), 
optical rotation ({\sc OptR}), 
electron density on the most positive atom ({\sc EDPA}), 
melting point ({\sc Mp}), 
heat of atomization ({\sc Ha}), 
heat of formation ({\sc Hf}), 
internal energy at 0K ({\sc U0}), 
energy of lowest unoccupied molecular orbital ({\sc Lumo}), 
isotropic polarizability ({\sc Alpha}), 
heat capacity at 298.15K ({\sc Cv}), 
solubility ({\sc Sl}), 
surface tension ({\sc SfT}),  
viscosity ({\sc Vis}),
isobaric heat capacities in liquid phase ({\sc IhcLiq}), 
isobaric heat capacities in solid phase ({\sc IhcSol}) and 
lipophilicity ({\sc Lp}).

We used data sets  provided  by HSDB from PubChem~\cite{pubchem} 
 for {\sc Kow},  {\sc Hc},  {\sc Vd} and  {\sc OptR}, 
 M.~Jalali-Heravi and M.~Fatemi~\cite{JF01} for {\sc EDPA},  
   Roy and Saha~\cite{RS03} for {\sc Mp}, {\sc Ha} and {\sc Hf}, 
 MoleculeNet~\cite{moleculenet} for {\sc U0},  {\sc Lumo},  {\sc Alpha}, {\sc Cv} and {\sc Sl},   
 Goussard et al.~\cite{GFPDNA20} for {\sc SfT} and  {\sc Vis}, 
R.~Naef~\cite{Naef}  for {\sc IhcLiq} and  {\sc IhcSol},  and 
 Figshare~\cite{figshare}  for {\sc Lp}.

Properties  {\sc  U0, Lumo, Alpha} and  {\sc Cv} share a common original data set $D^*$
with  more than 130,000 compounds, and 
we used  a set $D_\pi$ of 1,000 graphs randomly selected from $D^*$  
as a common data set of these four properties $\pi$
in this experiment.

We implemented Stages~1, 2 and 3 in Phase~1 as follows.

\medskip \noindent
{\bf Stage~1.  }
We set  a graph class   $ \mathcal{G}$   to be
the set of all chemical graphs with any graph structure, 
and set a branch-parameter ${\rho}$ to be 2. 

For each of the properties,  
 we first select a set $\Lambda$ of chemical elements 
 and then collect  a  data set  $D_{\pi}$ on chemical graphs
 over the set $\Lambda$ of chemical elements.  
 To construct the data set $D_{\pi}$,
  we eliminated  chemical compounds that do not satisfy 
  one of the following: the graph is connected,
  the number of carbon atoms is at least four,
  and   the number of non-hydrogen neighbors of each atom is
  at most 4.   
  
 Table~\ref{table:phase1a}    shows
  the size and range of data sets   that 
 we prepared for each chemical property in Stage~1,
 where  we denote the following:  
\begin{enumerate}[nosep,  leftmargin=*]
\item[-] 
  $\Lambda$: the set of elements used in the data set $D_{\pi}$; 
  $\Lambda$ is one of the following 11 sets: 
  $\Lambda_1=\{\ttH,\ttC,\ttO \}$; 
   $\Lambda_2=\{\ttH,\ttC,\ttO, \ttN \}$;
   $\Lambda_3=\{\ttH,\ttC,\ttO, \ttS_{(2)} \}$;
    $\Lambda_4=\{\ttH,\ttC,\ttO, \ttSi \}$; \\
   $\Lambda_5=\{\ttH,\ttC,\ttO, \ttN,\ttCl,\ttP_{(3)},\ttP_{(5)} \}$;
    $\Lambda_6=\{\ttH,\ttC,\ttO, \ttN,\ttS_{(2)},\ttF \}$;
   $\Lambda_7=\{\ttH,\ttC,\ttO, \ttN,\ttS_{(2)},\ttS_{(6)},\ttCl \}$;
   $\Lambda_8=\{\ttH, \ttC_{(2)},\ttC_{(3)},\ttC_{(4)},\ttO, \ttN_{(2)}, \ttN_{(3)} \}$; 
   $\Lambda_9=\{\ttH,\ttC,\ttO, \ttN, \ttS_{(2)},\ttS_{(4)},\ttS_{(6)},\ttCl\}$; \\
   $\Lambda_{10}=\{\ttH, \ttC_{(2)},\ttC_{(3)},\ttC_{(4)},\ttC_{(5)},\ttO,
   \ttN_{(1)}, \ttN_{(2)}, \ttN_{(3)}, \ttF \}$; and \\
    $\Lambda_{11}=\{\ttH, \ttC_{(2)},\ttC_{(3)},\ttC_{(4)},\ttO, \ttN_{(2)}, \ttN_{(3)},
     \ttS_{(2)},\ttS_{(4)},\ttS_{(6)},\ttCl \}$,
 where ${\tt e}_{(i)}$ for a chemical element ${\tt e}$ and an integer $i\geq 1$ 
 means that  a chemical element ${\tt e}$ with valence $i$. 

\item[-] 
 $|D_{\pi}|$:  the size of data set $D_{\pi}$ over $\Lambda$
  for the property $\pi$.
  
\item[-]   $ \underline{n},~\overline{n} $:  
  the minimum and maximum  values of the number 
  $n(\Co)$ of non-hydrogen atoms in 
  compounds $\Co$ in $D_{\pi}$.
\item[-] $ \underline{a},~\overline{a} $:  the minimum and maximum values
of $a(\Co)$ for $\pi$ over  compounds $\Co$ in  $D_{\pi}$.
\item[-]    $|\Gamma|$: 
the number of different edge-configurations
of interior-edges over the compounds in~$D_{\pi}$. 
\item[-]  $|\mathcal{F}|$: the number of non-isomorphic chemical rooted trees
 in the set of all 2-fringe-trees in  the   compounds in $D_{\pi}$.
 
\item[-] $K$: the number of descriptors in a feature vector $f(\Co)$.  
\end{enumerate}

\medskip \noindent
{\bf Stage~2.  }
We used  the  new  feature function  defined 
in our chemical model without suppressing hydrogen 
(see Appendix~\ref{sec:descriptor}  for the detail).
We standardize the range of each descriptor and
 the range $\{t\in \R \mid \underline{a}\leq t\leq \overline{a}\}$ 
 of property values   $a(\Co), \Co\in D_\pi$.

\medskip \noindent
{\bf Stage~3.  }
For each chemical property $\pi$, we select a penalty value $\lambda_\pi$
in the Lasso function 
from  36 different values from 0 to 100
by conducting linear regression as a preliminary experiment.
 
We conducted an experiment in Stage~3 to evaluate the performance
of the prediction function based on cross-validation.
For a property $\pi$, 
an execution of a {\em cross-validation}  consists of five trials of
constructing a prediction function as follows.
First partition the data set $D_{\pi}$ 
 into five subsets $D^{(k)}$, $k\in[1,5]$ randomly.
 For each $k\in[1,5]$,  the $k$-th trial 
  constructs a prediction function $\eta^{(k)}$  by conducting 
  a linear regression with the penalty term $\lambda_\pi$
     using the set $D_{\pi}\setminus D^{(k)}$ as a training data set.
We used   scikit-learn version 0.23.2  with Python 3.8.5 for executing linear regression 
with Lasso function.
For each property, we executed ten cross-validations and
we show the median of test $\mathrm{R}^2(\eta^{(k)},D^{(k)}), k\in[1,5]$
over all ten cross-validations.
Recall that a subset of descriptors is selected in linear regression with Lasso function
and let $K'$ denote the average number of selected descriptors over all 50 trials.  
The running time per trial in a cross-validation was at most one second. 
  
\begin{table}[h!]\caption{Results in Phase 1.} 
  \begin{center}
    \begin{tabular}{@{} c c r c  c  r r r c r  c   @{}}\toprule
      $\pi$ & $\Lambda$  &  $|D_{\pi}|$  &  $ \underline{n},~\overline{n} $ &   $\underline{a},~\overline{a}$ &
              $|\Gamma|$   &  $|\mathcal{F}|$ &   $K$ &  
              $\lambda_\pi$ & $K'$~ &    test $\mathrm{R}^2$  \\ \midrule
      {\sc Kow} & $\Lambda_2$  & 684  &  4,\,58  &    -7.5,\,15.6  & 25   & 166  & 223  & $6.4\mathrm{E}{-5}$  & 80.3 & 0.953   \\
      {\sc Kow} & $\Lambda_9$  & 899 & 4,\,69 &   -7.5,\,15.6 & 37   & 219    & 303   & $5.5\mathrm{E}{-5}$ &112.1 & 0.927    \\ 
      {\sc Hc} & $\Lambda_2$ & 255  &  4,\,63  &   49.6,\,35099.6&  17  & 106     & 154 &   $1.9\mathrm{E}{-4}$ & 19.2 & 0.946    \\
      {\sc Hc} & $\Lambda_7$  & 282  &  4,\,63 & 49.6,\,35099.6 & 21  & 118    & 177 &  $1.9\mathrm{E}{-4}$& 20.5 & 0.951    \\
      {\sc Vd} & $\Lambda_2$ & 474  &  4,\,30  & 0.7,\,20.6   &  21  &160  & 214 &  $1.0\mathrm{E}{-3}$ & 3.6 & 0.927    \\
      {\sc Vd} & $\Lambda_5$  & 551   & 4,\,30  & 0.7,\,20.6 & 24 &191 & 256 &  $5.5\mathrm{E}{-4}$& 8.0 &0.942    \\
      {\sc OptR} & $\Lambda_2$  & 147  &  5,\,44 & -117.0,\,165.0 & 21  & 55   &107 & $4.6\mathrm{E}{-4}$  & 39.2  & 0.823    \\
      {\sc OptR} &   $\Lambda_6$  & 157  &  5,\,69  & -117.0,\,165.0 & 25  & 62    &123 & $7.3\mathrm{E}{-4}$  & 41.7  & 0.825    \\
      {\sc EDPA} &  $\Lambda_1$ &  52  &   11,\,16  &  0.80,\,3.76 & 9   & 33   & 64 & $1.0\mathrm{E}{-4}$ & 10.9 &  0.999    \\
      {\sc Mp} & $\Lambda_2$ & 467 &4,\,122 &   -185.33,\,300.0  & 23 &142   & 197 & $3.7\mathrm{E}{-5}$ & 82.5 & 0.817   \\
      {\sc Ha} & $\Lambda_3$ & 115  &  4,\,11  & 1100.6,\,3009.6 &  8  & 83   &115 & $3.7\mathrm{E}{-5}$ & 39.0 &  0.997     \\
      {\sc Hf} & $\Lambda_1$ &  82  &  4,\,16  & 30.2,\,94.8  & 5 &    50  & 74 &  $1.0\mathrm{E}{-4}$ & 34.0 & 0.987    \\
      {\sc U0} & $\Lambda_{10}$ & 977 &  4,\,9  &   -570.6,\,-272.8   & 59  &190   & 297 & $1.0\mathrm{E}{-7}$ & 246.7 & 0.999   \\ 
      {\sc Lumo} &  $\Lambda_{10}$  & 977 &  4,\,9  &  -0.11,\,0.10    & 59  &190  & 297 &  $6.4\mathrm{E}{-5}$& 133.9 & 0.841  \\ 
      {\sc Alpha} & $\Lambda_{10}$ &  977    &   4,\,9    & 50.9,\,99.6 & 59 & 190  & 297  & $1.0\mathrm{E}{-5}$& 125.5 & 0.961  \\ 
      {\sc Cv} &  $\Lambda_{10}$ &  977    & 4,\,9  &    19.2,\,44.0 & 59  & 190    & 297 & $1.0\mathrm{E}{-5}$& 165.3 &0.961  \\ 
      {\sc Sl} & $\Lambda_9$ & 915 &4,\,55 &   -11.6,\,1.11  &42 & 207  & 300 & $7.3\mathrm{E}{-5}$ &130.6 &0.808   \\
      {\sc SfT}&  $\Lambda_4$  & 247  &  5,\,33 & 12.3,\,45.1 & 11  & 91  &128 & $6.4\mathrm{E}{-4}$  & 20.9  & 0.804    \\
      {\sc Vis} &  $\Lambda_4$  & 282 &  5,\,36   & -0.64,\,1.63   &  12 & 88     & 126 &   $8.2\mathrm{E}{-4}$ & 16.3 &0.893 \\
      {\sc IhcLiq} & $\Lambda_2$ & 770 &  4,\,78     &  106.3,\,1956.1    & 23 & 200     & 256 & $1.9\mathrm{E}{-5}$& 82.2 &0.987     \\
      {\sc IhcLiq} & $\Lambda_7$ & 865 &  4,\,78   &   106.3,\,1956.1  &  29 & 246    & 316 & $8.2\mathrm{E}{-6}$& 139.1 &0.986     \\
      {\sc IhcSol} & $\Lambda_8$  & 581 &   5,\,70    &    67.4,\,1220.9   & 33  & 124  & 192 & $2.8\mathrm{E}{-5}$ & 75.9 &0.985   \\
      {\sc IhcSol} & $\Lambda_{11}$ & 668  &  5,\,70   &   67.4,\,1220.9 &  40  & 140  & 228 & $2.8\mathrm{E}{-5}$ & 86.7 &0.982     \\
      {\sc Lp} & $\Lambda_2$ & 615 & 6,\,60 &  -3.62,\,6.84 & 32 &116  &  186 & $1.0\mathrm{E}{-4}$ & 98.5 &0.856     \\
      {\sc Lp} & $\Lambda_9$  & 936  &6,\,74 &   -3.62,\,6.84 &44  &136   & 231 & $6.4\mathrm{E}{-5}$ &130.4 &0.840   \\
      \bottomrule
  \end{tabular}\end{center}\label{table:phase1a}
\end{table}

 Table~\ref{table:phase1a}   shows the results on Stages~2 and 3,
 where  we denote the following:     
\begin{enumerate}[nosep,  leftmargin=*]
\item[-] 
  $\lambda_\pi$: the penalty value in the Lasso function selected
for a property $\pi$, where $a\mathrm{E}{b}$ means $a\times 10 ^{b}$.

\item[-] 
 $K'$: the average of the number of descriptors selected in the linear regression
  over all 50 trials   in ten cross-validations.
  
\item[-]
test $\mathrm{R}^2$: the median of test $\mathrm{R}^2$ over all 50 trials
  in ten cross-validations.  
\end{enumerate}
  
Recall that the adjacency-configuration for leaf-edges was introduced
as a new descriptor in this paper.
Without including this new descriptor,
the test $\mathrm{R}^2$ for property {\sc   Vis} was  0.790,
 that for {\sc  Lumo}  was 0.799  and that for {\sc Mp} was 0.796,  
 while the  test $\mathrm{R}^2$ for each of the other properties in Table~\ref{table:phase1a}
 was almost the same.
 
\begin{figure}[!ht]  \begin{center}
\includegraphics[width=.42\columnwidth]{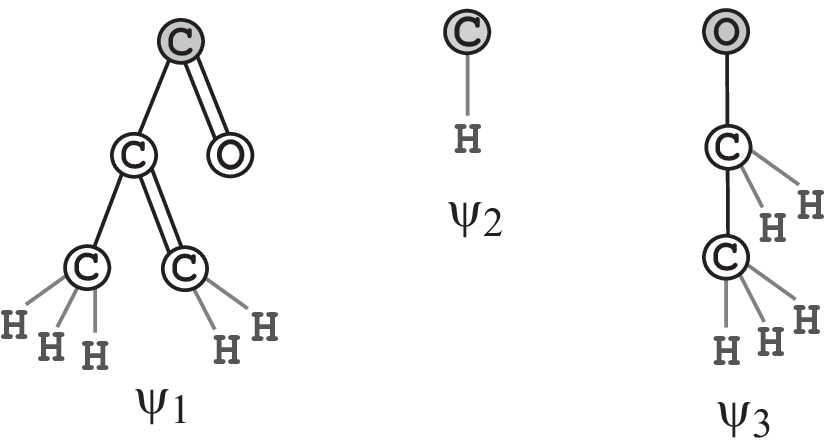}
\end{center} \caption{An illustration of chemical rooted trees
$\psi_1$, $\psi_1$ and $\psi_3$ 
that are selected in Lasso linear regression for constructing
a prediction function to property  {\sc Vd}, where the root is depicted with a gray circle.   } 
\label{fig:Vd_fringe-tree}  \end{figure}    

 From Table~\ref{table:phase1a}, we observe that a relatively large number of properties
 admit a good prediction function based on linear regression. 
The number $K'$ of descriptors used in linear regression is considerably small
for some properties.
For example of property {\sc Vd}, 
\begin{enumerate}
\item[] 
the four descriptors most frequently selected 
in the case of $\Lambda=\{\ttH,\ttO,\ttC,\ttN\}$  are 
the number of non-hydrogen atoms; 
the number of interior-vertices $v$ with $\deg_{\Co^\inte}(v)=1$; 
the number of fringe-trees r-isomorphic to the chemical rooted tree $\psi_1$
in Figure~\ref{fig:Vd_fringe-tree}; and
the number of leaf-edges with adjacency-configuration $(\ttO,\ttC,2)$. 
\item[]
the eight descriptors most frequently selected in the case of
 $\Lambda=\{\ttH,\ttO,\ttC,\ttN,\ttCl,\ttP_{(3)}, \ttP_{(5)}\}$
 are 
the number of non-hydrogen atoms; 
the number of interior-vertices $v$ with $\deg_{\Co^\inte}(v)=1$; 
the number of exterior-vertices $v$ with $\alpha(v)=\ttCl$; 
the number of interior-edges with edge-configuration $\gamma_i, i=1,2$,
where $\gamma_1=(\ttC2,\ttC2,2)$ and  $\gamma_2=(\ttC3,\ttC4,1)$; and
the number of fringe-trees r-isomorphic to the chemical rooted tree $\psi_i, i=1,2,3$
in Figure~\ref{fig:Vd_fringe-tree}.
\end{enumerate}

\clearpage

\begin{figure}[h!] \begin{center}
\includegraphics[width=.85\columnwidth]{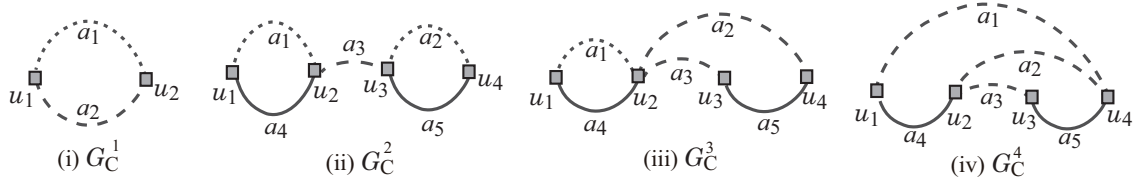}
\end{center} \caption{
(i)  Seed graph   $\GC^1$ for $I_{\mathrm{b}}^1$ and  $I_{\mathrm{d}}$;
(ii) Seed graph $\GC^2$   for $I_{\mathrm{b}}^2$; 
(iii) Seed graph  $\GC^3$   for $I_{\mathrm{b}}^3$; 
(iv)~Seed graph $\GC^4$  for $I_{\mathrm{b}}^4$. }
\label{fig:specification_example_polymer}
\end{figure}

\medskip \noindent
{\bf Results on Phase~2.  }
To execute  Stages~4  and 5 in Phase~2, 
we used a set of seven instances
$I_{\mathrm{a}}$, $I_{\mathrm{b}}^i, i\in[1,4]$, $I_{\mathrm{c}}$
 and $I_{\mathrm{d}}$ based on seed graphs prepared by  Shi et al.~\cite{SZAHZNA21}.
We here present their seed graphs $\GC$ 
(see Appendix~\ref{sec:specification} for the details of $I_{\mathrm{a}}$
and Appendix~\ref{sec:test_instances}  for the details of 
$I_{\mathrm{b}}^i, i\in[1,4]$, $I_{\mathrm{c}}$  and $I_{\mathrm{d}}$).
The seed graph  $\GC$ of instance $I_{\mathrm{a}}$ is given
 by the graph in Figure~\ref{fig:specification_example_1}(a).
The seed graph $\GC^1$ (resp., $\GC^i, i=2,3,4$)  of instances 
$I_{\mathrm{b}}^1$ and $I_{\mathrm{d}}$ (resp., $I_{\mathrm{b}}^i,  i=2,3,4$) is illustrated
 in~Figure~\ref{fig:specification_example_polymer}.
 
\begin{figure}[!htb]
\begin{center} 
 \includegraphics[width=.69\columnwidth]{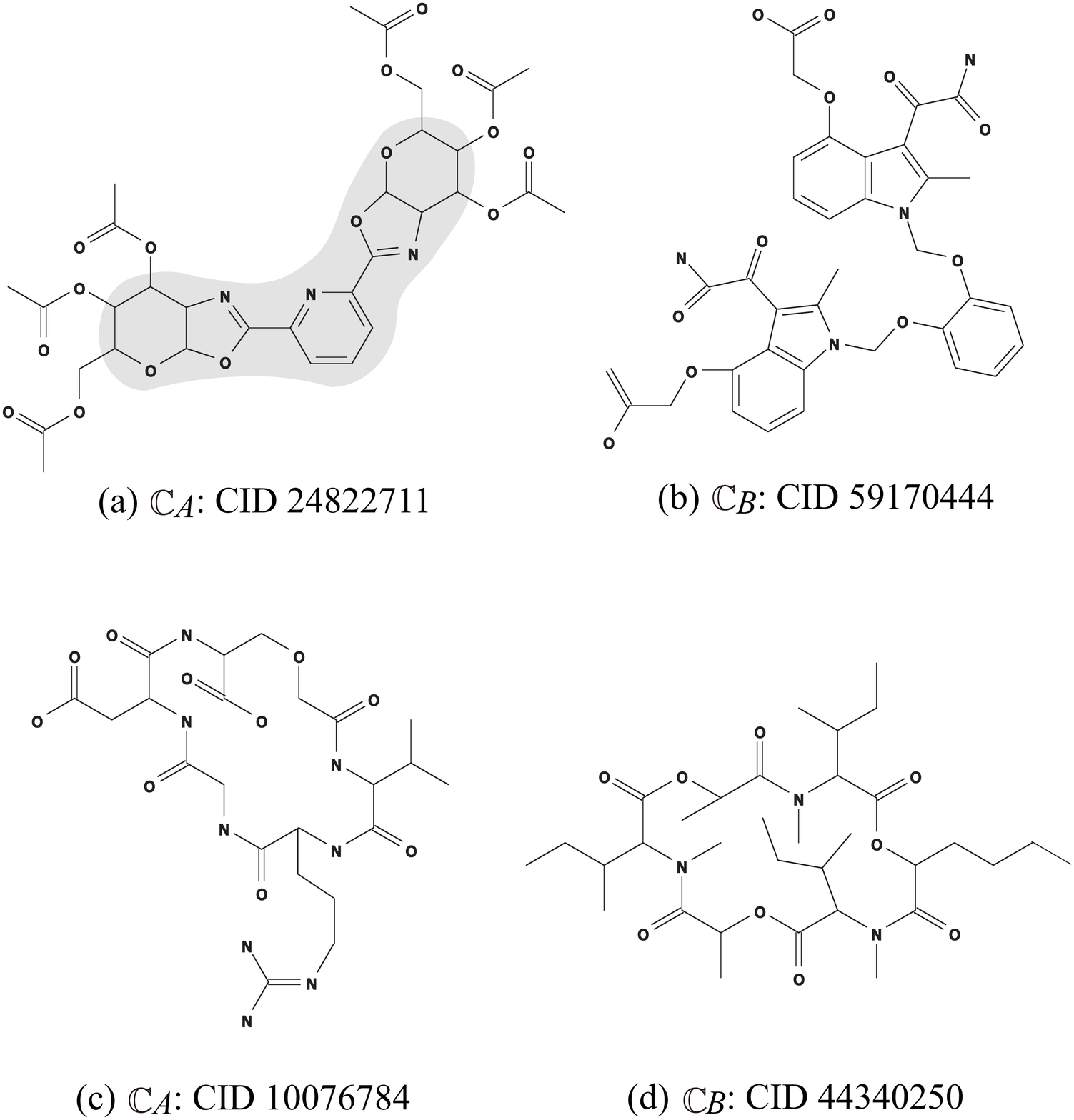}
\end{center}
\caption{An illustration of  chemical compounds 
  for instances  $I_{\rm c}$  and  $I_{\rm d}$: 
(a) $\Co_A$: CID~24822711;
(b)  $\Co_B$: CID~59170444; 
(c) $\Co_A$: CID~10076784;
(d)  $\Co_B$: CID~44340250,
where hydrogens are omitted. 
}
\label{fig:instance_I_c_I_d}  
\end{figure}

Instance  $I_{\mathrm{c}}$ has been introduced 
in order to infer a chemical graph $\Co^\dagger$ such that
the core of $\Co^\dagger$ is equal to the core of 
chemical graph $\Co_A$: CID~24822711 in Figure~\ref{fig:instance_I_c_I_d}(a)
and 
the frequency of each edge-configuration in the non-core of $\Co^\dagger$
is equal to that of chemical graph  $\Co_B$:  CID~59170444
 in  Figure~\ref{fig:instance_I_c_I_d}(b).
This means that the seed graph  $\GC$ of   $I_{\mathrm{c}}$
 is the core of $\Co_A$
which is indicated by a shaded area in  Figure~\ref{fig:instance_I_c_I_d}(a). 

Instance  $I_{\mathrm{d}}$ has been introduced 
in order to   infer a chemical monocyclic graph $\Co^\dagger$ such that
the frequency vector of  edge-configurations in  $\Co^\dagger$
is a vector obtained by merging those of chemical graphs 
$\Co_A$: CID~10076784   and $\Co_B$: CID~44340250 
in   Figure~\ref{fig:instance_I_c_I_d}(c) and (d), respectively.  

\medskip \noindent
{\bf Stage~4.  } 
We executed Stage~4 for five properties $\pi\in \{${\sc Hc, Vd, OptR,  IhcLiq, Vis}$\}$.  

 For the MILP formulation  $\mathcal{M}(x,y;\mathcal{C}_1)$
  in Section~\ref{sec:linear_regression}, 
we use the prediction function $\eta_{\w,b}$  
 that attained the median  test $\mathrm{R}^2$ in Table~\ref{table:phase1a}.
 To solve an MILP   in Stage~4, we used 
{\tt  CPLEX} version 12.10.
Tables~\ref{table:stages_4_5_Hc} to \ref{table:stages_4_5_Vis}  show
   the computational results of the experiment
in Stage~4 for the five properties, 
 where we denote the following:
\begin{enumerate} [nosep,  leftmargin=*]

\item[-]  
  $ \underline{y}^*,~\overline{y}^* $:  
 lower and upper bounds $\underline{y}^*, \overline{y}^*\in \R$ 
  on the value $a(\Co)$ of a chemical graph $\Co$ to be inferred; 
  
\item[-]  
 $\#$v (resp.,  $\#$c): 
 the number  of variables (resp., constraints)  in the MILP  in Stage~4;

\item[-]   
 I-time: the   time (sec.) to solve the MILP  in Stage~4; 

\item[-]  
    $n$:  the number  $n(\Co^\dagger)$  of  non-hydrogen atoms
     in the chemical graph $\Co^\dagger$   inferred in Stage~4;   and 
\item[-]  
  $\nint$:  the number  $\nint(\Co^\dagger)$ of interior-vertices in
  the chemical graph $\Co^\dagger$   inferred in Stage~4;
      
\item[-]  
$\eta(f(\Co^\dagger))$: the predicted property value 
$\eta(f(\Co^\dagger))$ of the chemical graph $\Co^\dagger$ inferred 
in Stage~4.
\end{enumerate}

\begin{figure}[!htb]
\begin{center} 
 \includegraphics[width=.98\columnwidth]{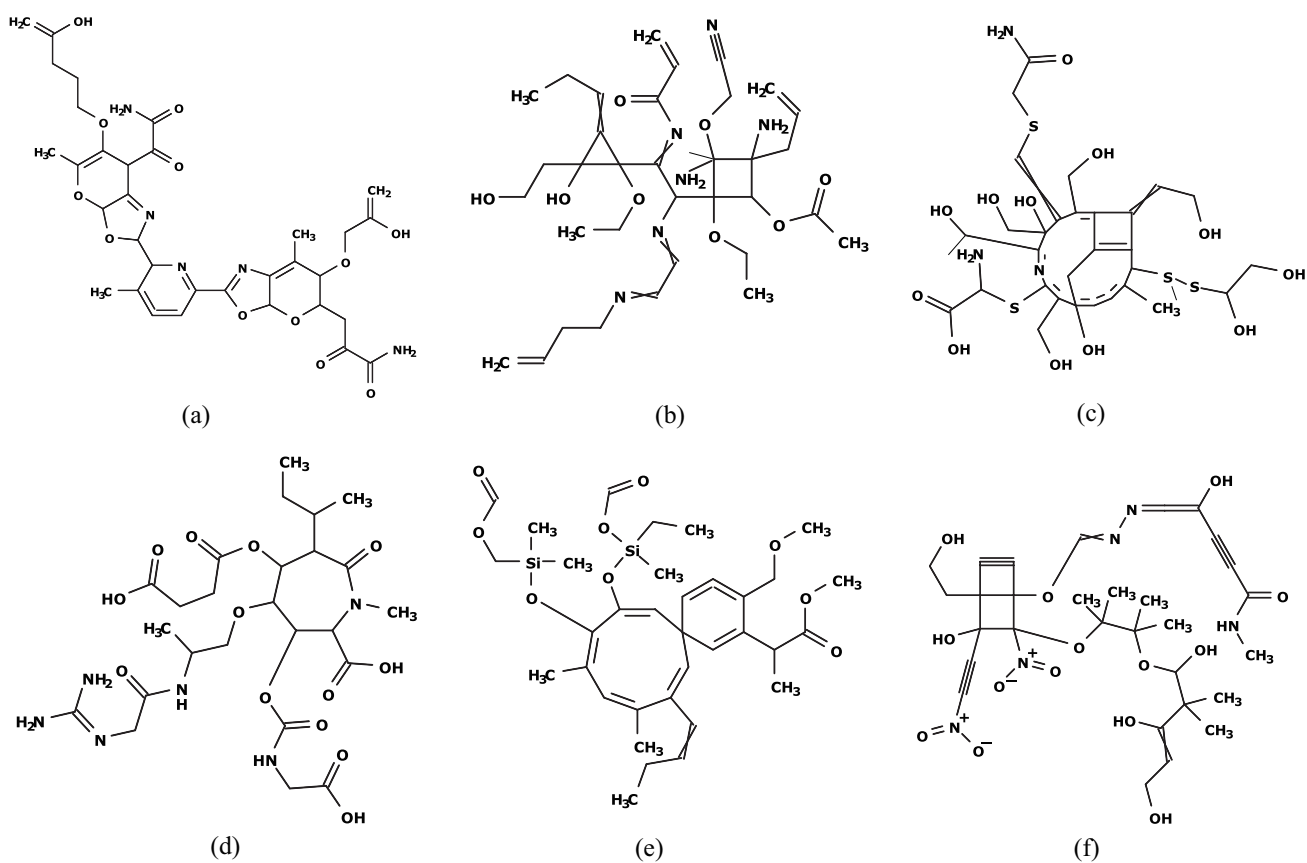}
\end{center}
\caption{ 
(a)  $\Co^\dagger$ with $\eta(f(\Co^\dagger))=13703.3$ inferred
 from   $I_{\mathrm{c}}$ with $(\underline{y}^*, \overline{y}^*) =(13700,13800)$  of  {\sc Hc};   
(b)  $\Co^\dagger$ with $\eta(f(\Co^\dagger))=21.62$ inferred
 from   $I_{\mathrm{b}}^2$ with $(\underline{y}^*, \overline{y}^*) =(21,22)$  of  {\sc Vd};   
(c)  $\Co^\dagger$ with $\eta(f(\Co^\dagger))= 70.9$ inferred
 from   $I_{\mathrm{b}}^4$ with $(\underline{y}^*, \overline{y}^*) =(70,71)$  of  {\sc OptR};   
(d)  $\Co^\dagger$ with $\eta(f(\Co^\dagger))=1198.8$ inferred
 from   $I_{\mathrm{d}}$ with $(\underline{y}^*, \overline{y}^*) =(1190,1210)$  of  {\sc IhcLiq};   
(e)   $\Co^\dagger$ with $\eta(f(\Co^\dagger))=1.880$ inferred 
from   $I_{\mathrm{b}}^3$ with $(\underline{y}^*, \overline{y}^*) =(1.85,1.90)$  of  {\sc Vis}; 
(f)   $\Co^\dagger$ inferred 
from   $I_{\mathrm{b}}^4$ with lower and upper bounds on 
the predicted property value $\eta_\pi(f(\Co^\dagger))$ of   property 
  $\pi\in\{${\sc Kow, Lp, Sl}$\}$ in Table~\ref{table:stage_4_KowLpSl}. 
}
\label{fig:MILP_solutions}  
\end{figure}

 \begin{table}[h!]\caption{ Results of Stages~4 and 5 for {\sc Hc}.}  
 \begin{center}
 \begin{tabular}{@{}  c     c  r r r r r c  r r r     @{}}\toprule                
 inst. & $ \underline{y}^*,~\overline{y}^* $ & $\#$v  &  $\#$c~\!\!  &  
   {\small I-time}\!\! & $n$~  &  \!\!$\nint$  & {\small $\eta(f(\Co^\dagger))$}\!\!  & 
                 {\small  D-time} &  {\small $\Co$-LB } &  {\small $\#\Co$}    \\ \midrule
   $I_{\mathrm{a}}$ & 5950,\,6050  &  9902 & 9255 & 4.6 & 44 & 25  & 5977.9   & 0.068 & 1  & 1       \\%
  $I_{\mathrm{b}}^1$ & 5950,\,6050 &  9404 & 6776 & 1.7 & 36 & 10  & 6007.1 & 0.048& 6 & 6   \\%
  $I_{\mathrm{b}}^2$ & 5950,\,6050 & 11729 & 9891 & 16.7 & 50 & 25  & 6043.7  & 38.7&  $2.4\!\! \times\!\! 10^5$ & 100    \\%
  $I_{\mathrm{b}}^3$ & 5950,\,6050 & 11510 & 9894 & 16.3 & 47 & 25 & 6015.4  &  0.353 &  8724& 100    \\%
  $I_{\mathrm{b}}^4$ & 5950,\,6050 & 11291 & 9897 & 9.0 & 49 & 26  & 5971.6 & 0.304     & 84  & 84    \\
  $I_{\mathrm{c}}$    & 13700,\,13800  &  6915 & 7278 &  0.7 & 50 & 33   & 13703.3 &    0.016 & 1  & 1   \\
  $I_{\mathrm{d}}$    & 13700,\,13800   &  5535 & 6781 & 4.9 & 44 & 23 & 13704.7  &   0.564 &  $4.3\!\!\times\!\!10^{5}$ & 100   \\
   \bottomrule
   \end{tabular}\end{center}\label{table:stages_4_5_Hc}
\end{table}
 Figure~\ref{fig:MILP_solutions}(a) illustrates  the chemical graph  $\Co^\dagger$  inferred
 from   $I_{\mathrm{c}}$ with $(\underline{y}^*, \overline{y}^*) =(13700,13800)$ 
  of  {\sc Hc} in Table~\ref{table:stages_4_5_Hc}. 
 
 \begin{table}[h!]\caption{ Results of Stages~4 and 5 for {\sc Vd}.}  
 \begin{center}
 \begin{tabular}{@{}  c     c  r r r r r c  r r r     @{}}\toprule                
 inst. & $ \underline{y}^*,~\overline{y}^* $ & $\#$v  &  $\#$c~\!\!  &  
   {\small I-time}\!\! & $n$~  &  \!\!$\nint$  & {\small $\eta(f(\Co^\dagger))$}\!\! & 
                 {\small  D-time} &  {\small $\Co$-LB } &  {\small $\#\Co$}    \\ \midrule
  $I_{\mathrm{a}}$ & 16,\,17  & 9481 & 9358 & 1.6 & 38 & 23  & 16.83 &0.070 & 1  & 1       \\%
  $I_{\mathrm{b}}^1$ & 16,\,17 &  9928 & 6986 & 1.5 & 35 & 12   & 16.68 & 0.206 & 48&48   \\%
  $I_{\mathrm{b}}^2$ & 21,\,22 & 12373 & 10101 & 10.0 & 48 & 25  & 21.62  & 0.104 &  20 & 20    \\%
  $I_{\mathrm{b}}^3$ & 21,\,22 & 12159 & 10104 & 6.5 & 48 & 25 & 21.95 &  3.65 &    $8.6\!\! \times\!\! 10^5$ &  100     \\%
  $I_{\mathrm{b}}^4$ & 21,\,22 & 11945 & 10107 & 8.1 & 48 & 25  & 21.34  & 0.057    & 6  & 6     \\
  $I_{\mathrm{c}}$    & 21,\,22  &  7073 & 7438 &  0.7 & 50 & 34   & 21.89  &   0.016 & 1   & 1   \\
  $I_{\mathrm{d}}$    &  17,\,18  &  5693 & 6942 & 2.1 & 41 & 23 & 17.94  &    0.161  &   216 & 100   \\
   \bottomrule
   \end{tabular}\end{center}\label{table:stages_4_5_Vd}
\end{table}
Figure~\ref{fig:MILP_solutions}(b) illustrates  the chemical graph  $\Co^\dagger$  inferred 
 from   $I_{\mathrm{b}}^2$ with $(\underline{y}^*, \overline{y}^*) =(21,22)$  of  {\sc Vd}
  in Table~\ref{table:stages_4_5_Vd}.   
 
\begin{table}[h!]\caption{ Results of Stages~4 and 5 for {\sc OptR}.}  
 \begin{center}
 \begin{tabular}{@{}  c     c  r r r r r c  r r r     @{}}\toprule                
 inst. & $ \underline{y}^*,~\overline{y}^* $ & $\#$v  &  $\#$c~\!\!  &  
   {\small I-time}\!\! & $n$~  &  \!\!$\nint$  & {\small $\eta(f(\Co^\dagger))$}\!\! & 
                 {\small  D-time} &  {\small $\Co$-LB } &  {\small $\#\Co$}    \\ \midrule
  $I_{\mathrm{a}}$ & 70,\,71  &  8962 & 9064 & 3.5 & 40 & 23  & 70.1  & 0.061  & 1  & 1       \\%
  $I_{\mathrm{b}}^1$ & 70,\,71 &  9432 & 6662 & 2.7 & 37 & 14   & 70.1  &0.185 & 2622 &100   \\%
  $I_{\mathrm{b}}^2$ & 70,\,71 & 11818 & 9773 & 10.0 & 50 & 25  &  70.8  & 0.041 &  4 & 4   \\%
  $I_{\mathrm{b}}^3$ & 70,\,71 & 11602 & 9776 & 10.2 & 50 & 25 & 70.2  &  0.241 &     60&  60    \\%
  $I_{\mathrm{b}}^4$ & 70,\,71 & 11386 & 9779 & 24.7 & 49 & 25  & 70.9 & 6.39     & $4.6\!\! \times\!\! 10^5$  & 100    \\
  $I_{\mathrm{c}}$    & -112,\,-111  &  6807 & 7170 &  1.8 & 50 & 32   & -111.9  &  0.016 & 1   & 1   \\
  $I_{\mathrm{d}}$    & 70,\,71  &  5427 & 6673 & 6.1 & 42 & 23 & 70.2 &    0.127   &   78768 & 100   \\
   \bottomrule
   \end{tabular}\end{center}\label{table:stages_4_5_OptR}
\end{table}
Figure~\ref{fig:MILP_solutions}(c) illustrates  the chemical graph  $\Co^\dagger$  inferred 
 from   $I_{\mathrm{b}}^4$ with $(\underline{y}^*, \overline{y}^*) =(70,71)$  of  {\sc OptR}
  in Table~\ref{table:stages_4_5_OptR}.

\begin{table}[h!]\caption{ Results of Stages~4 and 5 for {\sc IhcLiq}.}  
 \begin{center}
 \begin{tabular}{@{}  c     c  r r r r r c  r r r     @{}}\toprule                
 inst. & $ \underline{y}^*,~\overline{y}^* $ & $\#$v  &  $\#$c~\!\!  &  
   {\small I-time}\!\! & $n$~  &  \!\!$\nint$  & {\small $\eta(f(\Co^\dagger))$}\!\! & 
                 {\small  D-time} &  {\small $\Co$-LB } &  {\small $\#\Co$}    \\ \midrule
  $I_{\mathrm{a}}$ & 1190,\,1210  &  10180 & 9538 & 3.9 & 48 & 26  & 1208.5 & 0.071 & 2  & 2       \\%
  $I_{\mathrm{b}}^1$ & 1190,\,1210 &  10784 & 7191 & 2.4 & 35 & 14   & 1206.7 & 0.082 & 12 &12  \\%
  $I_{\mathrm{b}}^2$ & 1190,\,1210 & 13482 & 10302 & 14.1 & 47 & 25  & 1206.7   &  0.11 & 12 &12    \\%
  $I_{\mathrm{b}}^3$ & 1190,\,1210 & 13275 & 10301 & 9.0 & 49 & 27 & 1209.9 &  0.090 &  24 & 24   \\%
  $I_{\mathrm{b}}^4$ & 1190,\,1210 & 13128 & 10306 & 16.5 & 50 & 25  &1208.4 & 0.424  &     2388&  100    \\
  $I_{\mathrm{c}}$    & 1190,\,1210  &  7193 & 7560 &  0.8 & 50 & 33   & 1196.5 &    0.016 & 1   &1   \\
  $I_{\mathrm{d}}$    & 1190,\,1210  &  5813 & 7063 & 2.2 & 44 & 23 & 1198.8  & 5.63&  $5.2\!\!\times\!\!10^{5}$ & 100   \\
   \bottomrule
   \end{tabular}\end{center}\label{table:stages_4_5_IhcLiq}
\end{table}
 Figure~\ref{fig:MILP_solutions}(d) illustrates  the chemical graph  $\Co^\dagger$  inferred 
 from   $I_{\mathrm{d}}$ with $(\underline{y}^*, \overline{y}^*) =(1190,1210)$  of  {\sc IhcLiq}
  in Table~\ref{table:stages_4_5_IhcLiq}.

 \begin{table}[h!]\caption{ Results of Stages~4 and 5 for {\sc Vis}.}  
 \begin{center}
 \begin{tabular}{@{}  c     c  r r r r r c  r r r     @{}}\toprule                
 inst. & $ \underline{y}^*,~\overline{y}^* $ & $\#$v  &  $\#$c~\!\!  &  
   {\small I-time}\!\! & $n$~  &  \!\!$\nint$  & {\small $\eta(f(\Co^\dagger))$}\!\! & 
                 {\small  D-time} &  {\small $\Co$-LB } &  {\small $\#\Co$}    \\ \midrule
  $I_{\mathrm{a}}$ &  1.25,\,1.30   &  6847 & 8906 & 1.3 & 38 & 22  &  1.295 & 0.042  & 2  & 2       \\%
  $I_{\mathrm{b}}^1$ &  1.25,\,1.30  &  7270 & 6397 & 2.5 & 36 & 15   &  1.272 & 0.155  & 140&100   \\%
  $I_{\mathrm{b}}^2$ &  1.85,\,1.90 & 8984 & 9512 & 8.9 & 45 & 25  & 1.879  & 0.149 &  288 & 100    \\%
  $I_{\mathrm{b}}^3$ &  1.85,\,1.90 & 8741 & 9515 & 16.2 & 45 & 26 & 1.880 &  0.137 &     4928 &  100    \\%
  $I_{\mathrm{b}}^4$ &  1.85,\,1.90 & 8498 & 9518 & 8.1 & 45 & 25  &  1.851 & 0.13     & 660  & 100   \\
  $I_{\mathrm{c}}$    &  2.75,\,2.80   &  6813 & 7162 &  1.0 & 50 & 33   &  2.763 &    0.025  & 4   & 4   \\
  $I_{\mathrm{d}}$    &  1.85,\,1.90  &  5433 & 6665 & 2.7 & 41 & 23 & 1.881 &    0.138 &   4608 & 100   \\
   \bottomrule
   \end{tabular}\end{center}\label{table:stages_4_5_Vis}
\end{table}
Figure~\ref{fig:MILP_solutions}(e) illustrates  the chemical graph  $\Co^\dagger$  inferred 
from   $I_{\mathrm{b}}^3$ with $(\underline{y}^*, \overline{y}^*) =(1.85,1.90)$  of  {\sc Vis}~in Table~\ref{table:stages_4_5_Vis}.

From Tables~\ref{table:stages_4_5_Hc} to \ref{table:stages_4_5_Vis},
we observe that an instance with a large number of variables and constraints 
takes more running time than those with a smaller size in general.
In this experiment, we prepared several different types of instances:
 instances  $I_{\mathrm{a}}$ and $I_{\mathrm{c}}$ have restricted seed graphs,
the other  instances  have abstract seed graphs and
instances $I_{\mathrm{c}}$ and $I_{\mathrm{d}}$ have restricted set of fringe-trees.
All instances in this experiment are solved in a few seconds to around 30 seconds
with our MILP formulation.

\bigskip
\noindent 
{\bf 
Inferring a chemical graph with target values in multiple properties}
Once we obtained prediction functions $\eta_\pi$ for several properties $\pi$,
it is easy to include MILP formulations for these functions $\eta_\pi$
into a single MILP $\mathcal{M}(x,y;\mathcal{C}_1)$ so as to
infer a chemical graph that satisfies given target values $y^*$ for
these properties at the same time.
As an additional experiment in Stage~4, 
we inferred a chemical graph  that has a desired predicted value
each of three properties   {\sc Kow, Lp} and {\sc Sl},
where we used the prediction function
$\eta_\pi$ for each property $\pi\in\{${\sc Kow, Lp, Sl}$\}$ constructed in
Stage~3.
Table~\ref{table:stage_4_KowLpSl} shows the result of Stage~4
for inferring a chemical graph $\Co^\dagger$
 from instances $I_{\mathrm{b}}^2$,  $I_{\mathrm{b}}^3$ and  $I_{\mathrm{b}}^4$ 
with  $\Lambda=\{ \ttH,\ttC,\ttN,\ttO, \ttS_{(2)},\ttS_{(6)},\ttCl\}$, 
 where we denote the following: 
\begin{enumerate} [nosep,  leftmargin=*]
\item[-]  
$\pi$: one of the three properties  {\sc Kow, Lp} and {\sc Sl} 
 used in the experiment;

\item[-]  
  $ \underline{y}^*_\pi,~\overline{y}^*_\pi $:  
 lower and upper bounds $\underline{y}^*_\pi, \overline{y}^*_\pi\in \R$ 
  on  the predicted property value 
$\eta_\pi(f(\Co^\dagger))$ of   property 
  $\pi\in\{${\sc Kow, Lp, Sl}$\}$
   for a chemical graph $\Co^\dagger$ to be inferred;
  
\item[-]  
 $\#$v (resp.,  $\#$c): 
 the number  of variables (resp., constraints)  in the MILP  in Stage~4;  
 
\item[-]   
 I-time: the   time (sec.) to solve the MILP  in Stage~4;  
 
\item[-]  
    $n$:  the number  $n(\Co^\dagger)$  of  non-hydrogen atoms
     in the chemical graph $\Co^\dagger$   inferred in Stage~4;  
\item[-]  
 $\nint$:  the number  $\nint(\Co^\dagger)$ of interior-vertices in
  the chemical graph $\Co^\dagger$   inferred in Stage~4; and 
      
\item[-]  
$\eta_\pi(f(\Co^\dagger))$: the predicted property value 
$\eta_\pi(f(\Co^\dagger))$ of   property 
  $\pi\in\{${\sc Kow, Lp, Sl}$\}$
   for the chemical graph $\Co^\dagger$ inferred in Stage~4.
\end{enumerate}

 \begin{table}[h!]\caption{Results of Stage~4 for instance 
 $I_{\mathrm{b}}^i, i=2,3,4$ 
 with specified target values of three properties {\sc Kow, Lp} and {\sc Sl}.} 
 \begin{center}
 \begin{tabular}{@{}  c c  c  c  r r r r r c    @{}}\toprule                
  inst. & $\pi$ & $ \underline{y}_\pi^*,~\overline{y}_\pi^* $ & $\#$v  &  $\#$c~   &  
   {\small I-time}  & $n$~  &   $\nint$  & {\small $\eta_\pi(f(\Co^\dagger))$  }    \\ \midrule
                              & {\sc Kow}  &   -7.50,\,-7.40   &   &   &   &   &                            &  -7.41~~~     \\%
 $I_{\mathrm{b}}^2$  & {\sc Lp}  &  -1.40,\,-1.30   &  14574 & 11604  & 62.7 & 50  & 30 &  -1.33~~~    \\%
                               &   {\sc Sl}   &  -11.6,\,-11.5  &    &   &   &   &                          &   -11.52~~~     \\  
 \hline 
                              & {\sc Kow}  &   -7.40,\,-7.30   &   &   &   &   &                            &  -7.38~~~     \\%
 $I_{\mathrm{b}}^3$  & {\sc Lp}  &  -2.90,\,-2.80   &  14370 & 11596  & 35.5 & 48  & 25 &  -2.81~~~    \\%
                               &   {\sc Sl}   &  -11.6,\,-11.4  &    &   &   &   &                          &   -11.52~~~     \\  
\hline 
                              & {\sc Kow}  &   -7.50,\,-7.40   &   &   &   &   &                            &  -7.48~~~     \\%
 $I_{\mathrm{b}}^4$  & {\sc Lp}  &  -0.70,\,-0.60   &  14166 & 11588  & 71.7 & 49  & 26 &  -0.63~~~   \\%
                               &   {\sc Sl}   &  -11.4,\,-11.2  &    &   &   &   &                          &   -11.39~~~     \\  
   \bottomrule
   \end{tabular}\end{center}\label{table:stage_4_KowLpSl}
\end{table}
Figure~\ref{fig:MILP_solutions}(f) illustrates  the chemical graph  $\Co^\dagger$  inferred 
from  $I_{\mathrm{b}}^4$  with 
$(\underline{y}^*_{\pi_1}, \overline{y}^*_{\pi_1}) =(-7.50,$ $-7.40)$,
$(\underline{y}^*_{\pi_2}, \overline{y}^*_{\pi_2}) =(-0.70,-0.60 )$ and 
$(\underline{y}^*_{\pi_3}, \overline{y}^*_{\pi_3}) =(-11.4,-11.2)$
for $\pi_1=${\sc Kow}, $\pi_2=${\sc Lp} and  $\pi_3=${\sc Sl}, respectively.

\medskip \noindent
{\bf Stage~5.  } 
We executed Stage~5 to generate a more number of target chemical graphs $\Co^*$,
where we call a chemical graph $\Co^*$ a {\em chemical isomer} of
a target chemical graph $\Co^\dagger$ of a topological specification $\sigma$
if $f(\Co^*)=f(\Co^\dagger)$ and $\Co^*$ also satisfies the same topological specification $\sigma$.
We computed  chemical isomers  $\Co^*$ of 
each target chemical graph  $\Co^\dagger$ inferred in Stage~4.
We execute an  algorithm for generating chemical isomers of   $\Co^\dagger$
up to 100 when the number of all chemical isomers exceeds 100.
Such an algorithm can be obtained from the dynamic programming proposed
 by Tanaka et~al.~\cite{TZAHZNA21} with a slight modification. 
The algorithm first decomposes $\Co^\dagger$ into a set of acyclic chemical graphs,
next replaces each acyclic chemical graph $T$ with another  acyclic chemical graph $T'$ that admits
the same feature vector as that of $T$,
 and finally assembles the resulting acyclic chemical graphs
into a chemical isomer $\Co^*$ of $\Co^\dagger$. 
The algorithm can compute a lower bound 
on the total number of all chemical isomers of $\Co^\dagger$
without generating all of them.

Tables~\ref{table:stages_4_5_Hc} to  \ref{table:stages_4_5_Vis}  show
   the computational results of the experiment
in Stage~5 for the five properties, 
 where we denote the following:
\begin{enumerate}[nosep,  leftmargin=*]
\item[-]
 D-time: the running time (sec.) to execute the dynamic programming algorithm
 in Stage~5 to compute a lower bound on the number 
 of all chemical isomers  $\Co^*$ of  $\Co^\dagger$   
 and generate all (or up to 100) chemical isomers $\Co^*$;
\item[-]
 $\Co$-LB: a lower bound on the number of all chemical isomers $\Co^*$ of
$\Co^\dagger$; and

\item[-]
 $\#\Co$: the number of all (or up to 100) chemical isomers $\Co^*$ of  $\Co^\dagger$  
 generated in Stage~5.
\end{enumerate} 
 
From Tables~\ref{table:stages_4_5_Hc} to \ref{table:stages_4_5_Vis}, we observe that
 the running time for generating up to 100 target chemical graphs in Stage~5 is less than
 0.4 second for many cases.
 For some chemical graph $\Co^\dagger$, no chemical isomer was found by our algorithm.
 This is because each acyclic chemical graph in the decomposition of $\Co^\dagger$
 has no alternative acyclic chemical graph than the original one. 
On the other hand, some  chemical graph $\Co^\dagger$ such as the one in  $I_{\mathrm{d}}$
in  Tables~\ref{table:stages_4_5_Hc} admits extremely large number of 
chemical isomers $\Co^*$. 
Remember that we know such a lower bound $\Co$-LB  on the number of chemical isomers
without generating all of them.


\section{Concluding Remarks}\label{sec:conclude}
 
 In the previous applications of 
 the framework of inferring chemical graphs, 
 artificial neural network (ANN) and decision tree have been used 
 for the machine learning of Stage~3.
 In this paper, we used linear regression in Stage~3 for the first time
 and derived an MILP formulation that simulates the computation process
 of linear regression.
We also extended a way of specifying a target value $y^*$ in 
a property so that  the predicted value $\eta(f(\C^\dagger))$
of a target chemical graph $\C^\dagger$ is required to belong to
 an interval between two specified values $\underline{y}^*$ and $\overline{y}^*$.
In this paper, we modified a model of chemical compounds so that
 multi-valence chemical elements, cation and anion are treated, 
and  introduced the rank and the adjacency-configuration of leaf-edges
as new descriptors  in a feature vector of a chemical graph. 
We implemented the new system  of the framework and 
conducted computational experiments  for Stages~1 to 5.  
We found 18 properties for which
linear regression delivers a relatively good prediction function
 by using our feature vector based on the two-layered model.
We also observed that an MILP formulation for inferring a chemical graph
in Stage~4 can be solved efficiently over different types of test instances
with complicated topological specifications.
The experimental result  suggests that our method 
can infer chemical graphs with  up to 50 non-hydrogen atoms. 
 
 It is left as a future work to use other learning methods such as
 random forest, graph convolution networks and an ensemble method 
 in Stages~3 and 4 in the framework.

\clearpage
 \appendix
\centerline{\bf\LARGE Appendix}

\section{A Full Description of Descriptors}\label{sec:descriptor}

Associated with the two functions 
$\alpha$ and $\beta$ in a chemical graph $\Co=(H,\alpha,\beta)$,
we introduce   functions  
 $\ac: V(E)\to (\Lambda\setminus\{\ttH\})\times (\Lambda\setminus\{\ttH\})\times [1,3]$, 
 $\cs: V(E)\to (\Lambda\setminus\{\ttH\})\times [1,6]$ and
$\ec: V(E)\to ((\Lambda\setminus\{\ttH\})\times [1,6])\times ((\Lambda\setminus\{\ttH\})\times [1,6])\times [1,3]$
in the following. 

 To represent  a feature of the exterior  of  $\Co$, 
  a  chemical rooted tree in $\mathcal{T}(\Co)$ is
  called a {\em fringe-configuration} of $\Co$. 

We also represent leaf-edges in the exterior of $\Co$.
For a leaf-edge $uv\in E(\anC)$ with $\deg_{\anC}(u)=1$, we define
the {\em adjacency-configuration} of $e$ to be an ordered tuple
$(\alpha(u),\alpha(v),\beta(uv))$. 
Define 
\[ \Gac^\lf\triangleq \{(\ta,\tb,m)\mid \ta,\tb\in\Lambda, 
m\in[1,\min\{\val(\ta),\val(\tb)\}]\} \]
as a set of possible adjacency-configurations for leaf-edges. 

To  represent a feature of an interior-vertex $v\in V^\inte(\Co)$ such that
$\alpha(v)=\ta$  and  $\deg_{\anC}(v)=d$
(i.e., the number of non-hydrogen atoms adjacent to $v$ is $d$) 
   in a chemical   graph  $\Co=(H,\alpha,\beta)$,
 we use  a pair $(\ta, d)\in (\Lambda\setminus\{{\tt H}\})\times [1,4]$,
 which we call the {\em chemical symbol} $\cs(v)$ of the vertex $v$.
 We treat $(\ta, d)$ as a single symbol $\ta d$,  and  
define $\Ldg$   to be  the set of all chemical symbols
$\mu=\ta d\in  (\Lambda\setminus\{{\tt H}\})\times [1,4]$.  

We define a method for featuring interior-edges  as follows.
Let $e=uv\in E^\inte(\Co)$  be 
 an interior-edge $e=uv\in E^\inte(\Co)$ 
 such that $\alpha(u)=\ta$, $\alpha(v)=\tb$ and $\beta(e)=m$ 
   in a chemical graph  $\Co=(H,\alpha,\beta)$.
To feature this edge $e$, 
 we use a tuple $(\ta,\tb,m)\in (\Lambda\setminus\{{\tt H}\})
    \times (\Lambda\setminus\{{\tt H}\})\times [1,3]$,
 which we call the {\em adjacency-configuration} $\ac(e)$ of the edge $e$. 
 We introduce a total order $<$ over the elements in $\Lambda$
 to distinguish  between $(\ta,\tb, m)$ and $(\tb,\ta, m)$ 
 $(\ta\neq \tb)$ notationally.
 For a tuple  $\nu=(\ta,\tb, m)$,
 let $\overline{\nu}$ denote the tuple $(\tb,\ta, m)$.

Let $e=uv\in E^\inte(\Co)$  be 
an  interior-edge $e=uv\in E^\inte(\Co)$ 
 such that $\cs(u)=\mu$, $\cs(v)=\mu'$ and $\beta(e)=m$ 
   in a chemical  graph  $\Co=(H,\alpha,\beta)$.
To feature this edge $e$, 
 we use a tuple $(\mu,\mu',m)\in \Ldg\times \Ldg\times [1,3]$, 
 which we call  the {\em edge-configuration} $\ec(e)$ of the edge $e$. 
 We introduce a total order $<$ over the elements in $\Ldg$
 to distinguish between $(\mu,\mu', m)$ and $(\mu', \mu, m)$ 
 $(\mu \neq \mu')$ notationally. 
 For a tuple  $\gamma=(\mu,\mu', m)$,
 let $\overline{\gamma}$ denote the tuple $(\mu', \mu, m)$. 

Let $\pi$ be a chemical property for which we will construct
a prediction function $\eta$ from a feature
vector $f(\C)$ of a chemical graph $\C$ 
to a predicted value $y\in \mathbb{R}$
for the  chemical property of $\C$.

We first choose a set $\Lambda$ of chemical elements
 and then collect a data set  $D_{\pi}$ of
  chemical compounds  $C$ whose 
  chemical elements belong to $\Lambda$,
  where we regard  $D_{\pi}$ as a set of chemical graphs $\C$
  that represent the chemical compounds $C$  in  $D_{\pi}$.
To define the interior/exterior of 
chemical graphs  $\C\in D_{\pi}$,
we  next choose a branch-parameter ${\rho}$, where
 we recommend ${\rho}=2$.  
 
Let $\Lambda^\inte(D_\pi)\subseteq \Lambda$ 
(resp., $\Lambda^\ex(D_\pi)\subseteq \Lambda$)
denote the set  of chemical elements  used in
the set $V^\inte(\C)$ of interior-vertices
(resp., the set $V^\ex(\C)$ of  exterior-vertices) of $\C$
 over all chemical graphs $\C\in D_\pi$, 
and $\Gamma^\inte(D_\pi)$
denote the set of edge-configurations used in
the set $E^\inte(\C)$  of interior-edges in $\C$
 over all chemical graphs $\C\in D_\pi$. 
Let $\mathcal{F}(D_\pi)$ denote the set of
chemical rooted trees $\psi$  
r-isomorphic to a chemical rooted tree in $\mathcal{T}(\C)$
  over all chemical graphs $\C\in D_\pi$,
  where possibly a chemical rooted tree $\psi\in \mathcal{F}(D_\pi)$
  consists of a single chemical element $\ta\in \Lambda\setminus \{{\tt H}\}$.
  
We define an integer encoding of a finite set $A$ of elements
to be a bijection $\sigma: A \to [1, |A|]$, 
where we denote by $[A]$   the set $[1, |A|]$ of integers.
Introduce  an integer coding of each of the   sets 
$\Lambda^\inte(D_\pi)$, $\Lambda^\ex(D_\pi)$, 
$\Gamma^\inte(D_\pi)$ and $\mathcal{F}(D_\pi)$. 
Let $[\ta]^\inte$  
(resp., $[\ta]^\ex$)  denote   
the coded integer of  an element $\ta\in \Lambda^\inte(D_\pi)$
(resp., $\ta\in \Lambda^\ex(D_\pi)$),  
$[\gamma]$   denote  
the coded integer of  an element $\gamma$ in $\Gamma^\inte(D_\pi)$
and 
$[\psi]$   denote  an element $\psi$ in $\mathcal{F}(D_\pi)$. 
 
 Over 99\% of  chemical compounds $\C$ with up to
  100 non-hydrogen atoms in  PubChem  have degree at most 4
 in the hydrogen-suppressed graph $\anC$~\cite{AZSSSZNA20}. 
We assume that a chemical graph $\C$
 treated in this paper satisfies  $\deg_{\anC}(v)\leq 4$
in the hydrogen-suppressed graph $\anC$.
 
In our model, we  use an integer 
  $\mathrm{mass}^*(\ta)=\lfloor 10\cdot \mathrm{mass}(\ta)\rfloor$, 
 for each $\ta\in \Lambda$.
 
  We define the {\em feature vector} $f(\C)$ 
  of a  chemical graph $\C=(H,\alpha,\beta)\in D_{\pi}$ 
  to be a vector that consists of the following  
non-negative integer descriptors $\dcp_i(\C)$, $i\in [1,K]$, where 
$K = 14+ |\Lambda^\inte(D_\pi)|+|\Lambda^\ex(D_\pi)|
         +|\Gamma^\inte(D_\pi)|+|\mathcal{F}(D_\pi)|+|\Gac^\lf|$. 


\begin{enumerate}  
\item   
$\dcp_1(\C)$: the number  $|V(H)|-|\VH|$ of non-hydrogen atoms  in  $\C$.  
 
\item   
$\dcp_2(\C)$: the rank $\mathrm{r}(\C)$ of   $\C$.  

\item 
$\dcp_3(\C)$:  the number $|V^\inte(\C)|$ of interior-vertices in  $\C$.
  
\item 
$\dcp_4(\C)$: 
the average $\overline{\mathrm{ms}}(\C)$ of mass$^*$ 
over all atoms in $\C$; \\
 i.e., $\overline{\mathrm{ms}}(\C)\triangleq 
 \frac{1}{|V(H)|}\sum_{v\in V(H)}\mathrm{mass}^*(\alpha(v))$. 

\item 
$\dcp_i(\C)$,  $i=4+d,   d\in [1,4]$: 
the number $\dg_d^{\oH} (\C)$ 
 of non-hydrogen vertices $v\in V(H)\setminus \VH$
 of degree $\deg_{\anC}(v)=d$
 in the hydrogen-suppressed chemical graph $\anC$.  
 
\item 
$\dcp_i(\C)$,  $i=8+d,   d\in [1,4]$: 
the number $\dg_d^\inte(\C)$
 of interior-vertices of interior-degree  $\deg_{\C^\inte}(v)=d$
  in the interior $\C^\inte=(V^\inte(\C),E^\inte(\C))$ of  $\C$. 
  
   
\item $\dcp_i(\C)$, $i=12+m$,  $m\in[2,3]$: 
the number $\bd_m^\inte(\C)$
 of  interior-edges with bond multiplicity $m$ in  $\C$; 
 i.e., $\bd_m^\inte(\C)\triangleq \{e\in E^\inte(\C)\mid \beta(e)=m\}$.

\item $\dcp_i(\C)$, $i=14+[\ta]^\inte$, 
 $\ta\in \Lambda^\inte(D_\pi)$: 
 the frequency $\na_\ta^\inte(\C)=|V_\ta(\C)\cap V^\inte(\C) |$ 
 of chemical element $\ta$ in
 the set $V^\inte(\C)$ of  interior-vertices in  $\C$. 
 
\item $\dcp_i(\C)$, 
$i=14+|\Lambda^\inte(D_\pi)|+[\ta]^\ex$, 
 $\ta\in \Lambda^\ex(D_\pi)$: 
 the frequency $\na_\ta^\ex(\C)=|V_\ta(\C)\cap V^\ex(\C) |$
  of chemical element $\ta$ in
 the set $V^\ex(\C)$ of  exterior-vertices in  $\C$. 
 
\item $\dcp_i(\C)$, 
$i=14+|\Lambda^\inte(D_\pi)|+|\Lambda^\ex(D_\pi)|+ [\gamma]$, 
$\gamma \in \Gamma^\inte(D_\pi)$: 
the frequency $\ec_{\gamma} (\C)$ of edge-configuration $\gamma$
in the set $E^\inte(\C)$ of interior-edges   in  $\C$.

\item $\dcp_i(\C)$, 
$i= 14+|\Lambda^\inte(D_\pi)|+|\Lambda^\ex(D_\pi)|
+ |\Gamma^\inte(D_\pi)|+ [\psi]$,  
 $\psi \in \mathcal{F}(D_\pi)$: 
the frequency $\fc_{\psi}(\C)$ of fringe-configuration $\psi $
in the set of ${\rho}$-fringe-trees in  $\C$. 

\item $\dcp_i(\C)$, 
$i= 14+|\Lambda^\inte(D_\pi)|+|\Lambda^\ex(D_\pi)|
+ |\Gamma^\inte(D_\pi)|+|\mathcal{F}(D_\pi)|+ [\nu]$,  
 $\nu \in \Gac^\lf$: 
the frequency $\ac_{\nu}^\lf(\C)$ of adjacency-configuration $\nu$
in the set of leaf-edges in  $\anC$. %
\end{enumerate}

\section{Specifying Target Chemical Graphs}\label{sec:specification} 

Given a prediction function $\eta$ and 
a target value $y^*\in \mathbb{R}$, 
we call a chemical graph $\C^*$ such that $\eta(x^*)=y^*$
for the feature vector $x^*=f(\C^*)$ a {\em target chemical graph}.
This section  presents a set of rules for 
 specifying  topological substructure
  of a target chemical graph in a flexible way in Stage~4.

We first describe how to reduce a chemical graph $\C=(H,\alpha,\beta)$ into
an abstract form based on which our specification rules will be defined.
To illustrate the reduction process,
we use the chemical graph $\C=(H,\alpha,\beta)$
such that $\anC$ is given in Figure~\ref{fig:example_chemical_graph}.
 
\begin{figure}[h!] \begin{center}
\includegraphics[width=.65\columnwidth]{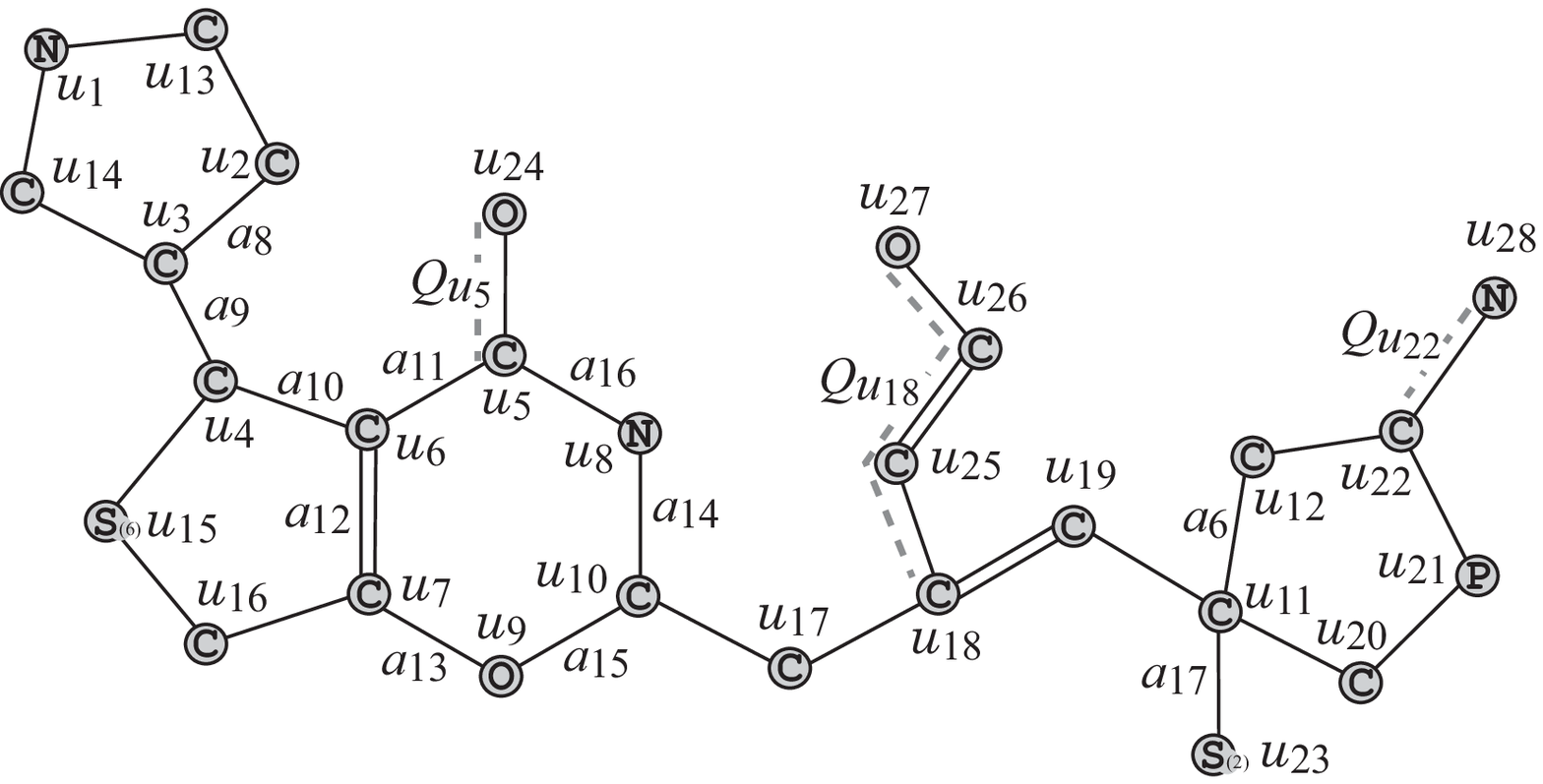}
\end{center} \caption{The interior $H^\inte$ of
chemical graph $\C$ with $\anC$ 
  in Figure~\ref{fig:example_chemical_graph} for ${\rho}=2$.
}
\label{fig:specification_example_interior} \end{figure}

\begin{figure}[h!] \begin{center}
\includegraphics[width=.98\columnwidth]{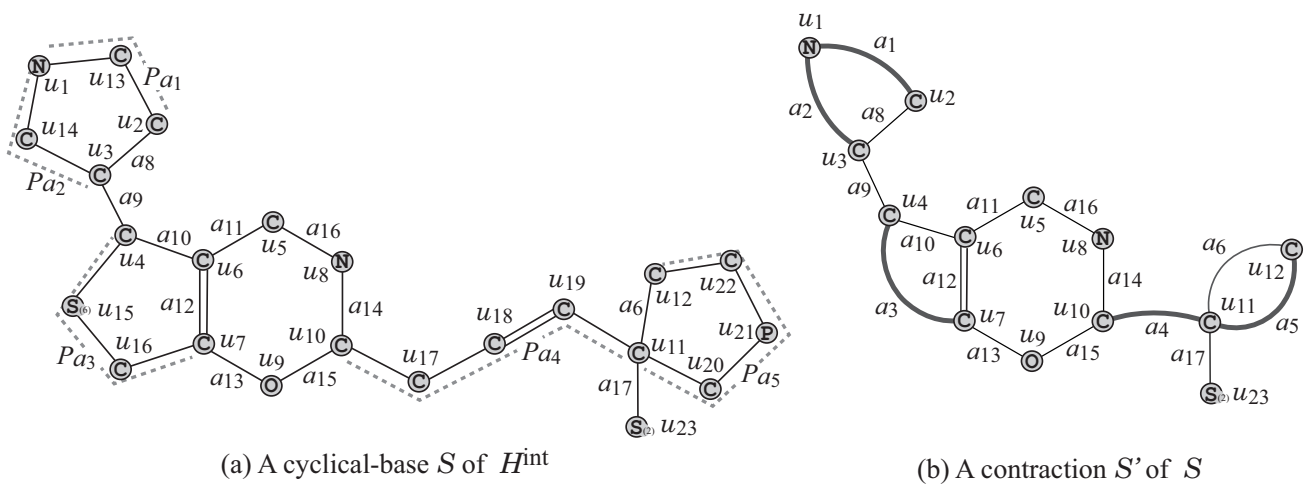}
\end{center} \caption{
(a) A cyclical-base  
$S=H^\inte- \bigcup_{u\in \{u_5,u_{18},u_{22}\}}(V(Q_u)\setminus \{u\})$
of the interior  $H^\inte$ in Figure~\ref{fig:specification_example_interior};
(b) A contraction $S'$ of  $S$ for a pure path set 
 $\mathcal{P}=\{P_{a_1},P_{a_2},\ldots,P_{a_5}\}$ 
in (a),
where a new edge obtained by contracting a pure path is depicted
with a thick line.   
}
\label{fig:specification_example_R2_3} \end{figure}

 \begin{enumerate}
 \item[R1] {\bf Removal of all ${\rho}$-fringe-trees: } 
The interior $H^\inte=(V^\inte(\C),E^\inte(\C))$ of $\C$ 
is obtained by removing the non-root vertices of 
each ${\rho}$-fringe-trees $\C[u]\in\mathcal{T}(\C), u\in V^\inte(\C)$. 
Figure~\ref{fig:specification_example_interior} illustrates
the interior $H^\inte$ of 
chemical graph $\C$ with ${\rho}=2$
  in Figure~\ref{fig:example_chemical_graph}. 
  
 \item[R2] {\bf Removal of some leaf paths: } 
 We call a $u,v$-path $Q$ in $H^\inte$  a {\em leaf path} if 
  vertex $v$ is a leaf-vertex of $H^\inte$
  and the degree of each internal vertex of $Q$  in $H^\inte$  is 2,
  where we regard that $Q$ is rooted at vertex $u$. 
A connected subgraph $S$ of the interior $H^\inte$ of $\C$  
is called a {\em cyclical-base}
if $S$ is obtained from $H$
by removing the vertices in $V(Q_u)\setminus \{u \}, u\in X$ 
for a subset $X$ of interior-vertices  and a set  $\{Q_u \mid u\in X\}$ of leaf 
 $u,v$-paths $Q_u$  such that    
 no two paths $Q_u$ and $Q_{u'}$ share a vertex.
Figure~\ref{fig:specification_example_R2_3}(a) illustrates
a cyclical-base  $S=H^\inte- \bigcup_{u\in X}(V(Q_u)\setminus \{u\})$
of the interior  $H^\inte$  
for a set 
$\{Q_{u_5}=(u_5,u_{24}), 
     Q_{u_{18}}=(u_{18},u_{25},u_{26},u_{27}),
     Q_{u_{22}}=(u_{22},u_{28})\}$ of leaf  paths 
in Figure~\ref{fig:specification_example_interior}.  

 \item[R3] {\bf Contraction of some pure paths: } 
 A path in $S$ is called {\em pure} 
 if  each internal vertex of the path  is of degree 2. 
 Choose a set $\mathcal{P}$ of several pure paths in $S$ 
 so that no two paths share  vertices except for their end-vertices. 
 A graph $S'$ is called a {\em contraction} of a graph $S$
  (with respect to $\mathcal{P}$) 
 if $S'$ is obtained from $S$ by replacing 
 each pure $u,v$-path  with a single edge $a=uv$,
 where $S'$ may contain multiple edges between the same pair of adjacent vertices.
Figure~\ref{fig:specification_example_R2_3}(b) illustrates
a contraction $S'$ obtained from 
the chemical graph  $S$
by contracting each $uv$-path $P_a\in  \mathcal{P}$ into a new edge $a=uv$,
where $a_1=u_1 u_{2},  a_2=u_1 u_{3},  a_3=u_4 u_{7}, a_4=u_{10}u_{11}$
and $a_5=u_{11}u_{12}$ and 
 $\mathcal{P}=\{
 P_{a_1}=(u_1,u_{13},u_{2}), 
 P_{a_2}=(u_{1},u_{14},u_{3}),
 P_{a_3}=(u_{4},u_{15},u_{16},u_{7}), 
 P_{a_4}=(u_{10},u_{17},u_{18},u_{19},u_{11}),
 P_{a_5}=(u_{11},u_{20},u_{21},u_{22},u_{12})\}$ of pure paths 
in Figure~\ref{fig:specification_example_R2_3}(a). 
\end{enumerate}

We will define a set of rules so that 
a chemical graph can be obtained 
from a graph (called a seed graph in the next section)
by applying processes R3 to R1 in a reverse way. 
We specify topological substructures of a target chemical graph
with a tuple  $(\GC,\sint,\sce)$  called  a {\em target specification}
defined under the set of the following rules. 

\subsection*{Seed Graph}

A  {\em seed graph} $\GC=(\VC,\EC)$ is defined
to be a graph (possibly with multiple edges) such that 
the edge set $\EC$ consists of four sets 
$\Et$, $\Ew$, $\Ez$ and $\Eew$, 
where each of them can be empty.
A seed graph plays a role of the most abstract form $S'$ in R3.  
Figure~\ref{fig:specification_example_1}(a) illustrates an example of a seed graph
$\GC$ with $\mathrm{r}(\GC)=5$,   
where $\VC=\{u_1,u_2,\ldots,u_{12},u_{23}\}$, 
$\Et=\{a_1,a_2,\ldots,a_5\}$, 
$\Ew=\{a_6\}$,
$\Ez=\{a_7\}$ and 
$\Eew=\{a_8,a_9,\ldots,a_{16}\}$.

 A {\em subdivision} $S$ of $\GC$  
is a graph constructed from a seed graph $\GC$ 
according to the following rules:
\begin{enumerate}[leftmargin=*]
\item[-]
Each edge $e=uv\in \Et$ is replaced
with a $u,v$-path $P_e$ of length at least 2;

\item[-] 
Each edge $e=uv\in \Ew$ is replaced
with a $u,v$-path $P_e$ of length at least 1
(equivalently $e$ is directly used or replaced with
a $u,v$-path $P_e$ of length at least 2);

\item[-] 
Each edge $e\in \Ez$ is either used or discarded, where 
 $\Ez$ is required to be chosen as a non-separating edge subset of
 $E(\GC)$ since otherwise the connectivity of a final chemical graph $\Co$
 is not guaranteed; 
$\mathrm{r}(\Co)= \mathrm{r}(\GC)-|E'|$ holds
for a subset $E'\subseteq \Ez$ of edges discarded 
in a  final chemical graph $\Co$; 
and 

\item[-]
Each edge $e\in \Eew$ is always used directly. 
\end{enumerate}

We allow a possible elimination of edges in $\Ez$ as an optional rule
in constructing a target chemical graph from a seed graph, 
even though such an operation has 
not been included in the process R3. 
A subdivision  $S$ plays a role of a cyclical-base   in R2. 
A target chemical graph $\C=(H,\alpha,\beta)$ will contain  $S$  as a subgraph
of the interior $H^\inte$ of $\C$.


\subsection*{Interior-specification}

A graph $H^*$ that serves as the interior $H^\inte$ of
a target chemical graph $\C$ will be constructed as follows.
First construct a subdivision  $S$ of a seed graph $\GC$ 
by replacing each edge $e=u u'\in \Et\cup\Ew$
with a pure $u,u'$-path $P_e$.
Next construct a supergraph $H^*$ of $S$ by 
attaching a leaf path $Q_v$ at each vertex $v\in \VC$ or
at an internal vertex $v\in V(P_e)\setminus\{u,u'\}$ 
of each pure $u,u'$-path $P_e$ for some edge $e=uu'\in \Et\cup\Ew$,
where possibly $Q_v=(v), E(Q_v)=\emptyset$ 
(i.e., we do not attach any new edges to $v$).
We introduce the following rules for specifying
 the size of $H^*$, the length $|E(P_e)|$  of
a pure path  $P_e$,  the length $|E(Q_v)|$ of
a   leaf path $Q_v$, the number of  leaf paths $Q_v$
and a bond-multiplicity of each interior-edge,
where we call the set of prescribed constants  
 an  {\em interior-specification}   $\sint$: 
\begin{enumerate}[leftmargin=*]
 \item[-]
  Lower and upper bounds $\nint_\LB, \nint_\UB\in \mathbb{Z}_+$ 
  on   the number of interior-vertices 
of a target chemical graph~$\C$. 
  
\item[-] 
For each edge $e=u u'\in \Et\cup\Ew$, 
\begin{description} 
\item[]
 a lower bound $\ell_{\LB}(e)$ and 
 an upper bound $\ell_{\UB}(e)$  on the length $|E(P_e)|$ of
 a pure $u,u'$-path $P_e$. 
(For a notational convenience, set 
$\ell_\LB(e):=0$, $\ell_\UB(e):=1$, $e\in \Ez$ and
$\ell_\LB(e):=1$, $\ell_\UB(e):=1$, $e\in \Eew$.)
   
\item[]  
 a lower bound $\bl_{\LB}(e)$ and 
 an upper bound $\bl_{\UB}(e)$ on the number of leaf paths $Q_v$ attached 
 at  internal vertices $v$ of a pure $u,u'$-path $P_e$.   

\item[] 
 a lower bound $\ch_{\LB}(e)$ and 
 an upper bound $\ch_{\UB}(e)$  on the maximum 
 length  $|E(Q_v)|$ of a leaf path $Q_v$ attached  
 at an internal vertex $v\in V(P_e)\setminus\{u,u'\}$ 
 of a pure $u,u'$-path $P_e$.   
\end{description} 

\item[-]
For each vertex $v\in \VC$, 
\begin{description} 
\item[]
 a lower bound $\ch_{\LB}(v)$ and 
 an upper bound $\ch_{\UB}(v)$  on  
 the number of leaf paths $Q_v$ attached to $v$,
 where $0\leq \ch_{\LB}(v)\leq \ch_{\UB}(v)\leq 1$.
 
\item[]
 a lower bound $\ch_{\LB}(v)$ and 
 an upper bound $\ch_{\UB}(v)$  on the
 length $|E(Q_v)|$ of a leaf path $Q_v$ attached to $v$. 
\end{description}  

\item[-] 
For each edge $e=u u'\in \EC$, 
a lower bound $\bd_{m, \LB}(e)$ 
and an  upper bound $\bd_{m, \UB}(e)$  on
the number of edges with bond-multiplicity $m\in [2,3]$ in
$u,u'$-path $P_e$, where we regard $P_e$, $e  \in \Ez\cup \Eew$ 
as single edge $e$.
\end{enumerate}

We call a graph $H^*$ that satisfies an interior-specification $\sint$
a {\em $\sint$-extension of $\GC$}, 
where the bond-multiplicity of each edge has been determined.

Table~\ref{table:interior-spec}  shows an example of 
an interior-specification  $\sint$ to the seed graph  $\GC$ in 
Figure~\ref{fig:specification_example_1}. 

\begin{table}[h!]\caption{Example~1 of an interior-specification  $\sint$. }
\begin{tabular}{ |  c | c |  } \hline 
$\nint_\LB=20$ & $\nint_\UB = 28$ \\\hline 
\end{tabular}

 \begin{tabular}{ |  c | c c c c c c |  } \hline
                        & $a_1$ &  $a_2$ &   $a_3$ &   $a_4$ &   $a_5$ &   $a_6$   \\\hline
 $\ell_\LB(a_i)$&  2 &  2 &  2 & 3 &  2 &  1 \\ \hline
 $\ell_\UB(a_i)$&  3 & 4 &  3 & 5 & 4 &  4 \\\hline
 $\bl_\LB(a_i)$&  0 &  0 &   0 & 1 &  1 &   0 \\ \hline
 $\bl_\UB(a_i)$&  1 & 1 &   0 & 2 & 1 &   0 \\\hline
 $\ch_\LB(a_i)$&  0 &  1 & 0 & 4 &  3 &  0 \\ \hline
 $\ch_\UB(a_i)$&  3 & 3 &  1 & 6 & 5 &  2 \\\hline
\end{tabular} 

\begin{tabular}{ |  c | c c c c c c   c c c c  c c c |  } \hline
                        & $u_1$ &  $u_2$ &   $u_3$ &   $u_4$ &   $u_5$ &   $u_6$ 
                       & $u_7$ &   $u_8$ &   $u_9$ &   $u_{10}$ &   $u_{11}$ 
                       &   $u_{12}$ &   $u_{23}$ \\\hline 
 $\bl_\LB(u_i)$&  0 &  0 &   0 & 0 &  0 &   0
                       & 0 &   0 &  0 &   0 &  0 &  0 &  0 \\ \hline
 $\bl_\UB(u_i)$&  1 & 1 &   1 & 1 & 1 &   0
                       & 0 &   0 &  0 &   0 &  0 &  0 &  0\\\hline
 $\ch_\LB(u_i)$&  0 &  0 &   0 & 0 &  1 &   0
                       & 0 &   0 &  0 &   0 &  0 &  0 &  0 \\ \hline
 $\ch_\UB(u_i)$&  1 & 0 &   0 & 0 & 3 &   0
                       & 1 &   1 &  0 &   1 &  2 & 4 &  1 \\\hline
\end{tabular} 

\begin{tabular}{ |  c | c c c c c c   c c c c c c  c c c c c |  } \hline
                               & $a_1$ &  $a_2$ &   $a_3$ &   $a_4$ &   $a_5$ &   $a_6$ 
                               & $a_7$ &  $a_8$ &   $a_9$ &   $a_{10}$ &   $a_{11}$ &   $a_{12}$ 
                               & $a_{13}$ &   $a_{14}$ &   $a_{15}$ &   $a_{16}$ &   $a_{17}$  \\\hline
 $\bd_{2, \LB}(a_i)$ &  0    &  0 &   0 & 1 &  0 &   0
                                &  0   &  0 &  0 & 0 &  0 &   1
                                &  0    &  0 &   0 & 0     & 0  \\ \hline
 $ \bd_{2, \UB}(a_i)$&  1    & 1 &   0 & 2  & 2 &   0  
                                &  0    & 0&   0 & 0 &  0 &   1
                                &  0    &  0 &   0 & 0   & 0   \\ \hline
 $\bd_{3, \LB}(a_i)$ &  0    &  0 &   0 & 0 &  0 &   0
                                &  0   &  0 &  0 & 0 &  0 &   0
                                &  0    &  0 &   0 & 0   & 0   \\ \hline
 $ \bd_{3, \UB}(a_i)$&  0    & 0 &   0 & 0  & 1 &   0 
                                &  0    &  0 &   0 & 0 &  0 &   0
                                &  0    &  0 &   0 &  0    & 0   \\ \hline
\end{tabular} 
\label{table:interior-spec}  
\end{table}

\begin{figure}[h!] \begin{center}
\includegraphics[width=.58\columnwidth]{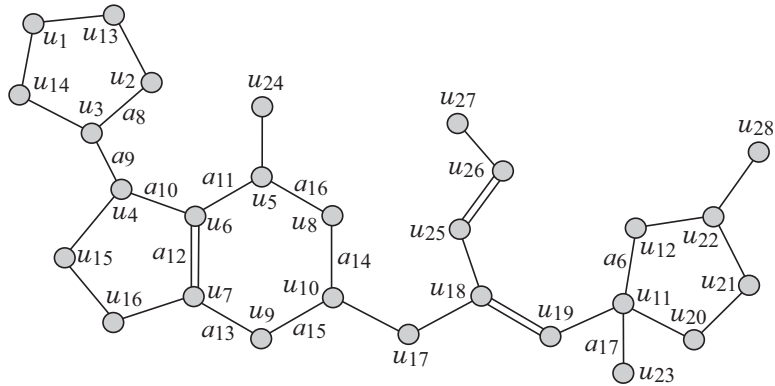}
\end{center} \caption{
An illustration of a graph 
$H^*$ that is obtained from  the seed graph  $\GC$ in 
Figure~\ref{fig:specification_example_1}
under the interior-specification  $\sint$ in 
Table~\ref{table:interior-spec},
where the vertices newly introduced by pure paths $P_{a_i}$
and leaf paths $Q_{v_i}$ are depicted with white squares and circles,
respectively.    }
\label{fig:specification_example_3} \end{figure}

Figure~\ref{fig:specification_example_3} illustrates an example of 
an $\sint$-extension $H^*$ of seed graph  $\GC$ in 
Figure~\ref{fig:specification_example_1}
under the interior-specification  $\sint$ in 
Table~\ref{table:interior-spec}.  

\subsection*{Chemical-specification}
 
 Let $H^*$ be a graph that serves as 
 the interior $H^\inte$ of a target chemical graph $\C$,
 where the bond-multiplicity of each edge in $H^*$ has be determined.
 Finally we introduce a set of rules for constructing 
   a target chemical graph $\C$ from $H^*$ 
   by choosing  a chemical element $\ta\in \Lambda$ 
and assigning a ${\rho}$-fringe-tree $\psi$
 to each interior-vertex $v\in V^\inte$. 
We introduce the following rules for specifying
the size of $\C$, a set of chemical rooted trees  
that are allowed to use as  ${\rho}$-fringe-trees 
and lower and upper bounds on the frequency of
a chemical element, a chemical symbol, 
and an edge-configuration,
where we call the set of prescribed constants   
 a  {\em chemical specification} $\sce$:   
 
\begin{enumerate}[leftmargin=*]
\item[-] 
Lower and upper bounds $n_\LB,  n^*\in \mathbb{Z}_+$
on the number of vertices, where $\nint_\LB \leq n_\LB\leq n^*$.
 
\item[-] 
Subsets  $\mathcal{F}(v) \subseteq \mathcal{F}(D_\pi), v\in \VC$ 
and $\mathcal{F}_E \subseteq \mathcal{F}(D_\pi)$ 
 of chemical rooted trees $\psi$ with $\h(\anpsi)\leq {\rho}$, where 
 we require that 
 every ${\rho}$-fringe-tree $\C[v]$ rooted at a vertex $v\in \VC$
 (resp., at an internal vertex $v$ not in $\VC$)   in  $\C$ 
 belongs to $\mathcal{F}(v)$ (resp.,   $\mathcal{F}_E$).  
Let  $\mathcal{F}^*:=\mathcal{F}_E\cup \bigcup_{v\in \VC}\mathcal{F}(v)$
and 
$\Lambda^\ex$ denote the set of  chemical elements assigned to non-root
vertices over all chemical rooted trees in $\mathcal{F}^*$.  
 
\item[-] 
A subset  $\Lambda^\inte\subseteq \Lambda^\inte(D_\pi)$, where 
 we require that every chemical element $\alpha(v)$ 
 assigned to an interior-vertex  $v$ in $\C$ belongs to $\Lambda^\inte$.
Let $\Lambda:= \Lambda^\inte\cup \Lambda^\ex$ and
 $\na_\ta(\C)$ (resp., $\na_\ta^\inte(\C)$ and $\na_\ta^\ex(\C)$) 
 denote the number of vertices   (resp.,   interior-vertices and  exterior-vertices)
  $v$ such that $\alpha(v)=\ta$   in  $\C$.
 
\item[-] 
A set $\Ldg^\inte\subseteq \Lambda\times [1,4]$  of chemical  symbols
and  a set $\Gamma^\inte \subseteq \Gamma^\inte(D_\pi)$  
of  edge-configurations  $(\mu,\mu' ,m)$ with $\mu \leq \mu'$, where 
 we require that the edge-configuration $\ec(e)$ of an interior-edge $e$ in $\C$ 
 belongs to $\Gamma^\inte$.
We do not distinguish  $(\mu,\mu' ,m)$ and $(\mu' , \mu,m)$.

\item[-] 
Define  $\Gac^\inte$ to be the set of   adjacency-configurations such that  
$\Gac^\inte:=\{(\ta, \tb, m) \mid (\ta d, \tb d',m)\in \Gamma^\inte\}$.   
Let  $\ac_\nu^\inte(\C), \nu\in \Gac^\inte$   
denote  the number of  interior-edges $e$ such that $\ac(e)=\nu$  in $\C$.
  
\item[-] 
 Subsets $\Lambda^*(v)\subseteq \{\ta\in \Lambda^\inte\mid \val(\ta)\geq 2\}$, 
 $v\in \VC$,  
 we require that every chemical element $\alpha(v)$ 
 assigned to   a vertex $v\in  \VC$
 in the seed graph  belongs to $\Lambda^*(v)$.  

\item[-] Lower and upper bound functions 
$\na_\LB,\na_\UB: \Lambda\to  [1,n^*]$  and 
$\na_\LB^\inte,\na_\UB^\inte: \Lambda^\inte\to  [1,n^*]$ 
on the number of   interior-vertices  $v$ such that  $\alpha(v)=\ta$  in $\C$. 

\item[-] Lower and upper bound functions  
$\ns_\LB^\inte,\ns_\UB^\inte: \Ldg^\inte\to  [1,n^*]$ 
  on the number of   interior-vertices $v$ such that $\cs(v)=\mu$  in $\C$.   

\item[-] Lower and upper bound functions  
$\ac_\LB^\inte,\ac_\UB^\inte: \Gac^\inte \to  \mathbb{Z}_+$ 
 on the number of  interior-edges $e$ such that $\ac(e)=\nu$  in $\C$. 

\item[-] Lower and upper bound functions  
$\ec_\LB^\inte,\ec_\UB^\inte: \Gamma^\inte \to  \mathbb{Z}_+$ 
 on the number of  interior-edges $e$ such that $\ec(e)=\gamma$  in $\C$.  
 
 \item[-] Lower and upper bound functions  
$\fc_\LB,\fc_\UB: \mathcal{F}^*\to  [0,n^*]$ 
  on the number of   interior-vertices $v$ 
  such that $\C[v]$ is r-isomorphic to $\psi\in \mathcal{F}^*$  in $\C$.   
  
 \item[-] Lower and upper bound functions  
$\ac^\lf_\LB,\ac^\lf_\UB: \Gac^\lf \to  [0,n^*]$ 
  on the number of  leaf-edges $uv$ in $\acC$
  with adjacency-configuration $\nu$.  
\end{enumerate}
 
We call a chemical graph $\C$ that satisfies a chemical specification $\sce$
a {\em $(\sint,\sce)$-extension of $\GC$},
and denote by $\mathcal{G}(\GC, \sint,\sce)$ the set of
all $(\sint,\sce)$-extensions of $\GC$. 

Table~\ref{table:chemical_spec}  shows an example of 
a chemical-specification  $\sce$ to the seed graph  $\GC$
 in Figure~\ref{fig:specification_example_1}. 
 

\begin{table}[h!]\caption{Example~2 of a chemical-specification  $\sce$.  
}
\begin{tabular}{ |  l |  } \hline
 $n_\LB=30$,  $n^* =50$. \\\hline
  branch-parameter:   ${\rho}=2$  \\\hline
\end{tabular}

\begin{tabular}{ |  l |  } \hline
 Each of sets $\mathcal{F}(v), v\in \VC$ and
 $\mathcal{F}_E$ is set to be \\
 the set $\mathcal{F}$  of chemical rooted trees $\psi$ with $\h(\anpsi)\leq {\rho}=2$
in Figure~\ref{fig:specification_example_1}(b). \\\hline
\end{tabular}

\begin{tabular}{ |  c | c |   } \hline
  $\Lambda=\{ \ttH,\ttC,\ttN,\ttO, \ttS_{(2)},\ttS_{(6)}, \ttP=\ttP_{(5)}\}$ & 
  $\Ldg^{\inte} =\{ \ttC2 , \ttC3,  \ttC4, \ttN2, \ttN3, \ttO2,
    \ttS_{(2)}2,  \ttS_{(6)}3, \ttP4   \}$  
\\\hline
\end{tabular}

\begin{tabular}{ |  c | l |  } \hline
  $\Gac^{\inte}$ &
  $ \nu_1 \!=\!(\ttC   , \ttC  , 1) ,   \nu_2 \!=\!(\ttC   , \ttC  , 2) ,   
   \nu_3 \!=\!(\ttC   , \ttN  , 1) ,  \nu_4 \!=\!(\ttC  , \ttO  , 1), 
    \nu_5 \!=\! (\ttC, \ttS_{(2)}, 1),\nu_6 \!=\!(\ttC  , \ttS_{(6)}, 1), 
    \nu_7 \!=\! (\ttC  , \ttP  , 1) $  \\ \hline
\end{tabular}

\begin{tabular}{ |  c | l |  } \hline
  $\Gamma^{\inte}$ &
  $ \gamma_1 \!=\! (\ttC 2 , \ttC 2, 1) ,
   \gamma_2 \!=\!(\ttC 2 , \ttC 3, 1) ,  
   \gamma_3 \!=\!(\ttC 2 , \ttC 3, 2) ,  
   \gamma_4 \!=\!(\ttC 2 , \ttC 4, 1) , 
   \gamma_5 \!=\!(\ttC 3 , \ttC 3, 1) , 
   \gamma_6 \!=\!(\ttC 3 , \ttC 3, 2) , $ \\
   &
  $   
    \gamma_7 \!=\!(\ttC 3 , \ttC 4, 1), 
   \gamma_8 \!=\!(\ttC 2 , \ttN 2, 1) ,  
   \gamma_9 \!=\!(\ttC 3 , \ttN 2, 1) ,  
   \gamma_{10} \!=\!(\ttC 3 , \ttO 2, 1), 
    \gamma_{11} \!=\!(\ttC 2 , \ttC 2, 2),  
    \gamma_{12} \!=\!(\ttC 2 , \ttO 2, 1) ,$ \\
   &
  $  
    \gamma_{13} \!=\!(\ttC 3 , \ttN3, 1), 
    \gamma_{14} \!=\!(\ttC 4, \ttS_{(2)} 2, 2),  
    \gamma_{15} \!=\!(\ttC 2 , \ttS_{(6)}3, 1), 
   \gamma_{16} \!=\!(\ttC 3 , \ttS_{\tiny (6)}3, 1), 
    \gamma_{17} \!=\!(\ttC 2, \ttP4, 2), $ \\
   &
  $  
    \gamma_{18} \!=\!(\ttC 3, \ttP4, 1)  
     $ \\ \hline
\end{tabular}
    
\begin{tabular}{ |  l|  } \hline
$\Lambda^*(u_1)=\Lambda^*(u_8)=\{{\tt C,  N}\}$, 
$\Lambda^*(u_9)=\{{\tt C, O}\}$, 
   $\Lambda^*(u)=\{\ttC\}$, $u\in \VC\setminus\{u_1,u_8,u_9\}$
   \\\hline
\end{tabular}

\begin{tabular}{ |  c | c c c c  c c c |  } \hline
                         & ${\tt H}$  & ${\tt C}$ &   ${\tt N}$ &     ${\tt O}$ 
                         & $\ttS_{(2)}$ & $\ttS_{(6)}$ & $\ttP$  \\\hline
 $\na_\LB(\ta)$ & 40 &  27 &  1 &   1 & 0 & 0 & 0   \\ \hline 
 $\na_\UB(\ta)$ & 65 & 37 & 4 &  8  &   1 &   1 &   1 \\\hline
\end{tabular} 
\begin{tabular}{ |  c | c c c  c c c   |  } \hline
   & $\ttC$ &   $\ttN$ &     $\ttO$  & $\ttS_{(2)}$ & $\ttS_{(6)}$ & $\ttP$  \\\hline
 $\na_\LB^{\inte}(\ta)$ &   9 &  1 &   0  & 0 & 0 & 0      \\ \hline
 $\na_\UB^{\inte}(\ta) $&  23 & 4 & 5 &   1 &   1 &   1  \\\hline
\end{tabular} 

\begin{tabular}{ |  c | c c c c c c  c c c   |  } \hline
    & $\ttC2$ &  $\ttC3$ &   $\ttC4$ & $\ttN2$ &   $\ttN3$ &   $\ttO2$
   & $\ttS_{(2)}2$ & $\ttS_{(6)}3$ & $\ttP4$  \\\hline
 $\ns_\LB^{\inte}(\mu)$ &  3 &  5 &   0 & 0 &  0 &   0 & 0 &  0 &   0    \\ \hline
 $\ns_\UB^{\inte}(\mu) $&  8 & 15 & 2 & 2 & 3 &  5  &   1 &   1 &   1   \\\hline
\end{tabular} 

\begin{tabular}{ |  c | c c c c c c c |  } \hline
         & $\nu_1 $ &   $\nu_2 $ & $\nu_3 $   & $\nu_4 $
         &   $\nu_5 $ & $\nu_6 $   & $\nu_7 $ \\\hline
 $\ac_\LB^{\inte}(\nu)$  &  0  &  0  & 0  & 0   & 0  & 0 & 0     \\ \hline
 $\ac_\UB^{\inte}(\nu)$ &  30 & 10 & 10 & 10 & 1 & 1 & 1 \\\hline
\end{tabular} 

\begin{tabular}{ |  c | c c c c c c c c c c c c c c c c c c |  } \hline
    & $\gamma_1 $ &   $\gamma_2 $ & $\gamma_3 $   & $\gamma_4 $ 
     & $\gamma_5 $
    & $\gamma_6 $ &   $\gamma_7 $ & $\gamma_8 $   & $\gamma_9 $ 
     & $\gamma_{10} $   & $\gamma_{11} $       & $\gamma_{12} $          
     & $\gamma_{13} $   & $\gamma_{14} $       & $\gamma_{15} $          
     & $\gamma_{16} $   & $\gamma_{17} $       & $\gamma_{18} $          
                            \\\hline
 $\ec_\LB^{\inte}(\gamma)$ &  0 &  0 & 0 &  0  & 0 &  0 &  0 & 0 &  0  & 0 & 0 & 0 
    &  0 & 0 &  0  & 0 & 0 & 0  \\ \hline
 $\ec_\UB^{\inte}(\gamma) $& 4 & 15 & 4 &  4  & 10 &  5 & 4 & 4 &  6 & 4 & 4 & 4
 &  2 & 2 &  2  & 2 & 2 & 2  \\\hline
\end{tabular}

\begin{tabular}{ |  c | c   c   |  } \hline 
& $\psi\in\{\psi_i\mid i=1,6,11\}$ 
& $\psi\in \mathcal{F}^*\setminus \{\psi_i\mid i=1,6,11\}$ \\\hline
 $\fc_\LB(\psi)$  &  1 &    0   \\ \hline 
 $\fc_\UB(\psi)$ &  10 &  3\\\hline
\end{tabular}

\begin{tabular}{ |  c | c   c   |  } \hline 
& $\nu\in\{(\ttC,\ttC,1),(\ttC,\ttC,2)  \}$ 
& $\nu\in \Gac^\lf \setminus \{(\ttC,\ttC,1),(\ttC,\ttC,2)  \}$   \\\hline
 $\ac^\lf_\LB(\nu)$  &  0 &    0   \\ \hline 
 $\ac^\lf_\UB(\nu)$ &  10 &  8 \\\hline
\end{tabular} 

\label{table:chemical_spec}
\end{table}

Figure~\ref{fig:example_chemical_graph} 
 illustrates an example $\Co$ of 
a   $(\sint,\sce)$-extension of $\GC$   obtained 
from the  $\sint$-extension $H^*$  
 in Figure~\ref{fig:specification_example_3} 
under the chemical-specification $\sce$ in Table~\ref{table:chemical_spec}.  
Note that $\mathrm{r}(\Co)= \mathrm{r}(H^*)= \mathrm{r}(\GC)-1=4$
 holds since the edge in $\Ez$ is discarded in $H^*$.

\clearpage

\section{Test Instances for Stages~4 and 5}\label{sec:test_instances} 

We prepared the following instances (a)-(d) for conducting experiments
of  Stages~4  and 5 in Phase~2. 
 
 In Stages~4 and 5, we  use five properties 
 $\pi\in \{${\sc Hc}, {\sc Vd}, {\sc OptR}, {\sc IhcLiq}, {\sc Vis}$\}$
 and define a set $\Lambda(\pi)$ of chemical elements as follows: \\
 ~~~ $\Lambda(${\sc Hc}$)=
 \{\ttH,\ttC,\ttN,\ttO,\ttS_{(2)}, \ttS_{(6)},\ttCl \}$,
 ~~~ $\Lambda(${\sc Vd}$)=
 \{\ttH,\ttC,\ttN,\ttO,\ttN,\ttCl,\ttP_{(3)},\ttP_{(5)}\}$, \\
 ~~~ $\Lambda(${\sc OptR}$)=
 \{\ttH,\ttC,\ttN,\ttO,\ttS_{(2)},\ttF\}$,
 ~~~ $\Lambda(${\sc IhcLiq}$)=
 \{\ttH,\ttC,\ttN,\ttO,\ttS_{(2)}, \ttS_{(6)},\ttCl\}$ and \\
~~~  $\Lambda(${\sc Vis}$)= 
 \{\ttH,\ttC,\ttO,\ttSi\}$. 
 
\begin{itemize} 
  \item[(a)]  $I_{\mathrm{a}} =(\GC,\sint,\sce)$: The instance
  introduced in Appendix~\ref{sec:specification} to explain the target specification.
For each property $\pi$, we replace
 $\Lambda=\{ \ttH,\ttC,\ttN,\ttO, \ttS_{(2)},\ttS_{(6)},\ttP_{(5)}\}$
in Table~\ref{table:chemical_spec} 
 with $\Lambda(\pi)\cap \{\ttS_{(2)},\ttS_{(6)},\ttP_{(5)}\}$
 and  remove from the $\sce$
 all chemical symbols,  edge-configurations and fringe-configurations
  that cannot be constructed from the replaced element set 
 (i.e., those containing a chemical element in 
 $\{\ttS_{(2)},\ttS_{(6)}, \ttP_{(5)}\}\setminus \Lambda(\pi)$).
 
 \end{itemize}
  
\begin{itemize} 
  \item[(b)] $I_\mathrm{b}^i=(\GC^i,\sint^i, \sce^i)$, $i=1,2,3,4$:
 An instance for inferring chemical graphs with rank at most 2.  
In the four instances $I_\mathrm{b}^i$, $i=1,2,3,4$, 
the following specifications in $(\sint,\sce)$ are common. 
\begin{enumerate}
\item[] 
Set  $\Lambda:=\Lambda(\pi)$
 for a given property $\pi\in \{${\sc Hc, Vd, OptR,  IhcLiq, Vis}$\}$, 
 set $\Ldg^\inte$ to be
the set of all possible symbols in $\Lambda\times[1,4]$  
that appear in the data set $D_\pi$  
and set $\Gamma^\inte$
to be the set  of  all  edge-configurations that appear in the data set $D_\pi$. 
Set  $\Lambda^*(v):= \Lambda$,  $v\in \VC$. 
 
\item[] 
The lower bounds  
 $\ell_\LB $, $\bl_\LB $, $\ch_\LB $,  
 $\bd_{2,\LB}$,   $\bd_{3,\LB}$,  
 $\na_\LB$,  $\na^\inte_\LB$,  $\ns^\inte_\LB$,  
$\ac^\inte_\LB$, $\ec^\inte_\LB$ and $\ac^\lf_\LB$  are all set to be 0.

\item[] 
The upper bounds  
 $\ell_\UB $, $\bl_\UB $, $\ch_\UB $,  
 $\bd_{2,\UB}$,   $\bd_{3,\UB}$,  
 $\na_\UB$,  $\na^\inte_\UB$,  $\ns^\inte_\UB$,  
$\ac^\inte_\UB$, $\ec^\inte_\UB$ and $\ac^\lf_\UB$ 
are all set to be an upper bound $n^*$  on $n(G^*)$.
 
\item[] 
   For each property $\pi$, let $\mathcal{F}(D_\pi)$ denote
    the set of 2-fringe-trees in the compounds in $D_\pi$,
   and select a subset $\mathcal{F}_\pi^i\subseteq  \mathcal{F}(D_\pi)$ with
   $|\mathcal{F}_\pi^i|=45-5i$, $i\in [1,5]$.
   For each instance $I_\mathrm{b}^i$, 
   set $\mathcal{F}_E :=\mathcal{F}(v):=  \mathcal{F}_\pi^i$, $v\in \VC$ and 
$\fc_\LB(\psi):=0, \fc_\UB(\psi):=10, \psi\in  \mathcal{F}_\pi^i$. 
\end{enumerate}
 
  Instance $I_\mathrm{b}^1$ is given   by the rank-1 seed graph $\GC^1$ 
  in Figure~\ref{fig:specification_example_polymer}(i)
  and   Instances $I_\mathrm{b}^i$, $i=2,3,4$ are
   given by  the rank-2 seed graph $\GC^i$, $i=2,3,4$ in 
   Figure~\ref{fig:specification_example_polymer}(ii)-(iv).

\begin{itemize} 
 \item[(i)]  For instance $I_\mathrm{b}^1$, select as a seed graph 
  the monocyclic graph   $\GC^1=(\VC,\EC=\Et\cup \Ew)$
  in Figure~\ref{fig:specification_example_polymer}(i),
  where $\VC=\{u_1,u_2\}$, $\Et=\{a_1\}$ and  $ \Ew=\{a_2\}$. 
Set $\nint_\LB:=5, \nint_\UB:=15, n_\LB:=35$ and $n^*:=38$.
We  include a linear constraint 
$\ell(a_1)\leq \ell(a_2)$ 
and $5\leq \ell(a_1)+\ell(a_2) \leq 15$  as part of the side constraint. 
  
 \item[(ii)]
 For instance $I_\mathrm{b}^2$, select as a seed graph 
  the  graph   $\GC^2=(\VC,\EC=\Et\cup \Ew\cup \Eew)$ 
  in Figure~\ref{fig:specification_example_polymer}(ii),
  where
$\VC=\{u_1,u_2,u_3,u_4\}$, 
$\Et=\{a_1,a_2\}$, 
$\Ew=\{a_3\}$  and 
$\Eew=\{a_4,a_5\}$. 
Set $\nint_\LB:=25, \nint_\UB:=30, n_\LB:=45$ and $n^*:=50$. 
We include a linear constraint $\ell(a_1)\leq \ell(a_2)$ 
and $\ell(a_1)+\ell(a_2)+\ell(a_3)\leq 15$. 
    
 \item[(iii)]
 For instance $I_\mathrm{b}^3$, select as a seed graph 
  the  graph   $\GC^3=(\VC,\EC=\Et\cup \Ew\cup \Eew)$ 
  in Figure~\ref{fig:specification_example_polymer}(iii),   where
$\VC=\{u_1,u_2,u_3,u_4\}$, 
$\Et=\{a_1\}$, 
$\Ew=\{a_2, a_3\}$  and 
$\Eew=\{a_4,a_5\}$. 
Set $\nint_\LB:=25, \nint_\UB:=30, n_\LB:=45$ and $n^*:=50$. 
We include   linear constraints 
$\ell(a_1)\leq \ell(a_2)+\ell(a_3)$, $\ell(a_2)\leq \ell(a_3)$
and $\ell(a_1)+\ell(a_2)+\ell(a_3)\leq 15$.  

 \item[(iv)] 
 For instance $I_\mathrm{b}^4$, select as a seed graph 
  the  graph   $\GC^4=(\VC,\EC=\Et\cup \Ew\cup \Eew)$ 
  in Figure~\ref{fig:specification_example_polymer}(iv),   where
$\VC=\{u_1,u_2,u_3,u_4\}$, 
$\Ew=\{a_1, a_2, a_3\}$  and 
$\Eew=\{a_4,a_5\}$. 
Set $\nint_\LB:=25, \nint_\UB:=30, n_\LB:=45$ and $n^*:=50$. 
We   include   linear constraints 
$\ell(a_2)\leq \ell(a_1)+1$,
$\ell(a_2)\leq \ell(a_3)+1$,  $\ell(a_1)\leq \ell(a_3)$  
and $\ell(a_1)+\ell(a_2)+\ell(a_3)\leq 15$. 
 \end{itemize}
 \end{itemize}

 
 We define instances in (c) and (d) 
 in order to find chemical graphs that have an intermediate structure of
 given two chemical cyclic graphs 
 $G_A=(H_A=(V_A,E_A),\alpha_A,\beta_A)$ 
and $G_B=(H_B=(V_B,E_B),\alpha_B,\beta_B)$.
Let
 $\Lambda_A^\inte$ and  $\Lambda_{\mathrm{dg},A}^\inte$ 
 denote the sets  of chemical elements
 and chemical symbols  of
 the interior-vertices in $G_A$, 
 $\Gamma_A^\inte$   denote the sets of edge-configurations of
  the interior-edges in $G_A$,   
  and 
  $\mathcal{F}_A$ denote the set of 2-fringe-trees in $G_A$.  
Analogously define sets
 $\Lambda_B^\inte$,    $\Lambda_{\mathrm{dg},B}^\inte$,   
 $\Gamma_B^\inte$ and   $\mathcal{F}_B$ 
 in $G_B$.

\begin{itemize}  
\item[(c)]  $I_{\mathrm{c}}=(\GC,\sint,\sce)$: 
An instance aimed to infer a chemical graph $G^\dagger$ such that
the core of $G^\dagger$ is equal to the core of $G_A$ and 
the frequency of each edge-configuration in the non-core of $G^\dagger$
is equal to that of  $G_B$. 
We use chemical compounds CID~24822711 and CID~59170444 in 
 Figure~\ref{fig:instance_I_c_I_d}(a) and (b)
 for $G_A$ and $G_B$, respectively.  \\
Set   a seed graph $\GC=(\VC,\EC=\Eew)$ to be the core of $G_A$. \\
Set  $\Lambda:=\{{\tt H,C,N,O}\}$, 
and  set $\Ldg^\inte$ to be
the set of all possible chemical symbols in $\Lambda\times[1,4]$.\\
Set 
$\Gamma^\inte:=\Gamma_A^\inte\cup \Gamma_B^\inte$ and 
  $\Lambda^*(v):=\{\alpha_A(v)\}$, $v\in \VC$.  \\
Set 
$\nint_\LB:=\min\{\nint(G_A), \nint(G_B)\}$, 
$\nint_\UB:=\max\{\nint(G_A), \nint(G_B)\}$, \\
$n_\LB:=\min\{n(G_A), n(G_B)\}-10$ 
and   $n^*:=\max\{n(G_A), n(G_B)\}+5$. \\
Set  lower bounds  
 $\ell_\LB $, $\bl_\LB $, $\ch_\LB $,  
 $\bd_{2,\LB}$,   $\bd_{3,\LB}$,  
 $\na_\LB$,  $\na^\inte_\LB$,  $\ns^\inte_\LB$, 
$\ac^\inte_\LB$ and  $\ac^\lf_\LB$  to be 0.\\
Set  upper bounds  
 $\ell_\UB $, $\bl_\UB $, $\ch_\UB $,  
 $\bd_{2,\UB}$,   $\bd_{3,\UB}$,  
 $\na_\UB$,  $\na^\inte_\UB$,  $\ns^\inte_\UB$, 
$\ac^\inte_\UB$   and  $\ac^\lf_\UB$ to be  $n^*$. \\
Set $\ec_\LB^\inte(\gamma)$ 
to be the number of core-edges  in $G_A$ with $\gamma\in \Gamma^\inte$ and  
 $\ec_\UB^\inte(\gamma)$  
to be the number interior-edges in $G_A$ and  $G_B$ 
with edge-configuration $\gamma$. \\
Let $\mathcal{F}_B^{(p)}, p\in [1,2]$ denote the set of chemical rooted 
trees r-isomorphic $p$-fringe-trees in $G_B$; \\
Set $\mathcal{F}_E :=\mathcal{F}(v):= 
 \mathcal{F}_B^{(1)}\cup \mathcal{F}_B^{(2)}$, $v\in \VC$ and
$\fc_\LB(\psi):=0, \fc_\UB(\psi):=10, \psi\in \mathcal{F}_B^{(1)}\cup \mathcal{F}_B^{(2)}$. 
 
  \item[(d)] $I_{\mathrm{d}}=(\GC^1,\sint, \sce)$:     
An instance aimed to infer a chemical monocyclic graph $G^\dagger$ such that
the frequency vector of  edge-configurations in  $G^\dagger$
is a vector obtained by merging those of $G_A$ and $G_B$.
We use chemical monocyclic compounds CID~10076784 and CID~44340250
in   Figure~\ref{fig:instance_I_c_I_d}(c) and (d) 
 for $G_A$ and $G_B$, respectively.  
Set a seed graph to be   the monocyclic seed graph  
 $\GC^1=(\VC,\EC=\Et\cup \Ew)$ with 
  $\VC=\{u_1,u_2\}$, $\Et=\{a_1\}$ and  $ \Ew=\{a_2\}$ 
  in Figure~\ref{fig:specification_example_polymer}(i). \\
Set  $\Lambda:=\{{\tt H,C,N,O}\}$,  
 $\Ldg^\inte:=\Lambda_{\mathrm{dg},A}^\inte 
                 \cup \Lambda_{\mathrm{dg},B}^\inte$ and 
$\Gamma^\inte:=\Gamma_A^\inte\cup \Gamma_B^\inte$. \\
Set 
$\nint_\LB:=\min\{\nint(G_A), \nint(G_B)\}$, 
$\nint_\UB:=\max\{\nint(G_A), \nint(G_B)\}$, \\
  $n_\LB:=\min\{n(G_A),n(G_B)\}$ and  
  $n^*:=\max\{n(G_A),n(G_B)\}$. \\
Set  lower bounds  
 $\ell_\LB $, $\bl_\LB $, $\ch_\LB $,  
 $\bd_{2,\LB}$,   $\bd_{3,\LB}$,  
 $\na_\LB$,  $\na^\inte_\LB$,  $\ns^\inte_\LB$, 
$\ac^\inte_\LB$  and  $\ac^\lf_\LB$ to be 0.\\
Set  upper bounds  
 $\ell_\UB $, $\bl_\UB $, $\ch_\UB $,  
 $\bd_{2,\UB}$,   $\bd_{3,\UB}$,  
 $\na_\UB$,  $\na^\inte_\UB$,  $\ns^\inte_\UB$,
$\ac^\inte_\UB$ and  $\ac^\lf_\UB$  to be   $n^*$. \\
For each edge-configuration
 $\gamma \in \Gamma^\inte$,  
let  $\x^*_A(\gamma^\inte)$  (resp., $\x^*_B(\gamma^\inte)$)   denote
 the number of interior-edges with $\gamma$ in $G_A$ (resp., $G_B$), 
 $\gamma \in \Gamma^\inte$ and   
set \\
$\x^*_{\min}(\gamma):=\min\{\x^*_A(\gamma), \x^*_B(\gamma)\}$, 
 $\x^*_{\max}(\gamma):=\max\{\x^*_A(\gamma), \x^*_B(\gamma)\}$, \\
$\ec_\LB^\inte(\gamma):=
\lfloor (3/4)\x^*_{\min}(\gamma)+(1/4)\x^*_{\max}(\gamma) \rfloor$
and  \\
$\ec_\UB^\inte(\gamma):=
\lceil (1/4)\x^*_{\min}(\gamma)+(3/4)\x^*_{\max}(\gamma) \rceil$. \\
Set $\mathcal{F}_E :=\mathcal{F}(v):=  \mathcal{F}_A\cup \mathcal{F}_B$, 
$v\in \VC$ and 
$\fc_\LB(\psi):=0, \fc_\UB(\psi):=10, \psi\in \mathcal{F}_A\cup \mathcal{F}_B$. \\
We  include a linear constraint 
$\ell(a_1)\leq \ell(a_2)$ 
and $5\leq \ell(a_1)+\ell(a_2) \leq 15$  as part of the side constraint. 
 \end{itemize}


\clearpage

\section{All Constraints in an MILP Formulation for Chemical Graphs}\label{sec:full_milp}


We define a standard encoding of a finite set $A$ of elements
to be a bijection $\sigma: A \to [1, |A|]$, 
where we denote by $[A]$   the set $[1, |A|]$ of integers
and by $[{\tt e}]$ the encoded element $\sigma({\tt e})$.
Let $\epsilon$ denote {\em null}, a fictitious chemical element 
that does not belong to any set of chemical elements,
chemical symbols, adjacency-configurations and
edge-configurations in the following formulation.
Given a finite set $A$, let $A_\epsilon$ denote the set $A\cup\{\epsilon\}$
and define a standard encoding of $A_\epsilon$
  to be a bijection $\sigma: A \to [0, |A|]$ such that
$\sigma(\epsilon)=0$, 
where we denote by $[A_\epsilon]$   the set $[0, |A|]$ of integers
and by $[{\tt e}]$ the encoded element $\sigma({\tt e})$,
where $[\epsilon]=0$.

 \bigskip 
 Let $\sigma=(\GC,\sint,\sce)$ be a target specification,
 ${\rho}$ denote  the branch-parameter in the specification $\sigma$
 and  $\C$ denote a chemical   graph in $\mathcal{G}(\GC, \sint,\sce)$. 

 \subsection{Selecting  a Cyclical-base} 
\label{sec:co}
 
Recall that  
\[ \begin{array}{ll}
   \Eew = \{e\in \EC\mid \ell_\LB(e)=\ell_\UB(e)=1 \}; &
   \Ez =\{e\in \EC\mid \ell_\LB(e)=0, \ell_\UB(e)=1 \}; \\
  \Ew=\{e\in \EC\mid \ell_\LB(e)=1,  \ell_\UB(e)\geq 2 \}; &
  \Et= \{e\in \EC\mid \ell_\LB(e)\geq 2 \}; \end{array} \]
\begin{enumerate} [leftmargin=*]
\item[-]
Every edge $a_i\in \Eew$ is  included in  $\anC$;

\item[-]
Each edge $a_i\in \Ez$ is   included in $\anC$ if necessary;
 
\item[-]
For each edge  $a_i  \in \Et$, edge $a_i$ is not included in $\anC$
and instead a path 
\[P_i=(\vC_{\tail(i)}, \vT_{j-1},\vT_{j},\ldots,
    \vT_{j+t}, \vC_{\hd(i)})\]
     of length at least 2
  from vertex $\vC_{\tail(i)}$ to vertex $\vC_{\hd(i)}$ 
  visiting some  vertices in $\VT$ is constructed in $\anC$; and  
 
\item[-]
For each edge $a_i  \in \Ew$, either  edge $a_i$   is directly used in $\anC$ or
the above path $P_i$ of length at least 2   is constructed in $\anC$.  
 \end{enumerate}
 
Let  $\tC\triangleq |\VC|$ and denote $\VC$ by 
$\{\vC_{i}\mid i\in [1,\tC]\}$.
Regard the seed graph $\GC$ as a digraph such that
each edge $a_i$ with end-vertices $\vC_{j}$ and $\vC_{j'}$
is directed from  $\vC_{j}$ to $\vC_{j'}$ when $j<j'$.
 For each directed edge $a_i  \in \EC $,
 let $\hd(i)$ and $\tail(i)$ denote the head and tail of $\eC(i)$;
 i.e., $a_i=(\vC_{\tail(i)}, \vC_{\hd(i)})$. 
  
Define 
 \[ \kC \triangleq  |\Et\cup \Ew| , ~~ \widetilde{\kC} \triangleq  |\Et| ,\]
 and denote   $\EC=\{a_i\mid i\in[1,\mC]\}$,
$\Et=\{a_k\mid k\in[1,\widetilde{\kC}]\}$,
$\Ew=\{a_k\mid k\in[\widetilde{\kC}+1,\kC]\}$,
$\Ez=\{a_i\mid i\in[\kC+1,\kC+|\Ez|]\}$ and 
$\Eew=\{a_i\mid i\in[\kC+|\Ez|+1,\mC]\}$.
Let $\Iew$ denote the set of indices $i$ of edges $a_i\in \Eew$.
Similarly for $\Iz$, $\Iw$  and $\It$.

To control the construction of such a path $P_i$
 for each edge  $a_k\in  \Et\cup \Ew $,
we regard the index $k\in [1,\kC]$ of each edge $a_k\in  \Et\cup \Ew$
as the ``color'' of the edge.
To introduce necessary linear constraints 
that can construct such a path $P_k$ properly   in our MILP,
we assign the color $k$ to the vertices $\vT_{j-1},\vT_{j},\ldots,$ 
$\vT_{j+t}$ in $\VT$
when the above path  $P_k$ is used in $\anC$.
 
For each index $s\in [1,\tC]$, let  
$\IC(s)$ denote the set of edges $e\in \EC$ incident to vertex $\vC_{s}$,
and 
 $\Eew^+(s)$ (resp., $\Eew^-(s)$) denote the set of 
 edges $a_i\in \Eew$ such that 
the tail (resp., head) of $a_i$ is vertex $\vC_{s}$.
Similarly for 
$\Ez^+(s)$,  $\Ez^-(s)$, $\Ew^+(s)$,  $\Ew^-(s)$,
$\Et^+(s)$ and $\Et^-(s)$.
Let $\IC(s)$ denote the set of indices $i$ of edges $a_i\in \IC(s)$.
Similarly for   
$\Iew^+(s)$,  $\Iew^-(s)$,
$\Iz^+(s)$,  $\Iz^-(s)$, 
$\Iw^+(s)$,  $\Iw^-(s)$,
$\It^+(s)$ and $\It^-(s)$.
Note that $[1, \kC]=\It\cup \Iw$ and 
$[\widetilde{\kC}+1,\mC]=\Iw\cup \Iz\cup\Iew$.

\smallskip\noindent
{\bf constants: } 
\begin{enumerate} [leftmargin=*]
\item[-] $\tC=|\VC|$, $\widetilde{\kC}=  |\Et|$, $\kC= |\Et\cup \Ew|$,
      $\tT=\nint_\UB-|\VC|$, $\mC=|\EC|$.
      Note that 
      $a_i\in \EC\setminus (\Et\cup \Ew)$ holds $i\in [\kC+1,\mC]$;   

\item[-] $\ell_\LB(k), \ell_\UB(k)\in [1, \tT]$, $k\in [1,\kC]$: 
lower and upper bounds on the length of path $P_k$;  
       
\item[-] $r_{\GC}\in[1,\mC]$: the rank $\mathrm{r}(\GC)$ of
seed graph $\GC$;   \newone
\end{enumerate}

\smallskip\noindent
{\bf variables: } 
\begin{enumerate}[leftmargin=*]
\item[-] $\eC(i)\in[0,1]$,  $i\in [1, \mC]$: 
$\eC(i)$ represents edge $a_i\in \EC$, $i\in [1,\mC]$  
 ($\eC(i)=1$, $i\in \Iew$;  $\eC(i)=0$, $i\in \It$)     
  ($\eC(i)=1$ $\Leftrightarrow $   edge $a_i$ is  used in  $\anC$);    
\item[-]  $\vT(i)\in[0,1]$,   $i\in [1,\tT]$:  
  $\vT(i)=1$ $\Leftrightarrow $ vertex $\vT_{i}$ is used in  $\anC$;   
\item[-]  $\eT(i)\in[0,1]$, $i\in [1,\tT+1]$:  $\eT(i)$ represents edge 
$\eT_{i}=(\vT_{i-1}, \vT_{i})\in \ET$,  
where $\eT_{1}$ and $\eT_{\tT+1}$ are fictitious edges
  ($\eT(i)=1$ $\Leftrightarrow $   edge $\eT_{i}$ is  used in  $\anC$);    
\item[-]  $\chiT(i)\in [0,\kC]$, $i\in [1,\tT]$: $\chiT(i)$ represents
 the color assigned to vertex $\vT_{i}$ 
  ($\chiT(i)=k>0$
   $\Leftrightarrow $  vertex $\vT_{i}$ is  assigned color $k$;
   $\chiT(i)=0$ means that vertex $\vT_{i}$ is not used in $\anC$);    
   
\item[-]  $\clrT(k)\in [\ell_\LB(k)-1, \ell_\UB(k)-1]$, $k\in [1,\kC]$, 
$\clrT(0)\in [0, \tT]$: the number of vertices 
$\vT_{i}\in \VT$  with color $c$;
%
\item[-]  $\dclrT(k)\in [0,1]$,   $k\in [0,\kC]$:
      $\dclrT(k)=1$    $\Leftrightarrow $ $\chiT(i)=k$ 
      for some $i\in [1,\tT]$;
      
\item[-]    $\chiT(i,k)\in[0,1]$,  $i\in [1,\tT]$, $k\in [0,\kC]$  
  ($\chiT(i,k)=1$    $\Leftrightarrow $ $\chiT(i)=k$);  
\item[-]   $\tldgC^+(i)\in [0,4]$, $i\in [1,\tC]$: 
the out-degree of vertex $\vC_{i}$ with the used edges $\eC$ in $\EC$; 

\item[-]   $\tldgC^-(i)\in [0,4]$, $i\in [1,\tC]$: 
the in-degree of vertex $\vC_{i}$  with the used edges $\eC$ in $\EC$; 

\item[-] $\mathrm{rank}$:  the rank $\mathrm{r}(\C)$ of a target 
chemical graph $\C$;   \newone
\end{enumerate}
  
\smallskip\noindent
{\bf constraints: }   
\begin{align} 
 \mathrm{rank} =  r_{\GC} -\sum_{i\in \Iz}(1-\eC(i)), && \label{eq:co_rank} \\
  \eC(i)=1,  ~~~  i\in \Iew,       &&  \label{eq:co_first}  \\
  \eC(i)=0,  ~~ \clrT(i)\geq 1,   ~~~  i\in \It,     &&   \label{eq:co_first} \\
  \eC(i)+ \clrT(i)\geq 1,  ~~~~~  \clrT(i)\leq \tT\cdot (1-\eC(i) ), 
~~~  i\in \Iw,    &&    \label{eq:co1c} 
\end{align}   
  
\begin{align}  
\sum_{ c\in \Iw^-(i)\cup \Iz^-(i)\cup \Iew^-(i) }\!\!\!\!\!\! \eC(c) 
 = \tldgC^-(i),  ~~ 
\sum_{ c\in \Iw^+(i)\cup \Iz^+(i)\cup \Iew^+(i) }\!\!\!\!\!\! \eC(c) 
 = \tldgC^+(i),  &&   i\in [1,\tC],   \label{eq:co_5}
\end{align}   
\begin{align} 
\chiT(i,0)=1 -\vT(i), ~~~
\sum_{k\in [0,\kC]} \chiT(i,k)=1,  ~~~ 
\sum_{k\in [0,\kC]}k\cdot \chiT(i,k)=\chiT(i),  && i\in[1,\tT],  \label{eq:co2} 
\end{align}   

\begin{align}  
\sum_{i\in[1,\tT]} \chiT(i,k)=\clrT(k), ~~
\tT\cdot \dclrT(k)\geq  \sum_{i\in [1,\tT]} \chiT(i,k)
\geq \dclrT(k), &&  k\in [0,\kC],    \label{eq:co3}   
\end{align}     
 
\begin{align}  
\vT(i-1)\geq \vT(i), && \notag \\
 \kC\cdot (\vT(i-1)-\eT(i )) \geq \chiT(i-1)-\chiT(i )
  \geq \vT(i-1) - \eT(i ), && i\in[2,\tT]. \label{eq:co_last} 
 \end{align}

\subsection{Constraints for Including Leaf Paths} 
\label{sec:int}

Let
$\widetilde{\tC}$  denote the number of vertices $u\in \VC$ such that 
$\bl_\UB(u)=1$ and assume that 
$\VC=\{u_1,u_2,\ldots, u_p\}$ so that 
\[ \mbox{ 
$\bl_\UB(u_i)=1$, $i\in [1,\widetilde{\tC}]$ and 
$\bl_\UB(u_i)=0$, $i\in[\widetilde{\tC}+1, \tC]$. }\]
Define the set of colors for the vertex set 
$\{u_i\mid i\in [1,\widetilde{\tC}] \}\cup \VT$
 to be $[1,\cF]$ with 
\[ \cF \triangleq \widetilde{\tC} + \tT 
=|\{u_i\mid i\in[1,\widetilde{\tC}]\}\cup \VT|. \]
Let each  vertex   $\vC_{i}$, $i\in[1,\widetilde{\tC}]$ 
(resp., $\vT_{i}\in \VT$)
  correspond to 
a color $i\in [1,\cF]$ (resp., $i+\widetilde{\tC} \in [1,\cF]$). 
When a path $P=(u, \vF_{j}, \vF_{j+1},\ldots, \vF_{j+t})$ 
from a vertex $u\in \VC\cup \VT$ 
  is used in $\anC$, we assign the color $i\in [1,\cF]$ of the vertex $u$
to the vertices $\vF_{j}, \vF_{j+1},\ldots, \vF_{j+t}\in \VF$.

\smallskip\noindent
{\bf constants: } 
\begin{enumerate}[leftmargin=*]
\item[-] $\cF$: the maximum number of different colors 
assigned to the vertices in $\VF$;  

\item[-]   $n^*$: an upper bound  
on the number $n(\C)$ of non-hydrogen atoms in $\C$;  

\item[-] $\nint_\LB, \nint_\UB \in [2,n^* ]$:
 lower and upper bounds on
the number of interior-vertices in $\C$; 

\item[-] $\bl_\LB(i) \in [0,1]$,  $i\in [1, \widetilde{\tC}]$: 
a lower   bound  on the number of leaf ${\rho}$-branches  in
the leaf path rooted  at a vertex $\vC_{i}$; 

\item[-]  $\bl_\LB(k),\bl_\UB(k)\in [0,\ell_\UB(k)-1]$, 
 $k\in[1,\kC]=\It\cup\Iw$: 
lower and upper bounds on the number of 
leaf ${\rho}$-branches in the trees rooted at internal vertices 
of a pure path $P_k$  for an edge $a_k\in \Ew\cup \Et$; 
\end{enumerate}

\smallskip\noindent
{\bf variables: } 
\begin{enumerate}[leftmargin=*]
  
\item[-]   $\nint_G\in [\nint_\LB, \nint_\UB]$: 
the number of interior-vertices in $\C$; 

\item[-]  $\vF(i)\in[0,1]$,   $i\in [1,\tF]$:
 $\vF(i)=1$ $\Leftrightarrow $ vertex $\vF_{i}$ is used in  $\C$;   
 
\item[-]  $\eF(i)\in[0,1]$, $i\in [1,\tF+1]$:  $\eF(i)$ represents edge 
$\eF_{i}=\vF_{i-1} \vF_{i}$,  
where $\eF_{1}$ and $\eF_{\tF+1}$ are fictitious edges
  ($\eF(i)=1$ $\Leftrightarrow $   edge $\eF_{i}$ is  used in  $\C$);    
\item[-]  $\chiF(i)\in [0,\cF]$, $i\in [1,\tF]$: $\chiF(i)$ represents
 the color assigned to  vertex $\vF_{i}$  
  ($\chiF(i)=c$ $\Leftrightarrow $  vertex $\vF_{i}$ is  assigned color $c$);   
  
\item[-]  $\clrF(c)\in [0, \tF]$, $c\in [0,\cF]$: the number of vertices $\vF_{i}$
 with color $c$;  
\item[-]  $\dclrF(c)\in [\bl_\LB(c), 1]$,  $c\in [1, \widetilde{\tC}]$:
      $\dclrF(c)=1$    $\Leftrightarrow $ $\chiF(i)=c$ for some $i\in [1,\tF]$;  
\item[-]  $\dclrF(c)\in[0,1]$,  $c\in [\widetilde{\tC}+1,\cF]$:
      $\dclrF(c)=1$    $\Leftrightarrow $ $\chiF(i)=c$ for some $i\in [1,\tF]$;  
\item[-]    $\chiF(i,c)\in[0,1]$,
 $i\in [1,\tF]$, $c\in [0,\cF]$:  
   $\chiF(i,c)=1$    $\Leftrightarrow $ $\chiF(i)=c$;     
\item[-]  $\bl(k,i)\in [0,1]$, $k\in[1,\kC]= \It\cup\Iw$,  $i\in[1,\tT]$: 
    $\bl(k,i)=1$ $\Leftrightarrow$ path $P_k$ contains vertex $\vT_{i}$ 
    as an internal vertex
    and the ${\rho}$-fringe-tree rooted at $\vT_{i}$ contains a leaf ${\rho}$-branch;
\end{enumerate}
  
\smallskip\noindent
{\bf constraints: }   
\begin{align} 
\chiF(i,0)=1 -\vF(i), ~~~
\sum_{c\in [0,\cF]} \chiF(i,c)=1,  ~~~ 
\sum_{c\in [0,\cF]}c\cdot \chiF(i,c)=\chiF(i),  &&  i\in[1,\tF],  \label{eq:int_first} 
\end{align}   

\begin{align}  
\sum_{i\in[1,\tF]} \chiF(i,c)=\clrF(c), ~~~ \tF\cdot \dclrF(c)\geq
\sum_{i\in [1,\tF]} \chiF(i,c)\geq \dclrF(c), &&  c\in [0,\cF],    \label{eq:int3}   
\end{align}   
 
\begin{align}  
 \eF(1)=\eF(\tF+1)=0,  && \label{eq:int4} 
 \end{align}   
 
\begin{align}  
\vF(i-1)\geq \vF(i), && \notag \\
 \cF\cdot (\vF(i-1)-\eF(i)) \geq \chiF(i-1)-\chiF(i) 
 \geq \vF(i-1)- \eF(i), && i\in[2,\tF], \label{eq:int6} 
 \end{align}



\begin{align}  
 \bl(k,i)\geq  \dclrF(\widetilde{\tC} + i)+\chiT(i,k)-1 , ~~~
 ~~~~~   k \in[1,\kC],   i\in[1,\tT], &&    
 \end{align}   
 
\begin{align}  
 \sum_{k \in[1,\kC],  i\in[1,\tT]} \bl(k,i)
 \leq \sum_{i\in[1,\tT]}\dclrF( \widetilde{\tC} +i),   &&    
  \label{eq:int12} 
 \end{align}   
  
  \begin{align}  
 \bl_\LB(k)\leq  \sum_{ i\in[1,\tT]} \bl(k,i) \leq  \bl_\UB(k) , ~~~~~~
     k \in[1,\kC], &&      
       \label{eq:int_last} 
 \end{align}

\begin{align}  
 \tC +\sum_{i\in [1,\tT]} \vT(i) + \sum_{i\in [1,\tF]} \vF(i) =\nint_G.  &&  
  \label{eq:int_last} 
 \end{align}   
 

\subsection{Constraints for Including Fringe-trees} \label{sec:ex}
 
 Recall that   $\mathcal{F}(D_\pi)$ denotes the set of 
chemical rooted trees $\psi$  
r-isomorphic to a chemical rooted tree in $\mathcal{T}(\C)$
  over all chemical graphs $\C\in D_\pi$,
  where possibly a chemical rooted tree $\psi\in \mathcal{F}(D_\pi)$
  consists of a single chemical element $\ta\in \Lambda\setminus \{{\tt H}\}$.

To express the condition that
the ${\rho}$-fringe-tree is chosen from a rooted tree $C_i$, $T_i$  or  $F_i$, 
we introduce the following set of variables and constraints.  
  
\smallskip\noindent
{\bf constants: } 
\begin{enumerate}[leftmargin=*]
\item[-]   $n_\LB$: a lower bound  
on the number $n(\C)$ of non-hydrogen atoms in $\C$,
where $n_\LB, n^*\geq \nint_\LB$;  

\item[-]   $\ch_{\LB}(i),\ch_{\UB}(i)\in [0,n^* ]$, $i\in [1,\tT]$: 
lower and upper bounds on $\h(\langle T_i\rangle)$ of the tree $T_i$ rooted 
at a vertex $\vC_{i}$; 

\item[-]   $\ch_{\LB}(k),\ch_{\UB}(k)\in [0,n^* ]$, $k \in[1,\kC]= \It\cup\Iw$: 
lower and upper bounds on the maximum  
 height $\h(\langle T \rangle)$ of the tree $T\in \F(P_k)$ rooted at 
an internal vertex of a path $P_k$   for an edge $a_k\in \Ew\cup \Et$;  


\item[-]  Prepare a coding of the set  $\mathcal{F}(D_\pi)$ and let 
    $[\psi]$ denote  the coded integer of  
     an element $\psi$ in $\mathcal{F}(D_\pi)$;  

\item[-]   Sets  $\mathcal{F}(v) \subseteq \mathcal{F}(D_\pi), v\in \VC$
and $\mathcal{F}_E \subseteq \mathcal{F}(D_\pi)$ 
 of chemical rooted trees $T$ with $\h(T)\in [1,{\rho}]$;  
 
\item[-]  Define
$\mathcal{F}^*:=\bigcup_{v\in \VC}\mathcal{F}(v)\cup \mathcal{F}_E$, 
 $\FrC_i:= \mathcal{F}(\vC_i)$, $i\in[1,\tC]$,
$\FrT_i:= \mathcal{F}_E$, $i\in[1,\tT]$  and 
$\FrF_i:= \mathcal{F}_E$, $i\in[1,\tF]$;

\item[-]    
 $\fc_\LB(\psi),\fc_\UB(\psi)\in[0,n^*], \psi\in \mathcal{F}^*$:
lower and upper bound functions  
  on the number of   interior-vertices $v$ 
  such that $\C[v]$ is r-isomorphic to $\psi $  in $\C$;
  
\item[-]  
$\FrX_i[p], p\in [1,{\rho}], \mathrm{X}\in\{\mathrm{C,T,F}\}$:
the set of  chemical rooted trees  $T\in  \FrX_i$
with   $\h(\langle T\rangle)= p$;  

\item[-]  
$n_{\oH}([\psi])\in [0, 3^{\rho}], \psi\in \mathcal{F}^*$: 
the number $n(\langle \psi\rangle)$ 
of non-root hydrogen vertices in a chemical rooted tree  $\psi$; 
 
\item[-]  
$\h_{\oH}([\psi])\in [0,{\rho}], \psi\in \mathcal{F}^*$: 
 the height $\h(\langle \psi\rangle)$  of the
 hydrogen-suppressed chemical rooted tree  $\langle \psi\rangle$; 

\item[-]  
$\deg_\mathrm{r}^{\oH}([\psi])\in [0,3], \psi\in \mathcal{F}^*$: 
the number $\deg_\mathrm{r}(\anpsi)$ of non-hydrogen children of the root $r$
 of a chemical rooted tree  $\psi$; 
 
\item[-]  
$\deghyd_\mathrm{r}([\psi])\in [0,3], \psi\in \mathcal{F}^*$: 
the number $\deg_\mathrm{r}(\psi)-\deg_\mathrm{r}(\anpsi)$ 
of hydrogen children of the root $r$ of a chemical rooted tree  $\psi$; 
 
\item[-] 
$\vion(\psi)\in [-3,+3], \psi\in \mathcal{F}^*$: 
  the ion-valence of the root in  $\psi$; 
  
\item[-] 
  $\ac^\lf_\nu(\psi), \nu\in \Gac^\lf$:
the frequency of leaf-edges with adjacency-configuration $\nu$ in $\psi$;
  
 \item[-] 
$\ac^\lf_\LB,\ac^\lf_\UB: \Gac^\lf \to  [0,n^*]$:
lower and upper bound functions    on the number of  leaf-edges $uv$ in $\acC$
  with adjacency-configuration $\nu$; 
\end{enumerate}

\smallskip\noindent
{\bf variables: }   
\begin{enumerate}[leftmargin=*]
\item[-]
  $n_G\in [n_\LB, n^*]$: the number $n(\C)$ of non-hydrogen atoms in $\C$;  
\item[-] $\vX(i)\in[0,1], i\in [1,\tX]$,   $\mathrm{X}\in\{\mathrm{T,F}\}$: 
 $\vX(i)=1$ $\Leftrightarrow $ vertex $\vX_{i}$ is used in $\C$; 
       
\item[-] 
$\dlfrX(i,[\psi])\in [0,1], 
    i\in[1,\tX], \psi\in \FrX_i, \mathrm{X}\in \{\mathrm{C,T,F}\}$:  
$\dlfrX(i,[\psi])=1$  $\Leftrightarrow $
 $\psi$ is  the ${\rho}$-fringe-tree rooted at vertex $\vX_i$ in $\C$;  

\item[-]    
$\fc([\psi])\in [\fc_\LB(\psi),\fc_\UB(\psi)], \psi\in \mathcal{F}^*$:
  the number of   interior-vertices $v$ 
  such that $\C[v]$ is r-isomorphic to $\psi$  in $\C$;  
  
\item[-]    
$\ac^\lf([\nu])\in [\ac^\lf_\LB(\nu),\ac^\lf_\UB(\nu)], \nu\in \Gac^\lf$: 
  the number of leaf-edge with adjacency-configuration $\nu$  in $\C$;  
  
\item[-]
 $\degXex(i)\in [0,3],  i\in [1,\tX],     \mathrm{X}\in\{\mathrm{C,T,F}\}$:
the number of non-hydrogen children of the root
 of  the ${\rho}$-fringe-tree rooted at vertex $\vX_i$ in $\C$;  
  
\item[-]  $\hyddegX(i)\in [0,4]$,  $i\in [1,\tX]$, 
 $\mathrm{X}\in \{\mathrm{C,T,F}\}$: 
 the number of  hydrogen atoms adjacent to  vertex $\vX_{i}$
 (i.e.,  $\hyddeg(\vX_{i})$) in $\C=(H,\alpha,\beta)$; 
 
\item[-] 
 $\eledegX(i)\in [-3,+3]$,  $i\in [1,\tX]$, 
 $\mathrm{X}\in \{\mathrm{C,T,F}\}$: 
 the  ion-valence $\vion(\psi)$ of vertex $\vX_{i}$
 (i.e.,  $\eledegX(i)=\vion(\psi)$ 
 for the ${\rho}$-fringe-tree $\psi$ rooted at $\vX_{i}$) in $\C=(H,\alpha,\beta)$; 
 
\item[-] $\hX(i)\in [0,{\rho}]$, $i\in [1,\tX]$,
$\mathrm{X}\in \{\mathrm{C,T,F}\}$: the height $\h(\langle T\rangle)$ of
the hydrogen-suppressed chemical rooted tree $\langle T\rangle$ of  
 the ${\rho}$-fringe-tree $T$ rooted at vertex $\vX_i$ in $\C$;  
\item[-] $\sigma(k,i)\in[0,1]$, $k \in[1,\kC]=\It\cup\Iw,  i\in [1,\tT]$: 
    $\sigma(k,i)=1$ $\Leftrightarrow$ 
    the ${\rho}$-fringe-tree $T_v$ rooted at  vertex $v=\vT_{i}$ 
      with color $k$  has the largest height $\h(\langle \T_v \rangle)$ among such trees
      $T_v, v\in \VT$;
\end{enumerate}

\smallskip\noindent
{\bf constraints: }    
\begin{align}    
\sum_{\psi\in \FrC_i}\!\!\dlfrC(i,[\psi]) =1, &&  i\in [1,\tC], \notag \\
\sum_{\psi\in \FrX_i }\!\!\dlfrX(i,[\psi]) =\vX(i),  
&&   i\in [1,\tX],  \mathrm{X}\in\{\mathrm{T,F}\},  \label{eq:ex_first}   
\end{align}     
 
\begin{align}    
\sum_{\psi\in \FrX_i }\!\! \deg_\mathrm{r}^{\oH}([\psi]) \cdot \dlfrX(i,[\psi]) 
 = \degXex(i),  &&  \notag \\
\sum_{\psi\in \FrX_i }\!\!   \deghyd_\mathrm{r}([\psi]) \cdot \dlfrX(i,[\psi])
 = \hyddegX(i),   &&  \notag \\
\sum_{\psi\in \FrX_i }\!\! \vion([\psi]) \cdot \dlfrX(i,[\psi]) 
 = \eledegX(i),
&&   i\in [1,\tX],  \mathrm{X}\in\{\mathrm{C,T,F}\},  \label{eq:ex_1}  
\end{align}    


\begin{align}    
\sum_{\psi\in \FrF_i[{\rho}] }\dlfrF(i,[\psi]) 
\geq \vF(i) - \eF(i+1),
      && i\in [1,\tF]~(\eF(\tF+1)=0), \label{eq:ex3}  
\end{align}   
 
\begin{align}   
\sum_{\psi\in \FrX_i } \h_{\oH}([\psi]) \cdot \dlfrX(i,[\psi]) =  \hX(i), && 
   i\in[1,\tX],  \mathrm{X}\in \{\mathrm{C,T,F}\}, \label{eq:ex5}  
\end{align}   

\begin{align}     
\sum\limits_{\substack{  \psi\in \FrX_i \\
              i\in [1,\tX],    \mathrm{X}\in \{\mathrm{C,T,F}\}  }}\!\!
                 n_{\oH}([\psi])  \cdot \dlfrX(i,[\psi])    
 +    \sum_{  i\in [1,\tX], \mathrm{X}\in \{\mathrm{T,F}\} } \vX(i)
   +\tC
  = n_G, ~~
 && 
  \label{eq:ex2} 
\end{align}   

\begin{align} 
\sum_{ i\in [1,\tX], \mathrm{X}\in\{\mathrm{C,T,F}\}} 
\dlfrX(i,[\psi]) =  \fc([\psi]), &&  \psi\in \mathcal{F}^*, 
 \label{eq:ex3} 
\end{align}    

\begin{align}    
\sum_{\psi\in \FrX_i, i\in[1,\tX], \mathrm{X}\in \{\mathrm{C,T,F}\}}
\ac^\lf_\nu(\psi)\cdot \dlfrX(i,[\psi]) = \ac^\lf([\nu]), &&
\nu\in \Gac^\lf, \label{eq:ex4}  
\end{align}

\begin{align}  
\hC(i)    \geq \ch_\LB(i)- n^* \cdot \dclrF(i),  ~~
\clrF(i)+{\rho} \geq \ch_\LB(i) , ~~~~~~~~~~~~~~   &&\notag \\
\hC(i)          \leq \ch_\UB(i) ,  ~~
\clrF(i)+{\rho} \leq \ch_\UB(i)+ n^* \cdot (1-\dclrF(i)),  ~ 
        &&   i\in [1,\widetilde{\tC}],         
            \label{eq:int14} 
 \end{align}   
 
\begin{align}  
 \ch_\LB(i) \leq  \hC(i)   \leq  \ch_\UB(i) ,  ~~ 
        &&    i\in [\widetilde{\tC}+1,\tC],       
             \label{eq:int14} 
 \end{align}   
 
\begin{align}   
 \hT(i)    \leq \ch_\UB(k) 
  + n^*\cdot (\dclrF( \widetilde{\tC}+ i)+1-\chiT(i,k)),  &&\notag \\
\clrF(\widetilde{\tC}+i)+{\rho}
 \leq \ch_\UB(k)+ n^*\cdot (2-\dclrF( \widetilde{\tC}+ i)-\chiT(i,k)),    
&& k \in[1,\kC],  i\in [1,\tT],      
    \label{eq:int15} 
 \end{align}   
 
\begin{align}   
 \sum_{i\in[1,\tT]}\sigma(k,i) =\dclrT(k),   &&   k \in[1,\kC],      
 \label{eq:int16} 
 \end{align}   
 
\begin{align}  
 \chiT(i,k)\geq \sigma(k,i), && \notag\\
 \hT(i)    \geq \ch_\LB(k) - n^*\cdot (\dclrF( \widetilde{\tC}+ i)+1-\sigma(k,i) ),
  && \notag\\ 
\clrF(\widetilde{\tC}+i)+{\rho}
 \geq \ch_\LB(k) - n^* \cdot (2-\dclrF( \widetilde{\tC}+ i)-\sigma(k,i)),   
&&  k \in[1,\kC],  i\in [1,\tT]. 
    \label{eq:ex_last} 
 \end{align}   

\subsection{Descriptor for the  Number of Specified Degree} 
\label{sec:Deg}

We include constraints to compute descriptors for degrees in $\C$. \\

\smallskip\noindent
{\bf variables: } 
\begin{enumerate}[leftmargin=*]
\item[-]  $\degX(i)\in [0,4]$,  $i\in [1,\tX]$, 
 $\mathrm{X}\in \{\mathrm{C,T,F}\}$: 
 the number of non-hydrogen atoms adjacent to  vertex $v=\vX_{i}$
 (i.e.,  $\deg_{\anC}(v)=\deg_H(v)-\hyddeg_{\C}(v)$) in $\C=(H,\alpha,\beta)$; 

\item[-] $\degCT(i)\in [0,4]$,  $i\in [1, \tC]$: the number of edges
from vertex $\vC_{i}$ to vertices $\vT_{j}$, $j\in [1,\tT]$;  
\item[-]  $\degTC(i)\in [0,4]$,  $i\in [1, \tC]$: the number of edges
from  vertices $\vT_{j}$, $j\in [1,\tT]$ to vertex $\vC_{i}$;    
\item[-]    $\ddgC(i,d)\in[0,1]$,  $i\in [1,\tC]$, $d\in [1,4]$, 
  $\ddgX(i,d)\in[0,1]$,  $i\in [1,\tX]$,
 $d\in [0,4]$,  $\mathrm{X}\in \{\mathrm{T,F}\}$: 
        $\ddgX(i,d)=1$ $\Leftrightarrow$   $\degX(i)+\hyddegX(i)=d$;  
       
\item[-]   $\dg(d)\in[\dg_\LB(d),\dg_\UB(d)]$,  $d \in[1,4]$:
    the number  of interior-vertices $v$ with 
       $\mathrm{deg}_H(\vX_{i})=d$   in $\C=(H,\alpha,\beta)$;

\item[-] $\degCint(i)\in [1,4]$,  $i\in [1, \tC]$, 
 $\degXint(i)\in [0,4]$,  $i\in [1, \tX], \mathrm{X}\in \{\mathrm{T,F}\}$: 
the interior-degree $\deg_{H^\inte}(\vX_i)$ 
  in the interior $H^\inte=(V^\inte(\C),E^\inte(\C))$ of  $\C$; i.e., 
the number of interior-edges incident to vertex $\vX_{i}$;

\item[-]    $\ddgCint(i,d)\in[0,1]$,  $i\in [1,\tC]$,  $d\in [1,4]$,  
  $\ddgXint(i,d)\in[0,1]$,  $i\in [1,\tX]$,
 $d\in [0,4]$,  $\mathrm{X}\in \{\mathrm{T,F}\}$: 
       $\ddgXint(i,d)=1$ $\Leftrightarrow$   $\degXint(i)=d$;  
       
\item[-]   $\dg^\inte(d)\in[\dg_\LB(d),\dg_\UB(d)]$,  $d \in[1,4]$:
    the number  of interior-vertices $v$ with
    the interior-degree  $\deg_{H^\inte}(v)=d$
  in the interior $H^\inte=(V^\inte(\C),E^\inte(\C))$ of  $\C=(H,\alpha,\beta)$.
  
\end{enumerate}
   
\smallskip\noindent
{\bf constraints: }   
\begin{align}   
\sum_{   k\in \It^+(i)\cup \Iw^+(i)} \dclrT(k) = \degCT(i), ~~
 \sum_{   k\in \It^-(i)\cup \Iw^-(i)} \dclrT(k) = \degTC(i), 
    &&    i\in [1, \tC],     \label{eq:Deg_first}  
\end{align}

\begin{align}   
\tldgC^-(i)+\tldgC^+(i)   + \degCT(i)  + \degTC(i) + \dclrF(i) = \degCint(i),  
    &&    i\in [1, \widetilde{\tC}],     \label{eq:Deg2}  
\end{align}   

\begin{align}      
\tldgC^-(i)+\tldgC^+(i)  + \degCT(i)  + \degTC(i)   = \degCint(i),  
     &&   i\in [\widetilde{\tC}+1,\tC],     \label{eq:Deg2b}  
\end{align}   

\begin{align}      
  \degCint(i)+ \degCex(i) = \degC(i),  
    &&    i\in [1, \tC],     \label{eq:Deg2c}  
\end{align}   

\begin{align}    
\sum_{\psi\in \FrC_i[{\rho}] }\dlfrC(i,[\psi]) \geq 2-\degCint(i)
      &&  i\in [1, \tC],    \label{eq:Deg2d} 
\end{align}   

\begin{align}   
  2\vT(i)   + \dclrF(\widetilde{\tC}+i)    =\degTint(i),   && \notag \\
 \degTint(i)+ \degTex(i)  =\degT(i),   && 
  i\in [1,\tT]~(\eT(1)=\eT(\tT+1)=0), \label{eq:Deg3}  
\end{align}   

\begin{align}
   \vF(i) +\eF(i+1)  =\degFint(i),  && \notag \\
   \degFint(i)  +\degFex(i)   = \degF(i),   && 
  i\in [1,\tF] ~(\eF(1)=\eF(\tF+1)=0),   \label{eq:Deg4}  
\end{align} 

\begin{align}   
\sum_{d\in [0,4]}\ddgX(i,d)=1, ~
\sum_{d\in [1,4]}d\cdot\ddgX(i,d)=\degX(i)+\hyddegX(i), && \notag \\
\sum_{d\in [0,4]}\ddgXint(i,d)=1, ~
\sum_{d\in [1,4]}d\cdot\ddgXint(i,d)=\degXint(i), && 
 i\in [1,\tX],  \mathrm{X}\in \{\mathrm{T, C, F}\}, \label{eq:Deg5}  
\end{align}   
 
\begin{align}   
\sum_{ i\in [1,\tC]} \ddgC(i,d) + \sum_{ i\in [1,\tT]} \ddgT(i,d) 
 + \sum_{ i\in [1,\tF] }  \ddgF(i,d) = \dg(d),   &&   \notag \\
\sum_{ i\in [1,\tC]} \ddgCint(i,d) + \sum_{ i\in [1,\tT]} \ddgTint(i,d) 
 + \sum_{ i\in [1,\tF] }  \ddgFint(i,d) = \dg^\inte(d),     
    && d\in [1,4].  
  \label{eq:Deg_last}  
\end{align}   

\subsection{Assigning Multiplicity} 
\label{sec:beta}

 We prepare an integer variable $\beta(e)$  
 for each edge $e$ in the scheme graph $\mathrm{SG}$ 
 to denote the bond-multiplicity of $e$ in a selected graph $H$ and
 include necessary constraints for the variables to satisfy in $H$. 
 
\smallskip\noindent
{\bf constants: }
\begin{enumerate}[leftmargin=*]
\item[-]
$\betar([\psi])$: the sum $\beta_\psi(r)$ of bond-multiplicities of edges
incident to  the root $r$ of a chemical rooted tree $\psi\in \mathcal{F}^*$; 
\end{enumerate}

\smallskip\noindent
{\bf variables: } 
\begin{enumerate}[leftmargin=*]
\item[-] $\bX(i)\in [0,3]$,   $i\in [2,\tX]$, $\mathrm{X}\in \{\mathrm{T,F}\}$:   
 the bond-multiplicity of edge  $\eX_{i}$ in $\C$;  
 
\item[-] $\bC(i)\in [0,3]$,     $i\in [\widetilde{\kC}+1,\mC]= \Iw\cup \Iz\cup\Iew$:    
     the bond-multiplicity of 
     edge  $a_{i}\in \Ew\cup \Ez\cup\Eew$ in $\C$;        
\item[-]
   $\bCT(k), \bTC(k)\in [0,3]$, $k\in [1, \kC]=\It\cup \Iw$: 
   the bond-multiplicity of the first (resp., last) edge of the pure path $P_k$ in $\C$;    
   
\item[-]
   $\bsF(c)\in [0,3], c\in [1,\cF=\widetilde{\tC} + \tT ]$:  
   the bond-multiplicity of the first edge of the leaf path $Q_c$
   rooted at vertex $\vC_{c}, c\leq\widetilde{\tC} $
    or $\vT_{c-\widetilde{\tC}}, c>\widetilde{\tC} $  in $\C$;   
    
\item[-] $\bXex(i)\in [0,4],  i\in [1,\tX],   \mathrm{X}\in\{\mathrm{C,T,F}\}$:
the sum $\beta_{\C[v]}(v)$ of bond-multiplicities of edges in the ${\rho}$-fringe-tree
$\C[v]$ rooted at  interior-vertex $v=\vX_{i}$;  

\item[-] $\delbX(i,m)\in [0,1]$, $i\in [2,\tX]$,   $m\in[0,3]$, 
       $\mathrm{X}\in \{\mathrm{T,F}\}$:  
  $\delbX(i,m)=1$  $\Leftrightarrow$  $\bX(i)=m$; 
\item[-] $\delbC(i,m)\in [0,1]$,  
   $i\in [\widetilde{\kC},\mC]=\Iw\cup \Iz\cup\Iew$,  $m\in[0,3]$:  
   $\delbC(i,m)=1$  $\Leftrightarrow$  $\bC(i)=m$; 
\item[-]
   $\delbCT(k,m), \delbTC(k,m)\in [0,1]$, $k\in [1, \kC]=\It\cup \Iw$,  $m\in[0,3]$:
     $\delbCT(k,m)=1$   (resp., $\delbTC(k,m)=1$)    $\Leftrightarrow$  
           $\bCT(k)=m$ (resp., $\bTC(k)=m$); 
           
\item[-]
   $\delbsF(c,m)\in [0,1]$, $c\in [1,\cF]$,  
    $m\in[0,3], \mathrm{X}\in \{\mathrm{C,T}\}$: 
     $\delbsF(c,m)=1$ $\Leftrightarrow$  $\bsF(c)=m$;  
\item[-] $\bd^\inte(m)\in[0, 2\nint_\UB]$, $m\in[1,3]$:
      the number of interior-edges with bond-multiplicity  $m$ in $\C$;  
      
\item[-] $\bdX(m)\in [0,2\nint_\UB],  \mathrm{X}\in \{\mathrm{C,T,CT,TC}\}$,
      $\bdX(m)\in [0,2\nint_\UB], \mathrm{X}\in \{\mathrm{F,CF,TF}\}$, $m\in[1,3]$:  
 the number of interior-edges $e\in \EX$ with bond-multiplicity  $m$ in  $\C$; 
\end{enumerate}
 
\smallskip\noindent
{\bf constraints: } 
\begin{align}    
\eC(i)\leq \bC(i)\leq 3\eC(i), 
  i\in [\widetilde{\kC}+1,\mC]=\Iw\cup \Iz\cup\Iew, \label{eq:beta_first} 
\end{align}   

\begin{align}   
  \eX(i)\leq \bX(i)\leq 3 \eX(i), 
  &&    i\in [2,\tX],   \mathrm{X}\in \{\mathrm{T, F}\},    \label{eq:beta1}  
\end{align}   

\begin{align}   
\dclrT(k)\leq \bCT(k)\leq 3 \dclrT(k), ~~~ 
\dclrT(k)\leq \bTC(k)\leq 3 \dclrT(k), &&  k\in [1, \kC], \label{eq:beta8} \\ 
\dclrF(c)\leq \bXF(c)\leq 3 \dclrF(c), &&   c\in [1,\cF], \label{eq:beta8}  
\end{align}

\begin{align} 
\sum_{m\in[0,3]} \delbX(i,m)=1,  ~~
\sum_{m\in[0,3]}m\cdot \delbX(i,m)=\bX(i), &&   i\in [2,\tX],  
  \mathrm{X}\in \{\mathrm{T,F}\},  \label{eq:beta10}    
\end{align}   

\begin{align} 
\sum_{m\in[0,3]} \delbC(i,m)=1,  ~~
\sum_{m\in[0,3]}m\cdot \delbC(i,m)=\bC(i), &&   i\in [\widetilde{\kC}+1,\mC],   \label{eq:beta11}    
\end{align}   
 
\begin{align}   
\sum_{m\in[0,3]} \delbCT(k,m)=1,    ~~ 
\sum_{m\in[0,3]}m\cdot\delbCT(k,m)=\bCT(k),
&&     k\in [1, \kC],  \notag \\      
\sum_{m\in[0,3]} \delbTC(k,m)=1,  ~~ 
\sum_{m\in[0,3]} m\cdot\delbTC(k,m)=\bTC(k),  &&
      k\in [1, \kC], \notag \\
\sum_{m\in[0,3]} \delbsF(c,m)=1,  ~~ 
\sum_{m\in[0,3]} m\cdot\delbsF(c,m)=\bsF(c),  &&    c\in [1,\cF],
   \label{eq:beta15}    
\end{align}    

 \begin{align}           
\sum_{\psi\in \FrX_i } \betar([\psi]) \cdot  \dlfrX(i,[\psi]) = \bXex(i),     
 &&  i\in [1,\tX],     \mathrm{X}\in\{\mathrm{C,T,F}\}, 
 \label{eq:beta16a}   
\end{align}

\begin{align} 
 \sum_{i\in [\widetilde{\kC}+1,\mC]} \delbC(i,m) =\bdC(m), ~~
  \sum_{i\in [2,\tT]} \delbT(i,m)    =\bdT(m),   \notag \\ 
   \sum_{k\in [1, \kC]}\delbCT(k,m)=\bdCT(m), ~~
   \sum_{k\in [1, \kC]}\delbTC(k,m)=\bdTC(m),    \notag \\ 
\sum_{i\in [2,\tF]}\!\!\! \delbF(i,m) =\bdF(m), ~~
 \sum_{c\in [1,\widetilde{\tC}]} \delbsF(c,m)  =\bdCF(m),   \notag \\ 
  \sum_{c\in [\widetilde{\tC}+1,\cF]}  \delbsF(c,m) =\bdTF(m),  
   \notag \\ 
 \bdC(m)+\bdT(m) + \bdF(m)
 +\bdCT(m)+\bdTC(m) +\bdTF(m)+\bdCF(m) = \bd^\inte(m),   \notag \\ 
  m\in [1,3].       \label{eq:beta_last} 
\end{align}

\subsection{Assigning Chemical Elements and  Valence Condition}
\label{sec:alpha} 

We include constraints so that each vertex $v$ in a selected graph $H$
satisfies the valence condition; i.e., 
$\beta_\C(v)=  \val(\alpha(v)) +\eledeg_\C(v)$, 
where $\eledeg_\C(v)=\vion(\psi)$ for the ${\rho}$-fringe-tree $\C[v]$
r-isomorphic to $\psi$. 
With these constraints, a chemical graph
   $\C=(H,\alpha,\beta)$ on a selected subgraph $H$
will be constructed. 
 
\smallskip\noindent
{\bf constants: }
 \begin{enumerate}[leftmargin=*]
\item[-] Subsets
 $\Lambda^\inte \subseteq \Lambda\setminus\{{\tt H}\}, 
 \Lambda^\ex \subseteq \Lambda$ of chemical elements,
 where we denote by $[{\tt e}]$ (resp., $[{\tt e}]^\inte$ and $[{\tt e}]^\ex$)  
 of a standard encoding of an element ${\tt e}$ in the set $\Lambda$ 
 (resp.,    $\Lambda^\inte_\epsilon$ and  $\Lambda^\ex_\epsilon$);  
\item[-]  A valence function: $\val: \Lambda \to [1,6]$;  

\item[-]  A function $\mathrm{mass}^*:\Lambda\to \mathbb{Z}$ 
(we let $\mathrm{mass}(\ta)$ denote  the observed mass of a chemical element  
$\ta\in \Lambda$, and define 
   $\mathrm{mass}^*(\ta)\triangleq
    \lfloor 10\cdot \mathrm{mass}(\ta)\rfloor$);  
        
\item[-]  
 Subsets $\Lambda^*(i)\subseteq \Lambda^\inte$, $i\in[1,\tC]$; 
 
\item[-] 
 $\na_\LB(\ta),\na_\UB(\ta)\in [0,n^* ]$,  $\ta\in  \Lambda$:
lower and upper bounds on the number of vertices  $v$    with $\alpha(v)=\ta$;  
\item[-] 
  $\na_\LB^\inte(\ta),\na_\UB^\inte(\ta)\in [0,n^* ]$,
 $\ta\in  \Lambda^\inte$:
lower and upper bounds on the number  of interior-vertices  
 $v$ with $\alpha(v)=\ta$; 

\item[-] 
$\alpha_\mathrm{r}([\psi])\in [\Lambda^\ex], \in \mathcal{F}^*$:
 the chemical element $\alpha(r)$  of the root $r$ of  $\psi$;

\item[-]   $\na_\ta^\ex([\psi])\in [0,n^*]$,  
$\ta\in \Lambda^\ex, \psi\in \mathcal{F}^*$: 
the frequency of chemical element $\ta$ in the set of  
non-rooted vertices   in   $\psi$, where possibly $\ta={\tt H}$;


\item[-]   $\mathrm{M}$: 
an upper bound for the average $\overline{\mathrm{ms}}(\C)$ of mass$^*$ 
over all atoms in $\C$;

\end{enumerate}
      
\smallskip\noindent
{\bf variables: } 
\begin{enumerate}[leftmargin=*]
\item[-]
   $\bCT(i),\bTC(i)\in [0,3], i\in [1,\tT]$:
the bond-multiplicity of edge $\eCT_{j,i}$ (resp., $\eTC_{j,i}$)
if one exists;  

\item[-] 
 $\bCF(i), \bTF(i)\in [0,3], i\in [1,\tF]$:
the bond-multiplicity of $\eCF_{j,i}$ (resp., $\eTF_{j,i}$)
if one exists;  

\item[-]  $\aX(i)\in [\Lambda^\inte_\epsilon ],
       \delaX(i,[\ta]^\inte)\in [0,1],  \ta\in \Lambda^\inte_\epsilon, i\in [1,\tX],
         \mathrm{X}\in \{\mathrm{C,T,F}\}$:  
$\aX(i)= [\ta]^\inte\geq 1$  (resp., $\aX(i)=0$)
  $\Leftrightarrow$ $\delaX(i,[\ta]^\inte)=1$ (resp., $\delaX(i,0)=0$)  
  $\Leftrightarrow$ $\alpha(\vX_{i})= \ta\in \Lambda$ 
(resp., vertex $\vX_{i}$ is not used in $\C$); 

\item[-] 
$\delaX(i,[\ta]^\inte)\in [0,1], i\in [1,\tX],  
\ta  \in \Lambda^\inte,   \mathrm{X}\in\{\mathrm{C,T,F}\}$:   
   $\delaX(i,[\ta]^\typ)=1$   $\Leftrightarrow$   $\alpha(\vX_{i})=\ta$;  
\item[-]  $\mathrm{Mass}\in \mathbb{Z}_+$: 
 $\sum_{v\in V(H)} \mathrm{mass}^*(\alpha(v))$;  
 
\item[-]  $\overline{\mathrm{ms}}\in \mathbb{R}_+$: 
 $\sum_{v\in V(H)} \mathrm{mass}^*(\alpha(v)) / |V(H)|$;  
 \item[-] 
$\delta_{\mathrm{atm}}(i)\in [0,1], i\in [n_\LB + \na_\LB({\tt H}), 
n^* + \na_\UB({\tt H})]$:   
   $\delta_{\mathrm{atm}}(i)=1$   $\Leftrightarrow$   $|V(H)| = i$;  

\item[-]   $\na([\ta])\in[\na_\LB(\ta),\na_\UB(\ta)]$,
 $\ta \in \Lambda$:
    the number  of vertices $v\in V(H)$
     with $\alpha(v)=\ta$, where possibly $\ta={\tt H}$; 
    
\item[-]   $\na^{\inte}([\ta]^\inte) \in[\na_\LB^\inte(\ta),\na_\UB^\inte(\ta)]$,
 $\ta \in \Lambda, \mathrm{X}\in \{\mathrm{C,T,F}\}$:
    the number  of interior-vertices  $v\in V(\C)$ 
    with $\alpha(v)=\ta$; 
    
\item[-]   $\naX^\ex([\ta]^\ex) , \na ^\ex([\ta]^\ex) \in [0,\na_\UB(\ta)]$,
 $\ta \in \Lambda$,   $\mathrm{X}\in \{\mathrm{C,T,F}\}$: 
    the number    of   exterior-vertices rooted at vertices $v\in\VX$ 
  and  the number    of   exterior-vertices $v$
     such that  $\alpha(v)=\ta$;
     

\end{enumerate} 
    
\smallskip\noindent
{\bf constraints: } 
\begin{align}    
 \bCT(k)-3(\eT(i)-\chiT(i,k)+1) \leq 
\bCT(i)\leq \bCT(k)+3(\eT(i)-\chiT(i,k)+1),  i\in [1,\tT], && \notag\\   
  \bTC(k)-3(\eT(i+1)-\chiT(i,k)+1) \leq 
\bTC(i)\leq \bTC(k)+3(\eT(i+1)-\chiT(i,k)+1),  i\in [1,\tT], && \notag\\ 
   k\in [1, \kC],   &&  \label{eq:alpha_first}  
\end{align}

\begin{align}   
 \bsF(c)-3(\eF(i)-\chiF(i,c)+1) \leq 
\bCF(i)\leq \bsF(c)+3(\eF(i)-\chiF(i,c)+1),   i\in [1,\tF], 
&&  c\in[1,\widetilde{\tC}] ,    \notag\\  
  \bsF(c)-3(\eF(i)-\chiF(i,c)+1) \leq 
\bTF(i)\leq \bsF(c)+3(\eF(i)-\chiF(i,c)+1),    i\in [1,\tF], 
&& c\in[\widetilde{\tC}+1,\cF] ,    \notag\\  
  \label{eq:alpha2}  
\end{align}

\begin{align}  
   \sum_{\ta\in \Lambda^\inte} \delaC(i,[\ta]^\inte)=1, ~~ 
   \sum_{\ta\in \Lambda^\inte} [\ta]^\inte\cdot\delaX(i,[\ta]^\inte)=\aC(i),   
     &&  i\in [1,\tC], \notag  \\
  \sum_{\ta\in \Lambda^\inte } \delaX(i,[\ta]^\inte)=\vX(i), ~~ 
   \sum_{\ta\in \Lambda^\inte} [\ta]^\inte\cdot\delaX(i,[\ta]^\inte)=\aX(i), 
    &&  i\in [1,\tX],  \mathrm{X}\in \{\mathrm{T,F}\},  
  \label{eq:alpha_first} 
\end{align}

\begin{align}  
\sum_{\psi\in \FrX_i } 
 \alpha_\mathrm{r}([\psi])\cdot \dlfrX(i,[\psi]) = \aX(i),   
 &&   i\in [1,\tX],   \mathrm{X}\in \{\mathrm{C,T,F}\},  
  \label{eq:alpha_1} 
\end{align}    
  

\begin{align}  
\sum_{j\in \IC(i)}\bC(j)  
+ \sum_{  k\in \It^+(i)\cup \Iw^+(i)} \bCT(k)
+ \sum_{  k\in \It^-(i)\cup \Iw^-(i)} \bTC(k)   &&  \notag\\
     + \bsF(i)  +\bCex(i) -\eledegC(i) 
     =
     \sum_{\ta\in \Lambda^\inte}\val(\ta)\delaC(i,[\ta]^\inte),  
 &&  i\in [1,\widetilde{\tC}],  \label{eq:alpha3} 
\end{align}

\begin{align}   
\sum_{j\in \IC(i)}\bC(j)  
+ \sum_{  k\in \It^+(i)\cup \Iw^+(i)} \bCT(k)
+ \sum_{  k\in \It^-(i)\cup \Iw^-(i)} \bTC(k)   &&  \notag\\
+\bCex(i) -\eledegC(i) 
   =
      \sum_{\ta\in \Lambda^\inte}\val(\ta)\delaC(i,[\ta]^\inte),  
  &&  i\in [\widetilde{\tC}+1,\tC],   \label{eq:alpha3b} 
\end{align} 

\begin{align}  
 \bT(i)+\bT(i\!+\!1)   +   \bTex(i)  
  + \bCT(i) + \bTC(i) \hspace{1cm} \notag\\
  + \bsF(\widetilde{\tC}+i) -\eledegT(i) 
   =
      \sum_{\ta\in \Lambda^\inte}\val(\ta)\delaT(i,[\ta]^\inte), 
   \notag \\
  i\in [1,\tT]~  (\bT(1)=\bT(\tT+1)=0),   \label{eq:alpha4} 
\end{align}
 
\begin{align} 
 \bF(i)+\bF(i\!+\!1) +\bCF(i) +\bTF(i)  \hspace{1cm}    \notag\\
  +\bFex(i)  -\eledegF(i) 
   =
      \sum_{\ta\in \Lambda^\inte}\val(\ta)\delaF(i,[\ta]^\inte),  
   \notag \\
  i\in [1,\tF] ~  (\bF(1)=\bF(\tF+1)=0),   \label{eq:alpha5} 
\end{align}

\begin{align}  
 \sum_{i\in [1,\tX] } \delaX(i,[\ta]^\inte) = \naX([\ta]^\inte) ,  
 &&  \ta\in \Lambda^\inte, \mathrm{X}\in \{\mathrm{C,T,F}\},  
     \label{eq:alpha6} 
\end{align}    

\begin{align}   
\sum_{\psi\in \FrX_i , i\in [1,\tX] } \na_\ta^\ex([\psi])\cdot  \dlfrX(i,[\psi])  
   = \naX^\ex([\ta]^\ex),     && 
 \ta\in \Lambda^\ex,  \mathrm{X}\in \{\mathrm{C,T,F}\},  \label{eq:alpha6} 
\end{align}    
       
\begin{align}  
 \naC([\ta]^\inte)+ \naT([\ta]^\inte)+ \naF([\ta]^\inte) =  \na^\inte([\ta]^\inte),  
    &&   \ta\in \Lambda^\inte,       \notag \\
  \sum_{  \mathrm{X}\in \{\mathrm{C,T,F}\} }
   \naX^\ex([\ta]^\ex)       =\na^\ex([\ta]^\ex),  
  &&  \ta\in \Lambda^\ex,    \notag \\
  \na^\inte([\ta]^\inte) + \na^\ex([\ta]^\ex)=\na([\ta]),  
 &&   \ta\in \Lambda^\inte\cap \Lambda^\ex,     \notag \\
  \na^\inte([\ta]^\inte)  =\na([\ta]),  
 &&      \ta\in \Lambda^\inte \setminus \Lambda^\ex,     \notag \\ 
   \na^\ex([\ta]^\ex)  =\na([\ta]),  
 &&      \ta\in \Lambda^\ex \setminus \Lambda^\inte,        
     \label{eq:alpha6} 
\end{align}

 \begin{align}    
 \sum_{\ta\in \Lambda^*(i)} \delaC(i,[\ta]^\inte) = 1,  
  &&  i\in [1,\tC],   \label{eq:alpha8} 
\end{align}

\begin{align}   
\sum_{ \ta\in\Lambda }\mathrm{mass}^*(\ta )\cdot \na([\ta])
 =\mathrm{Mass}, &&    \label{eq:alpha7} 
\end{align}  

 \begin{align}
 \sum_{i \in [n_\LB + \na_\LB({\tt H}), n^* + \na_\UB({\tt H})]} \delta_{\mathrm{atm}}(i) = 1, && \\
 \sum_{i \in [n_\LB + \na_\LB({\tt H}), n^* + \na_\UB({\tt H})]} i \cdot \delta_{\mathrm{atm}}(i) 
 = n_G + \na^\ex([{\tt H}]^\ex), &&  \\
 \mathrm{Mass} / i - \mathrm{M}\cdot (1 - \delta_{\mathrm{atm}}(i)) 
 \le
 \overline{\mathrm{ms}} 
 \le \mathrm{Mass} / i+ \mathrm{M}\cdot (1 - \delta_{\mathrm{atm}}(i)), 
   && i \in [n_\LB + \na_\LB({\tt H}), n^* + \na_\UB({\tt H})]. 
  \label{eq:alpha_last} 
 \end{align}

\subsection{Constraints for Bounds on the Number of Bonds}  
\label{sec:BDbond}

We include constraints for specification of lower and upper bounds
$\bd_\LB$ and $\bd_\UB$. 

\smallskip\noindent
{\bf constants: } 
\begin{enumerate}[leftmargin=*]
\item[-]
$\bd_{m, \LB}(i), \bd_{m, \UB}(i)\in [0,\nint_\UB]$, 
$i\in [1,\mC]$,  $m\in [2,3]$:  lower and upper bounds 
 on the number  of edges $e\in E(P_i)$ with bond-multiplicity $\beta(e)=m$
 in the pure path $P_i$ for edge $e_i\in \EC$; 
\end{enumerate}

\smallskip\noindent
{\bf variables : } 
\begin{enumerate}[leftmargin=*]
\item[-]
  $\bdT(k,i,m)\in [0,1]$, $k\in [1, \kC]$, $i\in [2,\tT]$, $m\in [2,3]$:  
  $\bdT(k,i,m)=1$  $\Leftrightarrow$ the pure path $P_k$ for edge $e_k\in \EC$ 
  contains edge $\eT_i$ with $\beta(\eT_i)=m$; 
\end{enumerate}
  
\smallskip\noindent
{\bf constraints: } 
\begin{align}    
\bd_{m,\LB}(i)\leq \delbC(i,m)\leq \bd_{m,\UB}(i), 
  i\in \Iew\cup \Iz, m\in [2,3], && 
  \label{eq:BDbond_first}  
\end{align}

\begin{align}   
\bdT(k,i,m)\geq \delbT(i,m)+\chiT(i,k)-1, 
~~~ k \in [1, \kC], i\in [2,\tT],  m\in [2,3], && 
 \label{eq:BDbond2}  
\end{align}   
 
\begin{align}    
\sum_{j\in[2,\tT]}\delbT(j,m) \geq 
\sum_{k\in[1, \kC], i\in [2,\tT]}\!\!\!\! \bdT(k,i,m) , 
~~ m\in [2,3],  \label{eq:BDbond3}  
\end{align}    

\begin{align}    
 \bd_{m, \LB}(k) \leq 
   \sum_{i\in [2,\tT]}\bdT(k,i,m) +\delbCT(k,m)+\delbTC(k,m)    
   \leq   \bd_{m, \UB}(k), ~~~~  \notag \\
    k\in [1, \kC],   m\in [2,3]. ~~ 
     \label{eq:BDbond_last}  
\end{align}

\subsection{Descriptor for the Number of  Adjacency-configurations}  
\label{sec:AC}

We call a tuple $(\ta,\tb,m)
\in (\Lambda\setminus\{{\tt H}\})\times (\Lambda\setminus\{{\tt H}\}) \times[1,3]$
an {\em adjacency-configuration}.
The adjacency-configuration of an edge-configuration
$(\mu=\ta d, \mu'=\tb d', m)$ is defined to be
 $(\ta,\tb,m)$.
We include constraints to compute the frequency of each adjacency-configuration
in an inferred chemical graph $\C$. 
 
\smallskip\noindent
{\bf constants: } 
\begin{enumerate}[leftmargin=*]
\item[-] A set  $\Gamma^\inte$ of edge-configurations $\gamma=(\mu,\mu',m)$ 
with  $\mu\leq \mu'$;

\item[-] 
Let  $\overline{\gamma}$ of an edge-configuration $\gamma=(\mu,\mu',m)$
denote the  edge-configuration $(\mu',\mu,m)$; 

\item[-] Let $\Gamma_{<}^\inte=\{(\mu,\mu',m)\in  \Gamma^\inte\mid \mu < \mu' \}$, 
$\Gamma_{=}^\inte=\{(\mu,\mu',m)\in  \Gamma^\inte\mid \mu= \mu' \}$
and    $\Gamma_{>}^\inte=\{\overline{\gamma}\mid 
    \gamma\in  \Gamma_{<}^\inte  \}$;
    
\item[-] 
Let  $\Gacs^\inte$, $\Gace^\inte$ and $\Gacl^\inte$ 
 denote  the sets of the adjacency-configurations of
edge-configurations in the sets 
$\Gamma_{<}^\inte$, $\Gamma_{=}^\inte$ and $\Gamma_{>}^\inte$, 
 respectively;
 
\item[-] 
Let  $\overline{\nu}$ of an adjacency-configuration $\nu=(\ta, \tb,m)$
denote the  adjacency-configuration $(\tb,\ta,m)$; 

\item[-] 
 Prepare a coding of   the  set 
$\Gac^\inte \cup \Gacl^\inte$  and let 
$[\nu]^\inte$   denote  
the coded integer of  an element $\nu$ in $\Gac^\inte \cup \Gacl^\inte$; 

\item[-] 
Choose subsets   $\tGacC,\tGacT,\tGacCT,\tGacTC, \tGacF, \tGacCF , \tGacTF
   \subseteq \Gac^\inte\cup\Gacl^\inte$;  
 To compute the frequency     of adjacency-configurations exactly,  set 
  $\tGacC:= \tGacT :=\tGacCT:=  \tGacTC :=\tGacF:=  \tGacCF := \tGacTF :=
  \Gac^\inte\cup\Gacl^\inte$; 

\item[-]  $\ac_\LB^\inte(\nu),  \ac_\UB^\inte(\nu) \in [0,2\nint_\UB ], 
\nu=(\ta,\tb,m)\in \Gac^\inte$: 
lower and upper bounds on the number 
  of interior-edges  $e=uv$  with $\alpha(u)=\ta$, 
 $\alpha(v)=\tb$ and $\beta(e)=m$; 
\end{enumerate}

\smallskip\noindent
{\bf variables: } 
\begin{enumerate}[leftmargin=*]
\item[-]
$\ac^\inte([\nu]^\inte) \in [\ac_\LB^\inte(\nu), \ac_\UB^\inte(\nu)], 
\nu\in \Gac^\inte$: 
the number of interior-edges  with  adjacency-configuration $\nu$; 
\item[-]
$\acC([\nu]^\inte)\in [0,\mC],  \nu\in \tGacC$, 
$\acT([\nu]^\inte)\in [0,\tT],   \nu\in \tGacT$, 
$\acF([\nu]^\inte)\in [0,\tF], \nu\in \tGacF$:  
the number of  edges $\eC\in \EC$ (resp.,  edges $\eT\in \ET$
and   edges $\eF\in \EF$)  with  adjacency-configuration $\nu$; 

\item[-]
$\acCT([\nu]^\inte)\in [0, \min\{\kC,\tT\} ], \nu\in \tGacCT$,
$\acTC([\nu]^\inte)\in [0,\min\{\kC,\tT\} ], \nu\in \tGacCT$, 
$\acCF([\nu]^\inte)\in [0,\widetilde{\tC}],  \nu\in \tGacCF$,
$\acTF([\nu]^\inte)\in [0,\tT], \nu\in \tGacTF$: 
the number of  edges   $\eCT\in \ECT$  
(resp.,   edges $\eTC\in \ETC$
and  edges $\eCF\in \ECF$ and $\eTF\in \ETF$)  with  adjacency-configuration $\nu$; 
%
\item[-]
$\dlacC(i,[\nu]^\inte)\in [0,1], 
  i\in [\widetilde{\kC}+1,\mC]=\Iw\cup \Iz\cup\Iew, \nu\in \tGacC$, 
$\dlacT(i,[\nu]^\inte)\in [0,1],  i\in [2,\tT],   \nu\in \tGacT$, 
$\dlacF(i,[\nu]^\inte)\in [0,1] , i\in [2,\tF],\nu\in \tGacF$:
$\dlacX(i,[\nu]^\inte)=1$  $\Leftrightarrow$
edge  $\eX_i$ has  adjacency-configuration $\nu$; 
\item[-]
$\dlacCT(k,[\nu]^\inte),\dlacTC(k,[\nu]^\inte)\in [0,1],
k\in [1, \kC]=\It\cup \Iw,  \nu\in \tGacCT$: 
$\dlacCT(k,[\nu]^\inte)=1$   (resp., $\dlacTC(k,[\nu]^\inte)=1$)  $\Leftrightarrow$
edge  $\eCT_{\tail(k),j}$ (resp.,  $\eTC_{\hd(k),j}$) 
for some $j\in [1,\tT]$ has  adjacency-configuration $\nu$;

\item[-]
$\dlacCF(c,[\nu]^\inte)\in [0,1],  c\in [1,\widetilde{\tC}],\nu\in \tGacCF$:
$\dlacCF(c,[\nu]^\inte)=1$    $\Leftrightarrow$
edge   $\eCF_{c,i}$  for some $i\in [1,\tF]$ has  adjacency-configuration $\nu$;

\item[-]
  $\dlacTF(i,[\nu]^\inte)\in [0,1],  i\in [1,\tT],  \nu\in \tGacTF$: 
   $\dlacTF(i,[\nu]^\inte)=1$  $\Leftrightarrow$
edge   $\eTF_{i,j}$
 for some $j\in [1,\tF]$ has  adjacency-configuration $\nu$;   
\item[-]
$\aCT(k),\aTC(k)\in [0, |\Lambda^\inte|],   k\in [1, \kC]$: 
$\alpha(v)$  of the edge $(\vC_{\mathrm{tail}(k)},v)\in \ECT$   
 (resp., $(v,\vC_{\mathrm{head}(k)})\in \ETC$) if any; 
 
\item[-]
$\aCF(c)\in [0, |\Lambda^\inte|], c\in [1,\widetilde{\tC}]$: 
 $\alpha(v)$  of the edge $(\vC_{c},v)\in \ECF$   if any; 
 
\item[-]
$\aTF(i)\in [0, |\Lambda^\inte|], i\in [1,\tT]$: 
 $\alpha(v)$  of the edge $(\vT_{i},v)\in \ETF$   if any; 
\item[-]
$\DlacCp(i),  \DlacCm(i), \in [0,|\Lambda^\inte|], 
  i\in [\widetilde{\kC}+1,\mC]$, 
$\DlacTp(i),\DlacTm(i)\in [0,|\Lambda^\inte|],  i\in [2,\tT]$,
$\DlacFp(i),\DlacFm(i)\in [0,|\Lambda^\inte|] , i\in [2,\tF]$: 
$\DlacXp(i)=\DlacXm(i)=0$ (resp., 
 $\DlacXp(i)=\alpha(u)$ and $\DlacXm(i)=\alpha(v)$) $\Leftrightarrow$
edge  $\eX_i=(u,v)\in \EX$   is used in $\C$ (resp., $\eX_i\not\in E(G)$); 

\item[-]
$\DlacCTp(k),\DlacCTm(k)\in [0,|\Lambda^\inte|],
k\in [1, \kC]=\It\cup \Iw$: 
$\DlacCTp(k)=\DlacCTm(k) =0$ 
(resp.,  $\DlacCTp(k)=\alpha(u)$ and $\DlacCTm(k)=\alpha(v)$) 
 $\Leftrightarrow$  
edge  $\eCT_{\tail(k),j}=(u,v)\in \ECT$   
  for some $j\in [1,\tT]$ is used in $\C$ (resp., otherwise);  
  
\item[-]
$\DlacTCp(k),\DlacTCm(k)\in [0,|\Lambda^\inte|],
k\in [1, \kC]=\It\cup \Iw$: 
Analogous with $\DlacCTp(k)$ and $\DlacCTm(k)$;

\item[-]
$\DlacCFp(c)\in [0,|\Lambda^\inte|],
\DlacCFm(c) \in [0,|\Lambda^\inte|],  c\in [1,\widetilde{\tC}]$: 
$\DlacCFp(c)=\DlacCFm(c) =0$ (resp., 
 $\DlacCFp(c)=\alpha(u)$ and $\DlacCFm(c)=\alpha(v)$) 
 $\Leftrightarrow$ 
edge  $\eCF_{c,i}=(u,v)\in \ECF$  
   for some $i\in [1,\tF]$ is used in $\C$ (resp., otherwise);  
\item[-]
  $\DlacTFp(i)\in [0,|\Lambda^\inte|],
  \DlacTFm(i)\in [0,|\Lambda^\inte|],  i\in [1,\tT]$:  
  Analogous with $\DlacCFp(c)$ and $\DlacCFm(c)$;
\end{enumerate}

\smallskip\noindent
{\bf constraints: } 

\begin{align} 
 \acC([\nu]^\inte) =0,  &&  \nu \in \Gac^\inte\setminus \tGacC , \notag \\
 \acT([\nu]^\inte) =0,  &&  \nu \in \Gac^\inte\setminus \tGacT , \notag \\
 \acF([\nu]^\inte) =0,  &&  \nu \in \Gac^\inte\setminus \tGacF , \notag \\
 \acCT([\nu]^\inte) =0,  &&  \nu \in \Gac^\inte\setminus \tGacCT , \notag \\
 \acTC([\nu]^\inte) =0,  &&  \nu \in \Gac^\inte\setminus \tGacTC , \notag \\
 \acCF([\nu]^\inte) =0,  &&  \nu \in \Gac^\inte\setminus \tGacCF , \notag \\
 \acTF([\nu]^\inte) =0,  &&  \nu \in \Gac^\inte\setminus \tGacTF , \notag \\
 \label{eq:AC_first} 
\end{align}    

\begin{align} 
 \sum_{(\ta, \tb,m)=\nu\in  \Gac^\inte}\acC([\nu]^\inte) 
      =\sum_{i\in [\widetilde{\kC}+1,\mC]}\delbC(i,m),  &&   m\in [1,3]   , \notag \\
 \sum_{(\ta, \tb,m)=\nu\in  \Gac^\inte}\acT([\nu]^\inte) 
      =\sum_{i\in [2,\tT]}\delbT(i,m) ,  &&   m\in [1,3] , \notag \\
 \sum_{(\ta, \tb,m)=\nu\in \Gac^\inte}\acF([\nu]^\inte)
      =\sum_{i\in [2,\tF]}\delbF(i,m) ,  &&   m\in [1,3]  , \notag \\
 \sum_{(\ta, \tb,m)=\nu\in \Gac^\inte}\acCT([\nu]^\inte)
     =\sum_{k\in [1, \kC]} \delbCT(k,m),  &&   m\in [1,3]  , \notag \\
 \sum_{(\ta, \tb,m)=\nu\in \Gac^\inte}\acTC([\nu]^\inte)
    =\sum_{k\in [1, \kC]} \delbTC(k,m),  &&   m\in [1,3]  , \notag \\
 \sum_{(\ta, \tb,m)=\nu\in \Gac^\inte}\acCF([\nu]^\inte)
    =\sum_{c\in [1,\widetilde{\tC}]} \delbsF(c,m),  &&   m\in [1,3]  , \notag \\
 \sum_{(\ta, \tb,m)=\nu\in \Gac^\inte}\acTF([\nu]^\inte) 
    =\sum_{c\in [\widetilde{\tC}+1, \cF]} \delbsF(c,m),  &&   m\in [1,3]  , \notag \\
 \label{eq:AC_first2} 
\end{align}    

\begin{align} 
\sum_{\nu=(\ta,\tb,m) \in \tGacC }\!\!\!\! m\cdot \dlacC(i, [\nu]^\inte) 
=\bC(i), && \notag  \\  
\DlacCp(i) +\sum_{\nu=(\ta,\tb,m) \in \tGacC }\!\!\!\! [\ta]^\inte\dlacC(i, [\nu]^\inte) 
=\aC(\tail(i)),  && \notag  \\
\DlacCm(i) +\sum_{\nu=(\ta,\tb,m) \in \tGacC }\!\!\!\! [\tb]^\inte\dlacC(i, [\nu]^\inte) 
=\aC(\hd(i)),  &&\notag  \\
\DlacCp(i)+\DlacCm(i) \leq 2|\Lambda^\inte|(1 - \eC(i)),
&& i\in [\widetilde{\kC}+1,\mC],  \notag  \\
\sum_{i\in [\widetilde{\kC}+1,\mC]}\!\!\!\! \dlacC(i, [\nu]^\inte) =\acC([\nu]^\inte),  
&&  \nu \in \tGacC ,   \label{eq:AC1}   
\end{align}

\begin{align} 
\sum_{\nu=(\ta,\tb,m) \in \tGacT }\!\!\!\! m\cdot \dlacT(i, [\nu]^\inte) 
=\bT(i), && \notag  \\  
\DlacTp(i) +\sum_{\nu=(\ta,\tb,m) \in \tGacT }\!\!\!\! [\ta]^\inte\dlacT(i, [\nu]^\inte) 
=\aT(i-1),  && \notag  \\
\DlacTm(i) +\sum_{\nu=(\ta,\tb,m) \in \tGacT }\!\!\!\! [\tb]^\inte\dlacT(i, [\nu]^\inte) 
=\aT(i),  &&\notag  \\
\DlacTp(i)+\DlacTm(i) \leq 2|\Lambda^\inte|(1 - \eT(i)),
&& i\in [2,\tT],   \notag  \\
 \sum_{ i\in [2,\tT]} \!\! \dlacT(i, [\nu]^\inte) =\acT([\nu]^\inte),    
&& \nu \in \tGacT ,    \label{eq:AC2} 
\end{align}    

\begin{align} 
\sum_{\nu=(\ta,\tb,m) \in \tGacF }\!\!\!\! m\cdot \dlacF(i, [\nu]^\inte) 
=\bF(i), && \notag  \\  
\DlacFp(i) +\sum_{\nu=(\ta,\tb,m) \in \tGacF }\!\!\!\! [\ta]^\inte\dlacF(i, [\nu]^\inte) 
=\aF(i-1),  && \notag  \\
\DlacFm(i) +\sum_{\nu=(\ta,\tb,m) \in \tGacF }\!\!\!\! [\tb]^\inte\dlacF(i, [\nu]^\inte) 
=\aF(i),  &&\notag  \\
\DlacFp(i)+\DlacFm(i) \leq 2|\Lambda^\ex|(1 - \eF(i)),
&& i\in [2,\tF],   \notag  \\
  \sum_{ i\in [2,\tF]} \!\! \dlacF(i, [\nu]^\inte) =\acF([\nu]^\inte),  
 &&  \nu \in \tGacF ,    \label{eq:AC3} 
\end{align}

\begin{align} 
 \aT(i)+|\Lambda^\inte|(1-\chiT(i,k)+\eT(i))\geq \aCT(k),  && \notag  \\  
\aCT(k)\geq \aT(i)- |\Lambda^\inte|(1-\chiT(i,k)+\eT(i)), 
&&  i\in [1,\tT],     \notag  \\  
\sum_{\nu=(\ta,\tb,m) \in \tGacCT }\!\!\!\! m\cdot \dlacCT(k, [\nu]^\inte) 
=\bCT(k), && \notag  \\  
\DlacCTp(k) +\sum_{\nu=(\ta,\tb,m) \in \tGacCT }\!\!\!\! [\ta]^\inte\dlacCT(k, [\nu]^\inte) 
=\aC(\tail(k)),  && \notag  \\
\DlacCTm(k) +\sum_{\nu=(\ta,\tb,m) \in \tGacCT }\!\!\!\! [\tb]^\inte\dlacCT(k, [\nu]^\inte) 
=  \aCT(k),  &&\notag  \\
\DlacCTp(k)+\DlacCTm(k) \leq 2|\Lambda^\inte|(1 - \dclrT(k)),
&& k\in [1, \kC],  \notag  \\
\sum_{k\in [1, \kC]}\!\! \dlacCT(k, [\nu]^\inte) =\acCT([\nu]^\inte),  
 && \nu \in  \tGacCT ,  \label{eq:AC5} 
\end{align}

\begin{align} 
 \aT(i)+|\Lambda^\inte|(1-\chiT(i,k)+\eT(i+1))\geq \aTC(k),   && \notag  \\  
\aTC(k)\geq \aT(i)- |\Lambda^\inte|(1-\chiT(i,k)+\eT(i+1)), 
&&  i\in [1,\tT],    \notag  \\  
\sum_{\nu=(\ta,\tb,m) \in \tGacTC }\!\!\!\! m\cdot \dlacTC(k, [\nu]^\inte) 
=\bTC(k), && \notag  \\  
\DlacTCp(k) +\sum_{\nu=(\ta,\tb,m) \in \tGacTC }\!\!\!\! [\ta]^\inte\dlacTC(k, [\nu]^\inte) 
=  \aTC(k),  &&\notag  \\
\DlacTCm(k) +\sum_{\nu=(\ta,\tb,m) \in \tGacTC }\!\!\!\! [\tb]^\inte\dlacTC(k, [\nu]^\inte) 
=\aC(\hd(k)),  &&   \notag  \\
\DlacTCp(k)+\DlacTCm(k) \leq 2|\Lambda^\inte|(1 - \dclrT(k)), && k\in [1, \kC],  \notag  \\  
\sum_{k\in [1, \kC]}\!\! \dlacTC(k, [\nu]^\inte) =\acTC([\nu]^\inte),  
 && \nu \in  \tGacTC ,  \label{eq:AC5} 
\end{align}

\begin{align} 
\aF(i)+|\Lambda^\inte|(1-\chiF(i,c)+\eF(i))\geq \aCF(c),  && \notag  \\  
\aCF(c)\geq \aF(i)- |\Lambda^\inte|(1-\chiF(i,c)+\eF(i)), 
   &&   i\in [1,\tF], \notag  \\   
\sum_{\nu=(\ta,\tb,m) \in \tGacCF }\!\!\!\! m\cdot \dlacCF(c, [\nu]^\inte) 
=\bsF(c), && \notag  \\  
\DlacCFp(c) +\sum_{\nu=(\ta,\tb,m) \in \tGacCF }\!\!\!\! [\ta]^\inte\dlacCF(c, [\nu]^\inte) 
= \aC(\hd(c)),  &&\notag  \\
\DlacCFm(c) +\sum_{\nu=(\ta,\tb,m) \in \tGacCF }\!\!\!\! [\tb]^\inte\dlacCF(c, [\nu]^\inte) 
=\aCF(c) ,   &&\notag  \\
\DlacCFp(c)+\DlacCFm(c) \leq 2\max\{|\Lambda^\inte|,|\Lambda^\inte|\}(1 - \dclrF(c)), 
&& c\in [1,\widetilde{\tC}],  \notag  \\
\sum_{c\in [1,\widetilde{\tC}]}\!\! \dlacCF(c, [\nu]^\inte) =\acCF([\nu]^\inte),  
 && \nu \in  \tGacCF ,  \label{eq:AC6} 
\end{align}    

\begin{align} 
\aF(j)+|\Lambda^\inte|(1-\chiF(j,i+\widetilde{\tC})+\eF(j))\geq \aTF(i), 
   && \notag  \\  
\aTF(i)\geq \aF(j)- |\Lambda^\inte|(1-\chiF(j,i+\widetilde{\tC})+\eF(j)), 
  &&  j\in [1,\tF],  \notag  \\   
\sum_{\nu=(\ta,\tb,m) \in \tGacTF }\!\!\!\! m\cdot \dlacTF(i, [\nu]^\inte) 
=\bsF(i+\widetilde{\tC}), && \notag  \\  
\DlacTFp(i) +\sum_{\nu=(\ta,\tb,m) \in \tGacTF }\!\!\!\! [\ta]^\inte\dlacTF(i, [\nu]^\inte) 
= \aT(i),  &&\notag  \\
\DlacTFm(i) +\sum_{\nu=(\ta,\tb,m) \in \tGacTF }\!\!\!\! [\tb]^\inte\dlacTF(i, [\nu]^\inte) 
=\aTF(i) ,   
&& \notag  \\
\DlacTFp(i)+\DlacTFm(i) \leq 2\max\{|\Lambda^\inte|,|\Lambda^\inte|\}
(1 - \dclrF(i+\widetilde{\tC})), 
&& i\in [1,\tT],  \notag  \\
\sum_{i\in [1,\tT]}\!\! \dlacTF(i, [\nu]^\inte) =\acTF([\nu]^\inte),  
 && \nu \in  \tGacTF ,  \label{eq:AC5} 
\end{align}

\begin{align} 
\sum_{\mathrm{X}\in\{\mathrm{C,T,F,CT,TC,CF,TF}\}}\!\!\!\!\!\!
   (\acX([\nu]^\inte)+\acX([\overline{\nu}]^\inte)) 
 =\ac^\inte([\nu]^\inte) , && \nu\in \Gacs^\inte,  \notag \\  
\sum_{\mathrm{X}\in\{\mathrm{C,T,F,CT,TC,CF,TF}\}} \!\!\!\!\!\!
    \acX([\nu]^\inte)
 =\ac^\inte([\nu]^\inte) , && \nu\in \Gace^\inte.    
   \label{eq:AC_last} 
\end{align}

\subsection{Descriptor for the Number of Chemical Symbols}  
\label{sec:CS}

We include constraints for computing
 the frequency of each chemical symbol in $\Ldg$.
 Let $\cs(v)$ denote the chemical symbol of an interior-vertex $v$ in 
 a chemical graph $\C$ to be inferred; i.e.,
 $\cs(v)=\mu=\ta d\in \Ldg$ such that $\alpha(v)=\ta$ and
  $\deg_{\anC}(v)=\deg_H(v)-\deghyd_\C(v)=d$ in $\C=(H,\alpha,\beta)$. 

\smallskip\noindent
{\bf constants: } 
\begin{enumerate}[leftmargin=*]
\item[-] A set  $\Ldg^\inte$  of chemical symbols;
 
\item[-]
 Prepare a coding of each of the two sets 
$\Ldg^\inte$    and let $[\mu]^\inte$  denote  
the coded integer of  an element $\mu \in \Ldg^\inte$; 

\item[-]
Choose subsets  $\tLdgC, \tLdgT, \tLdgF  \subseteq \Ldg^\inte$:   
 To compute the frequency of chemical symbols exactly,  set
  $\tLdgC:= \tLdgT :=  \tLdgF :=\Ldg^\inte$;  
  
\end{enumerate}

\smallskip\noindent
{\bf variables: }
\begin{enumerate}[leftmargin=*]
\item[-] $\ns^\inte([\mu]^\inte )\in[0,\nint_\UB]$,  $\mu\in \Ldg^\inte$: 
      the number of interior-vertices $v$  with $\cs(v)=\mu$;   
\item[-] 
   $\dlnsX(i,[\mu]^\inte)\in [0,1]$, $ i\in [1,\tX],\mu\in \Ldg^\inte$, 
   $\mathrm{X}\in \{\mathrm{C,T,F}\}$;
\end{enumerate}

\smallskip\noindent
{\bf constraints: } 
\begin{align}  
   \sum_{\mu\in \tLdgX\cup\{\epsilon\} } \dlnsX(i,[\mu]^\inte)=1, ~~ 
   \sum_{\mu=\ta d\in \tLdgX }[\ta]^\inte\cdot\dlnsX(i,[\mu]^\inte)=\aX(i), 
   \notag \\
   \sum_{\mu=\ta d\in \tLdgX }d\cdot\dlnsX(i,[\mu]^\inte)
   =\degX(i),
    \hspace{2cm}\notag \\ 
  ~~   i\in [1,\tX],   
  \mathrm{X}\in \{\mathrm{C,T,F}\},  
  \label{eq:CS_first} 
\end{align}

\begin{align}  
   \sum_{i\in [1,\tC]} \dlnsC(i,[\mu]^\inte)
   +    \sum_{i\in [1,\tT]} \dlnsT(i,[\mu]^\inte) 
 + \sum_{i\in [1,\tF]} \dlnsF(i,[\mu]^\inte)=\ns^\inte([\mu]^\inte),
    && \mu\in \Ldg^\inte.
  \label{eq:CS_last} 
\end{align}

\subsection{Descriptor for the Number of Edge-configurations}  
\label{sec:EC}

We include constraints to compute the frequency of each edge-configuration
in an inferred chemical graph $\C$. 
 
\smallskip\noindent
{\bf constants: } 
\begin{enumerate}[leftmargin=*]
\item[-] A set  $\Gamma^\inte$ of edge-configurations $\gamma=(\mu,\mu',m)$
with $\mu\leq \mu'$;

\item[-]  Let $\Gamma_{<}^\inte=\{(\mu,\mu',m)\in  \Gamma^\inte\mid \mu < \mu' \}$, 
$\Gamma_{=}^\inte=\{(\mu,\mu',m)\in  \Gamma^\inte\mid \mu= \mu' \}$
and   $\Gamma_{>}^\inte=\{(\mu',\mu,m)\mid 
    (\mu,\mu',m)\in  \Gamma_{<}^\inte  \}$;
    
\item[-] 
 Prepare a coding of  the set 
$\Gamma^\inte \cup \Gamma_{>}^\inte$  and let 
$[\gamma]^\inte$   denote  
the coded integer of  an element $\gamma$ in $\Gamma^\inte \cup \Gamma_{>}^\inte$; 
  
\item[-] 
Choose subsets   $\tGecC,\tGecT,\tGecCT,\tGecTC,\tGecF, \tGecCF , \tGecTF
 \subseteq \Gamma^\inte\cup\Gamma_{>}^\inte$;  
 To compute the frequency  of edge-configurations exactly,  set 
  $\tGecC:= \tGecT :=\tGecCT:=  \tGecTC :=\tGecF:=  \tGecCF := \tGecTF := \Gamma^\inte\cup\Gamma_{>}^\inte$;  
  
\item[-] $\ec_\LB^\inte(\gamma),  \ec_\UB^\inte(\gamma) \in [0,2\nint_\UB ], 
\gamma=(\mu,\mu',m)\in \Gamma^\inte$: 
lower and upper bounds on the number  of interior-edges  $e=uv$ 
with $\cs(u)=\mu$,  $\cs(v)=\mu'$ and $\beta(e)=m$; 
\end{enumerate}

\smallskip\noindent
{\bf variables: } 
\begin{enumerate}[leftmargin=*]
\item[-]
$\ec^\inte([\gamma]^\inte) \in [\ec_\LB^\inte(\gamma),\ec_\UB^\inte(\gamma)], \gamma\in \Gamma^\inte$: 
the number of interior-edges     with  edge-configuration $\gamma$;   
\item[-]
$\ecC([\gamma]^\inte)\in [0,\mC],  \gamma\in \tGecC$, 
$\ecT([\gamma]^\inte)\in [0,\tT],   \gamma\in \tGecT$, 
$\ecF([\gamma]^\inte)\in [0,\tF], \gamma\in \tGecF$:  
the number of  edges $\eC\in \EC$ (resp.,   edges $\eT\in \ET$
and  edges $\eF\in \EF$)   with  edge-configuration $\gamma$;  
\item[-]
$\ecCT([\gamma]^\inte)\in [0, \min\{\kC,\tT\} ], \gamma\in \tGecCT$,
$\ecTC([\gamma]^\inte)\in [0,\min\{\kC,\tT\} ], \gamma\in \tGecCT$, 
$\ecCF([\gamma]^\inte)\in [0,\widetilde{\tC}],  \gamma\in \tGecCF$,
$\ecTF([\gamma]^\inte)\in [0,\tT], \gamma\in \tGecTF$:
the number of  edges $\eCT\in \ECT$  
(resp.,   edges $\eTC\in \ETC$
and   edges $\eCF\in \ECF$ and $\eTF\in \ETF$)  with  edge-configuration $\gamma$;  
\item[-]
$\dlecC(i,[\gamma]^\inte)\in [0,1], 
  i\in [\widetilde{\kC}+1,\mC]=\Iw\cup \Iz\cup\Iew, \gamma\in \tGecC$,  
$\dlecT(i,[\gamma]^\inte)\in [0,1],  i\in [2,\tT],   \gamma\in \tGecT$,  
$\dlecF(i,[\gamma]^\inte)\in [0,1] , i\in [2,\tF],\gamma\in \tGecF$: 
$\dlecX(i,[\gamma]^\typ)=1$  $\Leftrightarrow$
edge  $\eX_i$ has  edge-configuration $\gamma$;  
\item[-]
$\dlecCTC(k,[\gamma]^\inte),\dlecTCC(k,[\gamma]^\inte)\in [0,1],
k\in [1, \kC]=\It\cup \Iw,  \gamma\in \tGecCT$: 
$\dlecCTC(k,[\gamma]^\inte)=1$   (resp., $\dlecTCC(k,[\gamma]^\inte)=1$)
  $\Leftrightarrow$
edge  $\eCT_{\tail(k),j}$ (resp.,  $\eTC_{\hd(k),j}$) 
   for some $j\in [1,\tT]$ has  edge-configuration $\gamma$; 
   
\item[-]
$\dlecCFC(c,[\gamma]^\inte)\in [0,1],  c\in [1,\widetilde{\tC}],\gamma\in \tGecCF$:
$\dlecCFC(c,[\gamma]^\inte)=1$    $\Leftrightarrow$
edge   $\eCF_{c,i}$  for some $i\in [1,\tF]$ has  edge-configuration $\gamma$; 

\item[-]
  $\dlecTFT(i,[\gamma]^\inte)\in [0,1],  i\in [1,\tT],  \gamma\in \tGecTF$:
    $\dlecTFT(i,[\gamma]^\inte)=1$  $\Leftrightarrow$
edge     $\eTF_{i,j}$ for some $j\in [1,\tF]$ has  edge-configuration $\gamma$; 

\item[-]
$\degCTT(k),\degTCT(k)\in [0, 4],   k\in [1, \kC]$: 
$\deg_{\anC}(v)$  of an end-vertex $v\in \VT$ of  
the edge $(\vC_{\mathrm{tail}(k)},v)\in \ECT$  
 (resp., $(v,\vC_{\mathrm{head}(k)})\in \ETC$)  if any; 
 
\item[-]
$\degCFF(c)\in [0, 4], c\in [1,\widetilde{\tC}]$: 
 $\deg_{\anC}(v)$  of an end-vertex $v\in \VF$ of  
 the edge $(\vC_{c},v)\in \ECF$   if any; 
 
\item[-]
$\degTFF(i)\in [0, 4], i\in [1,\tT]$: 
 $\deg_{\anC}(v)$   of an end-vertex $v\in \VF$ of  
  the edge $(\vT_{i},v)\in \ETF$   if any;  
\item[-]
$\DlecCp(i),  \DlecCm(i), \in [0,4], 
  i\in [\widetilde{\kC}+1,\mC]$,  
$\DlecTp(i),\DlecTm(i)\in [0,4],  i\in [2,\tT]$,  
$\DlecFp(i),\DlecFm(i)\in [0,4] , i\in [2,\tF]$: 
$\DlecXp(i)=\DlecXm(i)=0$ (resp., 
 $\DlecXp(i)=\deg_{\anC}(u)$
  and $\DlecXm(i)=\deg_{\anC}(v)$) $\Leftrightarrow$  
edge  $\eX_i=(u,v)\in \EX$  is used in ${\anC}$ (resp., $\eX_i\not\in E({\anC})$);  
\item[-]
$\DlecCTp(k),\DlecCTm(k)\in [0,4],
k\in [1, \kC]=\It\cup \Iw$: 
$\DlecCTp(k)=\DlecCTm(k) =0$ 
(resp.,  $\DlecCTp(k)=\deg_{\anC}(u)$
 and $\DlecCTm(k)=\deg_{\anC}(v)$) 
 $\Leftrightarrow$  
edge  $\eCT_{\tail(k),j}=(u,v)\in \ECT$   
  for some $j\in [1,\tT]$ is used in ${\anC}$ (resp., otherwise); 
  
\item[-]
$\DlecTCp(k),\DlecTCm(k)\in [0,4],
k\in [1, \kC]=\It\cup \Iw$: 
Analogous with $\DlecCTp(k)$ and $\DlecCTm(k)$;

\item[-]
$\DlacCFp(c), \DlecCFm(c) \in [0,4],  c\in [1,\widetilde{\tC}]$: 
$\DlecCFp(c)=\DlecCFm(c) =0$ (resp., 
 $\DlecCFp(c)=\deg_{\anC}(u)$ 
 and $\DlecCFm(c)=\deg_{\anC}(v)$) 
 $\Leftrightarrow$  
edge  $\eCF_{c,j}=(u,v)\in \ECF$  
   for some $j\in [1,\tF]$ is used in ${\anC}$ (resp., otherwise);  
\item[-]
  $\DlecTFp(i),  \DlecTFm(i)\in [0,4],  i\in [1,\tT]$:
  Analogous with $\DlecCFp(c)$ and $\DlecCFm(c)$;
\end{enumerate}
  
\smallskip\noindent
{\bf constraints: } 

\begin{align} 
 \ecC([\gamma]^\inte) =0,  &&  \gamma \in \Gamma^\inte\setminus \tGecC , \notag \\
 \ecT([\gamma]^\inte) =0,  &&  \gamma \in \Gamma^\inte\setminus \tGecT , \notag \\
 \ecF([\gamma]^\inte) =0,  &&  \gamma \in \Gamma^\inte\setminus \tGecF , \notag \\
 %
 %
 \ecCT([\gamma]^\inte) =0,  &&  \gamma \in \Gamma^\inte\setminus \tGecCT , \notag \\
 \ecTC([\gamma]^\inte) =0,  &&  \gamma \in \Gamma^\inte\setminus \tGecTC , \notag \\
 \ecCF([\gamma]^\inte) =0,  &&  \gamma \in \Gamma^\inte\setminus \tGecCF , \notag \\
 \ecTF([\gamma]^\inte) =0,  &&  \gamma \in \Gamma^\inte\setminus \tGecTF , \notag \\
 \label{eq:EC_first} 
\end{align}

\begin{align} 
 \sum_{(\mu, \mu',m)=\gamma\in  \Gamma^\inte}\ecC([\gamma]^\inte) 
      =\sum_{i\in [\widetilde{\kC}+1,\mC]}\delbC(i,m),  &&   m\in [1,3]   , \notag \\
 \sum_{(\mu, \mu',m)=\gamma\in  \Gamma^\inte}\ecT([\gamma]^\inte) 
      =\sum_{i\in [2,\tT]}\delbT(i,m) ,  &&   m\in [1,3] , \notag \\
 \sum_{(\mu, \mu',m)=\gamma\in \Gamma^\inte}\ecF([\gamma]^\inte)
      =\sum_{i\in [2,\tF]}\delbF(i,m) ,  &&   m\in [1,3]  , \notag \\
 \sum_{(\mu, \mu',m)=\gamma\in \Gamma^\inte}\ecCT([\gamma]^\inte)
     =\sum_{k\in [1, \kC]} \delbCT(k,m),  &&   m\in [1,3]  , \notag \\
 \sum_{(\mu, \mu',m)=\gamma\in \Gamma^\inte}\ecTC([\gamma]^\inte)
    =\sum_{k\in [1, \kC]} \delbTC(k,m),  &&   m\in [1,3]  , \notag \\
 \sum_{(\mu, \mu',m)=\gamma\in \Gamma^\inte}\ecCF([\gamma]^\inte)
    =\sum_{c\in [1,\widetilde{\tC}]} \delbsF(c,m),  &&   m\in [1,3]  , \notag \\
 \sum_{(\mu, \mu',m)=\gamma\in \Gamma^\inte}\ecTF([\gamma]^\inte) 
    =\sum_{c\in [\widetilde{\tC}+1, \cF]} \delbsF(c,m),  &&   m\in [1,3]  , \notag \\
 \label{eq:EC_first2} 
\end{align}

\begin{align}  
\sum_{\gamma=(\ta d,\tb d',m) \in \tGecC }\!\!\!\! [(\ta,\tb,m)]^\inte\cdot \dlecC(i, [\gamma]^\inte) 
= \sum_{\nu \in \tGacC } [\nu]^\inte\cdot \dlacC(i, [\nu]^\inte) , && \notag  \\  
\DlecCp(i) +\sum_{\gamma=(\ta d,\mu',m) \in \tGecC }\!\!\!\! 
  d\cdot \dlecC(i, [\gamma]^\inte) 
=\degC(\tail(i)),  && \notag  \\
\DlecCm(i) +\sum_{\gamma=(\mu,\tb d,m) \in \tGecC }\!\!\!\!
  d\cdot\dlecC(i, [\gamma]^\inte) 
= \degC(\hd(i)),  &&\notag  \\
\DlecCp(i)+\DlecCm(i) \leq 8(1 - \eC(i)),
&& i\in [\widetilde{\kC}+1,\mC],  \notag  \\
\sum_{i\in [\widetilde{\kC}+1,\mC]}\!\!\!\! \dlecC(i, [\gamma]^\inte) =\ecC([\gamma]^\inte),  
&&  \gamma \in \tGecC ,   \label{eq:EC1}   
\end{align}

\begin{align} 
\sum_{\gamma=(\ta d,\tb d',m) \in \tGecT }\!\!\!\! [(\ta,\tb,m)]^\inte\cdot \dlecT(i, [\gamma]^\inte) 
= \sum_{\nu \in \tGacT} [\nu]^\inte\cdot \dlacT(i, [\nu]^\inte) , && \notag  \\  
\DlecTp(i) +\sum_{\gamma=(\ta d,\mu',m) \in \tGecT }\!\!\!\!
  d\cdot  \dlecT(i, [\gamma]^\inte) 
   =\degT(i-1 ),  && \notag  \\
\DlecTm(i) +\sum_{\gamma=(\mu,\tb d,m) \in \tGecT }\!\!\!\! 
 d\cdot \dlecT(i, [\gamma]^\inte) 
 =\degT(i),     &&\notag  \\
\DlecTp(i)+\DlecTm(i) \leq 8(1 - \eT(i)),
&& i\in [2,\tT],   \notag  \\
 \sum_{ i\in [2,\tT]} \!\! \dlecT(i, [\gamma]^\inte) =\ecT([\gamma]^\inte),    
&& \gamma \in \tGecT ,    \label{eq:EC2} 
\end{align}     

\begin{align} 
\sum_{\gamma=(\ta d,\tb d',m) \in \tGecF }\!\!\!\! 
[(\ta,\tb,m)]^\inte\cdot \dlecF(i, [\gamma]^\inte) 
= \sum_{\nu \in \tGacF } [\nu]^\inte\cdot \dlacF(i, [\nu]^\inte) , && \notag  \\  
\DlecFp(i) +\sum_{\gamma=(\ta d,\mu',m) \in \tGecF }\!\!\!\! 
  d\cdot \dlecF(i, [\gamma]^\inte) 
=\degF(i-1 ),    && \notag  \\
\DlecFm(i) +\sum_{\gamma=(\mu,\tb d, m) \in \tGecF }\!\!\!\! 
 d\cdot  \dlecF(i, [\gamma]^\inte) 
=\degF(i,0 ),    &&\notag  \\
\DlecFp(i)+\DlecFm(i) \leq 8(1 - \eF(i)),
&& i\in [2,\tF],   \notag  \\
  \sum_{ i\in [2,\tF]} \!\! \dlecF(i, [\gamma]^\inte) =\ecF([\gamma]^\inte),  
 &&  \gamma \in \tGecF ,    \label{eq:EC3} 
\end{align}    
  

\begin{align} 
\degT(i)+4(1-\chiT(i,k)+\eT(i))\geq \degCTT(k),  
 && \notag  \\  
\degCTT(k)\geq \degT(i)- 4(1-\chiT(i,k)+\eT(i)), &&  i\in [1,\tT],    \notag  \\  
\sum_{\gamma=(\ta d,\tb d',m) \in \tGecCT }\!\!\!\! 
[(\ta,\tb,m)]^\inte\cdot \dlecCTC(k, [\gamma]^\inte) 
= \sum_{\nu \in \tGacCT} [\nu]^\inte\cdot \dlacCT(k, [\nu]^\inte) , && \notag  \\  
\DlecCTp(k) +\sum_{\gamma=(\ta d,\mu',m) \in \tGecCT }\!\!\!\! 
  d\cdot \dlecCTC(k, [\gamma]^\inte) 
=\degC(\tail(k)),    && \notag  \\
\DlecCTm(k) +\sum_{\gamma=(\mu,\tb d, m) \in \tGecCT }\!\!\!\! 
 d\cdot  \dlecCTC(k, [\gamma]^\inte) 
= \degCTT(k),     &&\notag  \\
\DlecCTp(k)+\DlecCTm(k) \leq 8(1 - \dclrT(k)),
&& k\in [1, \kC],  \notag  \\
\sum_{k\in [1, \kC]}\!\! \dlecCTC(k, [\gamma]^\inte) =\ecCT([\gamma]^\inte),  
 && \gamma \in  \tGecCT ,  \label{eq:EC5} 
\end{align}

\begin{align} 
 \degT(i)+4(1-\chiT(i,k)+\eT(i+1))\geq \degTCT(k),  
   && \notag  \\  
\degTCT(k)\geq \degT(i)- 4(1-\chiT(i,k)+\eT(i+1)), 
&&  i\in [1,\tT],    \notag  \\  
\sum_{\gamma=(\ta d,\tb d',m) \in \tGecTC }\!\!\!\! 
[(\ta,\tb,m)]^\inte\cdot \dlecTCC(k, [\gamma]^\inte) 
= \sum_{\nu \in \tGacTC} [\nu]^\inte\cdot \dlacTC(k, [\nu]^\inte) , && \notag  \\  
\DlecTCp(k) +\sum_{\gamma=(\ta d,\mu',m) \in \tGecTC }\!\!\!\! 
  d\cdot \dlecTCC(k, [\gamma]^\inte) 
= \degTCT(k),     &&\notag  \\
\DlecTCm(k) +\sum_{\gamma=(\mu,\tb d, m) \in \tGecTC }\!\!\!\! 
 d\cdot  \dlecTCC(k, [\gamma]^\inte) 
=\degC(\hd(k)),    && \notag  \\
\DlecTCp(k)+\DlecTCm(k) \leq 8(1 - \dclrT(k)),
&& k\in [1, \kC],  \notag  \\
\sum_{k\in [1, \kC]}\!\! \dlecTCC(k, [\gamma]^\inte) =\ecTC([\gamma]^\inte),  
 && \gamma \in  \tGecTC ,  \label{eq:EC5} 
\end{align}

\begin{align} 
\degF(i)+4(1-\chiF(i,c)+\eF(i))\geq \degCFF(c), 
   && \notag  \\  
\degCFF(c)\geq \degF(i)- 4(1-\chiF(i,c)+\eF(i)), 
 && i\in [1,\tF],   \notag  \\   
\sum_{\gamma=(\ta d,\tb d',m) \in \tGecCF }\!\!\!\! 
[(\ta,\tb,m)]^\inte\cdot \dlecCFC(c, [\gamma]^\inte) 
= \sum_{\nu \in \tGacCF} [\nu]^\inte\cdot \dlacCF(c, [\nu]^\inte) , && \notag  \\  
\DlecCFp(c) +\sum_{\gamma=(\ta d,\mu',m) \in \tGecCF }\!\!\!\! 
  d\cdot \dlecCFC(c, [\gamma]^\inte) 
=\degC(c),    && \notag  \\
\DlecCFm(c) +\sum_{\gamma=(\mu,\tb d, m) \in \tGecCF }\!\!\!\! 
 d\cdot  \dlecCFC(c, [\gamma]^\inte) 
= \degCFF(c),     &&\notag  \\
\DlecCFp(c)+\DlecCFm(c) \leq 8(1 - \dclrF(c)),
&& c\in [1,\widetilde{\tC}],  \notag  \\
\sum_{c\in [1,\widetilde{\tC}]}\!\! \dlecCFC(c, [\gamma]^\inte) =\ecCF([\gamma]^\inte),  
 && \gamma \in  \tGecCF ,  \label{eq:EC6} 
\end{align}

\begin{align} 
\degF(j)+4(1-\chiF(j,i+\widetilde{\tC})+\eF(j))\geq \degTFF(i), 
  && \notag  \\  
\degTFF(i)\geq \degF(j)- 4(1-\chiF(j,i+\widetilde{\tC})+\eF(j)), 
 && j\in [1,\tF],   \notag  \\ 
\sum_{\gamma=(\ta d,\tb d',m) \in \tGecTF }\!\!\!\! 
[(\ta,\tb,m)]^\inte\cdot \dlecTFT(i, [\gamma]^\inte) 
= \sum_{\nu \in \tGacTF} [\nu]^\inte\cdot \dlacTF(i, [\nu]^\inte) , && \notag  \\  
\DlecTFp(i) +\sum_{\gamma=(\ta d,\mu',m) \in \tGecTF }\!\!\!\! 
  d\cdot \dlecTFT(i, [\gamma]^\inte) 
=\degT(i),    && \notag  \\
\DlecTFm(i) +\sum_{\gamma=(\mu,\tb d, m) \in \tGecTF }\!\!\!\! 
 d\cdot  \dlecTFT(i, [\gamma]^\inte) 
= \degTFF(i),     &&\notag  \\
\DlecTFp(i)+\DlecTFm(i) \leq 8(1 - \dclrF(i+\widetilde{\tC})),
&& i\in [1,\tT],  \notag  \\
\sum_{i\in [1,\tT]}\!\! \dlecTFT(i, [\gamma]^\inte) =\ecTF([\gamma]^\inte),  
 && \gamma \in  \tGecTF ,  \label{eq:EC6} 
\end{align}

\begin{align} 
\sum_{\mathrm{X}\in\{\mathrm{C,T,F,CT,TC,CF,TF}\}}(\ecX([\gamma]^\inte)
+\ecX([\overline{\gamma}]^\inte)) 
 =\ec^\inte([\gamma]^\inte) , &&  \gamma\in \Gamma_{<}^\inte,  \notag \\  
\sum_{\mathrm{X}\in\{\mathrm{C,T,F,CT,TC,CF,TF}\}} \ecX([\gamma]^\inte) 
 =\ec^\inte([\gamma]^\inte) , && \gamma\in \Gamma_{=}^\inte. 
   \label{eq:EC_last} 
\end{align}


\subsection{Constraints for Standardization  of Feature Vectors} 
\label{sec:NSFV}

  By introducing a tolerance $\varepsilon>0$   in the conversion 
 between integers and reals, we include the following constraints for 
standardizing of a  feature vector   $x=(x(1),x(2),\ldots,x(K))$: 
\begin{equation}\label{eq:standardization2}
\frac{(1 - \varepsilon)(x(j)  - \min(\dcp_j;D_\pi))}
{ \max(\dcp_j;D_\pi) - \min(\dcp_j;D_\pi) }
\leq \widehat{x}(j)   \leq 
\frac{(1 + \varepsilon) ( x(j) - \min(\dcp_j;D_\pi))}
{ \max(\dcp_j;D_\pi) - \min(\dcp_j;D_\pi) }, 
~ j\in [1,K].  
\end{equation} 
An example of  a tolerance is $\varepsilon=1\times 10^{-5}$.

We use the same conversion for descriptor $x_j = \overline{\mathrm{ms}}$. 


\end{document}